\documentclass{article}
\usepackage{PRIMEarxiv}
\usepackage{kotex}
\usepackage{multirow}
\usepackage{makecell}
\usepackage{xcolor}
\usepackage{float}
\usepackage{hyperref}
\usepackage{algorithm}
\usepackage{float}
\usepackage{amsmath}
\usepackage{amsfonts}
\usepackage{booktabs}
\usepackage{algorithmic}
\usepackage{graphicx}     
\usepackage{subcaption}   
\usepackage{float} 
\graphicspath{{media/}}     

\title{Variational Approach for Job Shop Scheduling\thanks{This paper includes supplementary material.}}

\author{
  Seung Heon Oh\\
    Republic of Korea Navy \\
    \texttt{suenghuny@snu.ac.kr} \\
   \And
    Jiwon Baek, Hyunjin Oh, Kiyoung Cho, Heechang Yoon \\
  Department of Naval Architecture and Ocean Engineering \\
  Seoul National University \\
   \And
  Jong Hun Woo \\
  Department of Naval Architecture and Ocean Engineering \\
  Seoul National University \\
    Research Institute of Marine Systems Engineering\\
    Seoul National University \\
  \texttt{j.woo@snu.ac.kr} \\
}

\begin{document}
\maketitle
\thispagestyle{fancy} 

\begin{abstract}
This paper proposes a novel Variational Graph-to-Scheduler (VG2S) framework for solving the Job Shop Scheduling Problem (JSSP), a critical task in manufacturing that directly impacts operational efficiency and resource utilization. Conventional Deep Reinforcement Learning (DRL) approaches often face challenges such as non-stationarity during training and limited generalization to unseen problem instances because they optimize representation learning and policy execution simultaneously. To address these issues, we introduce variational inference to the JSSP domain for the first time and derive a probabilistic objective based on the Evidence of Lower Bound (ELBO) with maximum entropy reinforcement learning. By mathematically decoupling representation learning from policy optimization, the VG2S framework enables the agent to learn robust structural representations of scheduling instances through a variational graph encoder. This approach significantly enhances training stability and robustness against hyperparameter variations. Extensive experiments demonstrate that the proposed method exhibits superior zero-shot generalization compared with state-of-the-art DRL baselines and traditional dispatching rules, particularly on large-scale and challenging benchmark instances such as DMU and SWV.
\end{abstract}

\section{Introduction}
\textcolor{black}{The} Job Shop Scheduling Problem (JSSP) is a fundamental challenge in manufacturing, directly impacting the operational efficiency and resource utilization of various manufacturing systems. While traditional methods—such as heuristics, meta-heuristics, and mathematical programming—have been widely used, their need for instance-specific optimization often limits their flexibility in dynamic shop-floor environments. Recently, Deep Reinforcement Learning (DRL) has emerged as a powerful alternative in various scheduling domains, including manufacturing \cite{waschneck2018deep, park2019reinforcement, cho2022minimize}, logistics \cite{zong2022rbg, phiboonbanakit2021hybrid}, traffic control \cite{9505616, 8676356}, and defense \cite{oh2025wta, li2023weapon}. By leveraging learned policies, it achieves both superior performance and strong generalization across different scenarios. Unlike conventional approaches that require time-consuming re-optimization for every new task, a trained DRL model can be immediately deployed to diverse problem instances \cite{wang2021dynamic, zhang2024collab, liu2020actor, wang2024design}, providing \textcolor{black}{the time-efficient decision-making necessary for manufacturing operations}.
\par
\textcolor{black}{Fundamentally, this capability hinges on how well the model captures the intertwined dependencies among the production components, such as jobs, machines, and operations \cite{gabel2012distributed}. Therefore, recent research~\cite{JSSP_uncertain_processing_time, oh2022distributional, park2019reinforcement, Smart_JSSP, wang2021dynamic, WU2023106401} has focused on designing effective \textcolor{black}{Neural Network (NN)} architectures that can 
transform raw JSSP instances into structurally informative representations for 
decision-making.} From an architectural perspective, early studies applying DRL to JSSP utilized Multi-Layered Perceptron (MLP) \cite{JSSP_uncertain_processing_time, gabel2012distributed,oh2022distributional, park2019reinforcement} or Convolutional Neural Network (CNN) \cite{Smart_JSSP, wang2021dynamic, WU2023106401}. \textcolor{black}{These NNs are highly effective at distilling implicit scheduling-relevant features from JSSP instances represented in vector or matrix formats \cite{Dynamic_JSSP}. However, processing JSSP instances represented as vectors or matrices through NNs poses difficulties in generalizing across varying problem sizes~\cite{park2021learning, wang2024design,
chen2022deep, SAC_JSSP, Dynamic_JSSP}, a critical limitation in real-world manufacturing systems.}
\par
\textcolor{black}{Such a limitation of conventional NN-based approaches stems from the inherent nature of JSSP.} JSSP constraints are represented as a disjunctive graph, embodying unique topological relationships that cannot be readily captured in Euclidean space \cite{1555942}. \textcolor{black}{This characteristic makes JSSP inherently more difficult to solve than other combinatorial 
optimization problems such as the traveling salesman problem and vehicle routing problem \cite{zhang2020learning}.} Indeed, \textcolor{black}{even numerous non-DRL algorithms \textcolor{black}{for JSSP}—including the fast branch and bound algorithm \cite{brucker1994branch} and the shifting bottleneck procedure \cite{adams1988shifting}—have been developed specifically to exploit this disjunctive graph structure \cite{chen2022deep}}. Consequently, \textcolor{black}{this underscores the need for DRL methods capable of effectively handling such graph structures.}
\par
The key to leveraging this topological structure in the context of DRL lies in effective representation learning. \textcolor{black}{Graph Neural Networks (GNNs) are specifically designed to process graph-structured data, making them well-suited for capturing complex relational patterns inherent in JSSP.} \textcolor{black}{This capability has made the GNN a natural framework for learning representations of the disjunctive graph structure in JSSP}, where operations and machines form intricate networks of precedence and resource constraints \cite{Dynamic_JSSP}. GNN-based studies in JSSP can be categorized along two key aspects related to representation learning: (1) graph representation and (2) architecture design. \textit{Graph representation} defines \textbf{what} information is encoded: how JSSP entities (operations, machines) and relationships (precedence, disjunctive constraints) are structured as graph topology and node/edge features. \textit{Architecture design} defines \textbf{how} this information is processed: which GNN architectures and training methodologies extract representations and learn policies.
\par
In terms of \textit{graph representation}, \textcolor{black}{it has been progressively refined to accommodate the unique characteristics of each problem variant}. \cite{park2021learning} and \cite{zhang2020learning} propose a foundational framework representing operations as featured nodes, which has been widely adopted in subsequent works \cite{distributed_JSSP, hierachical, SAC_JSSP, Dynamic_JSSP, Strategy_JSSP, yuan2023solving}. Building on this foundation, researchers have developed specialized representations for diverse JSSP variants: \textcolor{black}{Invalid Action Masking (IAM)} \cite{yuan2023solving}, heterogeneous graphs for dynamic environments \cite{Dynamic_JSSP, SAC_JSSP, Strategy_JSSP}, stitched disjunctive graphs for distributed multi-factory settings \cite{hierachical, distributed_JSSP, huang2023novel}, and heterogeneous representations for flexible job shop problems \cite{moon2024learning, zhang2024collab, cho2025solving}. These studies have effectively expanded the scope of JSSP modeling by developing versatile graph structures that can encode the unique constraints of each problem variant.

\par
In terms of \textit{architecture design}, \textcolor{black}{prior} works have primarily focused on adopting existing GNN architectures. Various GNN architectures have been applied—including Graph Isomorphism Network (GIN) \cite{zhang2020learning, distributed_JSSP, hierachical, SAC_JSSP, yuan2023solving, huang2023novel}, standard GNN layers \cite{park2021learning}, random walk embeddings \cite{chen2022deep}, and Graph Attention Network (GAT) \cite{Dynamic_JSSP, Strategy_JSSP, oh2025framework}. Some recent works \cite{chen2022deep, oh2025framework} have explored graph-to-sequence architectures that structurally decouple the processing of static graph features and dynamic scheduling decisions.
\textcolor{black}{Despite variations in graph representation and network architecture, these methods are universally trained in an end-to-end manner.} This paradigm enables the joint optimization of representation-level feature extraction and policy-level task execution.
\par
\textcolor{black}{While recent GNN-based end-to-end DRL approaches \textcolor{black}{have demonstrated effectiveness} in the JSSP domain, they rely on the implicit assumption that these networks possess the representational capacity to effectively capture the complex constraints of JSSP, thereby generalizing across diverse problem instances. In practice, however, the efficacy of these end-to-end approaches is compromised by their limited generalizability.} While these models show potential in classical JSSP settings, even state-of-the-art approaches \cite{SAC_JSSP, oh2025framework, yuan2023solving} experience significant performance degradation under minor deviations from those conditions. This is evidenced by their reduced performance on the Demirkol et al. (DMU) and Storer et al. (SWV) benchmark datasets, which are widely recognized for their inherent difficulty \cite{demirkol1998benchmarks, xie2022hybrid}. \textcolor{black}{This limited generalization poses a critical barrier to deployment in real-world manufacturing systems, which are inherently characterized by high variability and complexity.}
\par
\textcolor{black}{These shortcomings are rooted in how the standard end-to-end learning paradigm handles 
representation learning: it tightly couples representation learning with policy learning, 
meaning that representations are driven solely by scalar reward signals. However, such signals contain little meaningful information about the target 
representation structure \cite{slac, stooke2021decoupling}. Consequently, these methods fail to secure robust representations, ultimately yielding policies with limited generalization capacity \cite{slac, higgins2017darla, stooke2021decoupling}. These representational bottlenecks further lead to unstable training, sensitivity to hyperparameters, and overall performance degradation \cite{slac}. 
The JSSP domain is not immune to these shortcomings either, as evidenced by the 
limited generalization observed on challenging benchmarks such as DMU and SWV. To bridge this gap in the JSSP domain, we propose a variational approach that explicitly decouples representation learning from policy learning, with the following contributions:
}
\begin{itemize}
\item We derive a rigorous probabilistic objective based on the Evidence Lower Bound (ELBO) with maximum entropy reinforcement learning (RL) for JSSP. 
 \textcolor{black}{This provides a theoretical framework not only to handle the latent stochasticity inherent in the manufacturing environment but also to explicitly decouple the representation learning of graph-structured JSSP instances from the reward-driven policy learning.}

\item We introduce VG2S, a Variational Graph-to-Scheduler framework that integrates a variational graph encoder with a sequence-based policy decoder. \textcolor{black}{By adopting a two-phase training strategy \textcolor{black}{based on ELBO}, our approach achieves high learning stability.}

\item \textcolor{black}{We empirically demonstrate that, in the JSSP domain, variational representation learning via two-phase training yields a structurally aware latent space, and provide empirical evidence that this structured latent space is a key contributor to superior zero-shot generalization.}

\end{itemize}

\section{Problem definition}
A JSSP is characterized by the triple $\langle\mathcal{J}, \mathcal{O}, \mathcal{M}\rangle$, which comprises a set of $n$ jobs $\mathcal{J} = \{J_1, \ldots, J_n\}$, a set of $m$ machines $\mathcal{M} = \{M_1, \ldots, M_m\}$, and a set of operations $\mathcal{O}$. 
Each job $J_j \in \mathcal{J}$ is subject to precedence constraints, requiring its constituent operations to be processed in a fixed, predetermined sequence. 
\textcolor{black}{An operation $O_{ij}$ belongs to job $J_j$, is processed on machine $M_i \in \mathcal{M}$, and has a processing duration $p_{ij}$.}
\par
The objective is to minimize the makespan $C_{\max}$, defined as the total duration required to complete all jobs. This optimization is conducted under several standard assumptions: operations are non-preemptive, each machine can process only one operation at a time, and all jobs are available simultaneously at the start of the schedule. Furthermore, any time associated with machine setup or job transportation is considered negligible.
\par
A JSSP instance is formally represented as a disjunctive graph $G = (V, C \cup D)$, where $V$ denotes the set of all operations. 
The set of conjunctive edges $C$ encodes precedence constraints; specifically, a directed edge $(u, v) \in C$ indicates that operation $u$ must be completed before operation $v$ can begin within the same job. 
The set of disjunctive edges $D$ represents machine-sharing relations, where an undirected edge connects any two operations required to be processed on the same machine, signifying that their execution order must be determined to avoid resource conflicts.
\begin{equation*}
\begin{aligned}
    \text{minimize} &\quad C_{\max} \\
     \text{subject to}&\quad y_{ij} - y_{kj} \geq p_{kj} \quad \forall (k, j) \to (i, j) \in C \\
     &\quad y_{ij} - y_{il} \geq p_{il} \quad \text{or} \quad y_{il} - y_{ij} \geq p_{ij} \\
     &\quad \forall (i, j) \text{ and } (i, l), \; i = 1, \ldots, m \\
     &\quad C_{\max} - y_{ij} \geq p_{ij} \quad \forall (i, j) \in V \\
     &\quad y_{ij} \geq 0 \quad \forall (i, j) \in V 
\end{aligned}
\label{static}
\end{equation*}
where the decision variable $y_{ij}$ denotes the starting time of operation $O_{ij}$, and $p_{ij}$ is its processing time.

\section{Representing JSSP instance}
To effectively model a JSSP instance, we distinguish between its static attributes and dynamic characteristics. \textcolor{black}{This separation in terms of data representation naturally corresponds to the architectural decoupling between the variational graph encoder and the policy decoder, introduced in Section~\ref{VG2S framework for jssp}.} Static information, representing the invariant topology of the problem, is modeled as a heterogeneous graph in Section~\ref{het graph}. Dynamic information, elaborated in Section~\ref{state feature}, captures the state transitions of the evolving partial schedule.
\subsection{Static information}
\label{het graph}
\textcolor{black}{We represent each JSSP instance as a heterogeneous graph 
$G = (\mathcal{V}, (\mathcal{E}_1, \mathcal{E}_2, \mathcal{E}_3))$, 
where each node $v \in \mathcal{V}$ represents an operation and is 
associated with six distinct features from \cite{oh2025framework}.} The edges $\mathcal{E}_1, \mathcal{E}_2,$ and $\mathcal{E}_3$ represent precedence, successor, and machine-sharing dependencies. To ensure generalization across diverse JSSP instances, operation features are normalized to reflect relative values within each instance. These features ($\mathbf{x}_{ij}^1, \dots, \mathbf{x}_{ij}^6$) capture instance-specific semantics as follows:
\begin{equation*}
\begin{aligned}
\mathbf{x}_{ij}^1 = \frac{p_{ij}}{\sum_{M_{i'} \in \mathcal{M}} p_{i'j}}, \quad \mathbf{x}_{ij}^2 =\frac{p_{ij}}{\max_{M_{i'} \in \mathcal{M}} p_{i'j}} , \quad \mathbf{x}_{ij}^3 = \frac{p_{ij}}{\sum_{J_{j'} \in \mathcal{J}} p_{ij'}}
\end{aligned}
\label{processing time}
\end{equation*}

\begin{equation*}
\begin{aligned}
\mathbf{x}_{ij}^4 = \frac{\sum_{M_{i'} \in \mathcal{A}_{ij}}p_{i'j}}{\sum_{M_{i'} \in \mathcal{M}} p_{i'j}}, \quad \mathbf{x}_{ij}^5 = \frac{|\mathcal{A}_{ij}|}{|\mathcal{M}|}, \quad \mathbf{x}_{ij}^6 = \frac{\sum_{M_{i'} \in \mathcal{M}} p_{i'j}}{\max_{j'}\sum_{M_{i'} \in \mathcal{M}} p_{i'j'}}
\end{aligned}
\label{historical work}
\end{equation*}
$\mathcal{A}_{ij}$ denotes the set of machines comprising the current machine for $O_{ij}$ and all machines previously visited by job $J_j$. The node set $\mathcal{V}$ is represented as a feature matrix $\mathbf{X} \in \mathbb{R}^{N_O \times 6}$, where each row $\mathbf{x}_{u} = (\mathbf{x}_{u}^1, \dots, \mathbf{x}_{u}^6)$ is the feature vector of operation $O_u$. The total number of nodes is defined as $N_O = nm + 2$ to account for two additional dummy nodes: a source node $(0,0,0,0,0,0)$ representing the start of the schedule, and a sink node $(0,0,1,1,1,0)$ representing its completion. $\mathcal{N}_{\mathcal{E}_\cdot}(u)$ defines the set of neighboring operations of $O_u$ connected via edges in $\mathcal{E}_\cdot$. Specifically:
\begin{itemize}
    \item $\mathcal{N}_{\mathcal{E}_1}(u)$: the precedent operation index of $O_u$
    \item $\mathcal{N}_{\mathcal{E}_2}(u)$: the successor operation index of $O_u$
    \item $\mathcal{N}_{\mathcal{E}_3}(u)$: machine-sharing operation indices of $O_u$
\end{itemize}
\par
Feature-wise ablation results are provided in Supplementary Material F.1, 
and \textcolor{black}{an analysis on the effect of precedence edges} is provided in Supplementary Material F.2.

\textbf{Remark.} We adopt two notations for operations: $O_{ij}$ represents job $j$ on machine $i$ to highlight specific job-machine associations, while $O_u$ serves as a simplified single-index notation within the context of the NN architecture. \textcolor{black}{Here, $O_0$ and $O_{nm+1}$ denote the source and sink dummy nodes, respectively, where $u \in \{0, 1, \ldots, nm+1\}$.}

\subsection{Dynamic information}
\label{state feature}
The JSSP instance $G$ is processed by the initial state function $g_0$ to produce the initial state $s_1 = g_0(G)$. The environment then follows the transition dynamics $s_{t+1} = g(s_t, a_t, G)$, where $s_t$ denotes the \textcolor{black}{true} state at each decoding timestep $t$ and $a_t$ represents the action indicating the selected operation at timestep $t$. Upon the state transition, the selected action is assigned to the designated machine and executed immediately as the machine becomes available, ensuring zero unnecessary idle time (i.e., semi-active scheduling; Details are provided in the Supplementary Material A).
\par
\textcolor{black}{The state features \textcolor{black}{$s_{1,t}^{ij}$--$s_{6,t}^{ij}$} are employed to capture the dynamic scheduling information at each timestep. \textcolor{black}{$s_{1,t}^{ij}$--$s_{4,t}^{ij}$} are directly adopted from \cite{oh2025framework}, while \textcolor{black}{$s_{5,t}^{ij}$} and \textcolor{black}{$s_{6,t}^{ij}$} are derived following the state feature design perspective established therein.} We define $\mathcal{A}(s_t)$ as the set of available operations at state $s_t$. The state features $s_{1, t}^{ij}$ and $s_{2, t}^{ij}$ denote the earliest start and finish times for each action $a_t = O_{ij} \in \mathcal{A}(s_t)$ and are defined as follows:
\begin{align*}
s_{1,t}^{ij} &=\max\{\mathbf{m}^i_t, \mathbf{j}^j_t\}\\
s_{2,t}^{ij} &=\max\{\mathbf{m}^i_t, \mathbf{j}^j_t\}+p_{ij}
\label{dynamics2}
\end{align*}
The dynamic variables $\mathbf{m}^i_t$ and $\mathbf{j}^j_t$ represent the ready times for machine $i$ and job $j$ at time $t$, respectively. Details of $\mathbf{m}^i_t$ and $\mathbf{j}^j_t$ are provided in the Supplementary Material A. Here, $s_{1,t}^{ij}$ and $s_{2,t}^{ij}$ are set to zero whenever an operation $O_{ij}$ is outside the set $\mathcal{A}(s_t)$. To ensure consistent input scales for the model, we normalize these values into the range between 0 and 1, using $\max_{ij}s_{1,t}^{ij}$ and $\max_{ij}s_{2,t}^{ij}$ as the scaling factors for each feature. We employ the \textit{efficient lower bound} from \cite{oh2025framework} by utilizing state features $s^{ij}_{3,t}$ and $s^{ij}_{4,t}$ as follows: 
\begin{align*}
s^{ij}_{3,t} &= \max\{\mathbf{m}_i(\tilde{s}^{ij}(s_t))+\mathbf{r}_i(\tilde{s}^{ij}(s_t)),
\mathbf{j}_j(\tilde{s}^{ij}(s_t))+\mathbf{r}_j(\tilde{s}^{ij}(s_t))\}\\
s^{ij}_{4,t} &= \max_{\substack{J_{\hat{j}}\in\mathcal{J}\\O_{\hat{i}\hat{j}}\in \mathcal{A}(\tilde{s}^{ij}(s_t))}}\max\{\mathbf{m}_{\hat{i}}(\tilde{s}^{ij}(s_t))+\mathbf{r}_{\hat{i}}(\textcolor{black}{\tilde{s}^{ij}(s_t)}),
\mathbf{j}_{\hat{j}}(\tilde{s}^{ij}(s_t))+\mathbf{r}_{\hat{j}}(\tilde{s}^{ij}(s_t))\}
\end{align*}
Here, the variables $\mathbf{m}_i(s)$ and $\mathbf{j}_j(s)$ correspond to the machine $i$ and job $j$ ready times in state $s$, while $\mathbf{r}_i(s)$ and $\mathbf{r}_j(s)$ signify their respective remaining processing times in state $s$. Despite being less restrictive than standard bounds, this formulation is selected for its straightforward implementation and minimal computational requirements during execution. The successor state resulting from the selection of operation $O_{ij}$ at $s_t$ is denoted by $\tilde{s}^{ij}(s_t)$. In cases where $O_{ij}$ is the terminal operation of job $J_j$, the state feature $s^{ij}_{3,t}$ along with the terms $\mathbf{m}_{\hat{i}}(\tilde{s}^{ij}(s_t))+\mathbf{r}_{\hat{i}}(\tilde{s}^{ij}(s_t))$ and $\mathbf{j}_{\hat{j}}(\tilde{s}^{ij}(s_t))+\mathbf{r}_{\hat{j}}(\tilde{s}^{ij}(s_t))$ are defined to be zero. State features $s^{ij}_{5,t}$ and $s^{ij}_{6,t}$ are introduced to capture the global progress of the scheduling process as follows:
\begin{align*}
s^{ij}_{5,t} &= \max_{J_{\hat{j}}\in\mathcal{J}} \mathbf{C}_{\hat{j}}(\tilde{s}^{ij}(s_t))\\
s^{ij}_{6,t} &= \operatorname{mean}_{J_{\hat{j}}\in\mathcal{J}} \mathbf{C}_{\hat{j}}(\tilde{s}^{ij}(s_t))
\end{align*}
Here, $\mathbf{C}_{\hat{j}}(\textcolor{black}{\tilde{s}^{ij}(s_t}))$ denotes the number of completed operations for job $J_{\hat{j}}$ in the \textcolor{black}{successor} state $\tilde{s}^{ij}(s_t)$ transitioned by selecting operation $O_{ij}$. Specifically, $s^{ij}_{5,t}$ represents the maximum number of completed operations among all jobs, indicating the progress of the most advanced job. Meanwhile, $s^{ij}_{6,t}$ represents the average number of completed operations across all jobs, providing a global measure of the overall completion status of the entire instance. The state features $s^{ij}_{5,t}$ and $s^{ij}_{6,t}$ are normalized by the total number of machines $m$. Consequently, the state feature vector $s_t^u \in \mathbb{R}^6$ of $O_u$ indicates $(s^{u}_{1,t}, s^{u}_{2,t}, s^{u}_{3,t},  s^{u}_{4,t}, s^{u}_{5,t}, s^{u}_{6,t})$. \textcolor{black}{Note that state features are defined only for non-dummy operations, i.e., $u \in \{1, \ldots, nm\}$, as dummy nodes $O_0$ and $O_{nm+1}$ are never included in $\mathcal{A}(s_t)$. \textcolor{black}{Ablation studies validating the contribution of each dynamic state feature group are provided in Supplementary Material F.3.}}

\section{Evidence lower bound with maximum entropy RL for JSSP}
\label{ELBO with maximum entropy RL for JSSP}
While state transitions within a single JSSP instance are deterministic, achieving robust generalization over diverse scheduling instances requires addressing the uncertainty inherent across the problem space. \textcolor{black}{Handling this uncertainty underpins the probabilistic formulation of representation learning for JSSP in our framework.}
To model \textcolor{black}{the} uncertainty, we introduce the distribution of instances $p(G)$, representing the likelihood of encountering a specific problem instance $G$, and the conditional probability of achieving optimality $p(O=1|G)$. 
Here, $O$ denotes a binary random variable indicating whether optimality is achieved. 
Consequently, the objective function is formulated as maximizing the joint probability $p(G, O=1)$ as follows:
\begin{equation}
\label{objective}
p(G, O=1) = p(G) \cdot p(O=1|G)
\end{equation}
This objective function mathematically formalizes the goal of robust generalization by integrating the instance distribution $p(G)$, thereby preventing overfitting to specific instances and promoting consistent performance across a wide range of problems. By marginalizing $p(G, O=1)$ over the latent variable $z$ and the action sequence $a$, we can express the lower bound on the log-likelihood of the objective function as \eqref{first elbo} -- \eqref{last elbo}. Derivations are provided in the Supplementary Material B.
\begin{align}
\log p(G,O) &= \log \sum_{z,a}[p(G,O,z,a) \cdot \frac{q(z,a|G)}{q(z,a|G)}] \label{first elbo}  \\
&\geq \underbrace{\mathbb{E}_{z,a \sim q}[\log p(G|z)] - D_{\text{KL}}[q(z|G)||p(z)]}_{\text{Reconstruction loss term}}+ \underbrace{\mathbb{E}_{z,a \sim q}[Q(z,a) - \log \pi(a|z,G) + \log p(a|G,z)]}_{\text{Policy loss term}}
\label{last elbo}
\end{align}
Here, the posterior $q(z,a|G)$ factorizes into the posterior $q(z|G)$ and the policy $\pi(a|z,G)$: $q(z,a|G) = q(z|G)\cdot  \pi(a|z,G)$. The derivation in \eqref{first elbo} -- \eqref{last elbo} assumes that $p(O=1|G, z, a)$ is proportional to the exponential of $Q(z, a)$: $p(O=1|G, z, a) \propto \exp Q(z, a)$. Here, $Q(z, a)$ represents the scheduling objective score (e.g., negative makespan) for a given latent variable $z$ and action sequence $a$. \textcolor{black}{By adopting this exponential formulation—a standard approach in maximum entropy RL —the unbounded objective score is mapped into a differentiable probability space. This framework establishes a mathematical link between reward-driven policy learning and probabilistic optimality. Consequently, as the negative makespan $Q(z,a)$ is maximized, the probability of the optimal trajectory exponentially approaches unity.} \eqref{last elbo} can be decomposed into ELBO consisting of two terms: reconstruction loss and policy loss term. This corresponds to a single-step scheduling adaptation of variational \textcolor{black}{representation learning} with multi-step maximum entropy RL \cite{slac}. By maximizing this ELBO, we can effectively optimize the original objective function of \eqref{objective}. In particular, under the uniform prior assumption for $\log p(a|G,z)$, the policy loss term reduces to a maximum entropy RL formulation with baseline, corresponding to the policy gradient with maximum entropy in \eqref{PG with maximum entropy}.
\begin{equation}
\label{PG with maximum entropy}
J(\pi) = \mathbb{E}_{z \sim q}\mathbb{E}_{a \sim \pi}[Q(z,a)-V(z) - \log \pi(a|z, G)]
\end{equation}
where $J(\pi)$ is the maximum entropy RL objective. This analysis provides theoretical justification \textcolor{black}{for decoupling variational representation learning from reward-based policy learning}. Derivations are provided in the Supplementary Material B.

\section{Variational graph-to-scheduler framework for JSSP}
\label{VG2S framework for jssp}
\begin{figure}
    \centering
    \includegraphics[width=\textwidth]{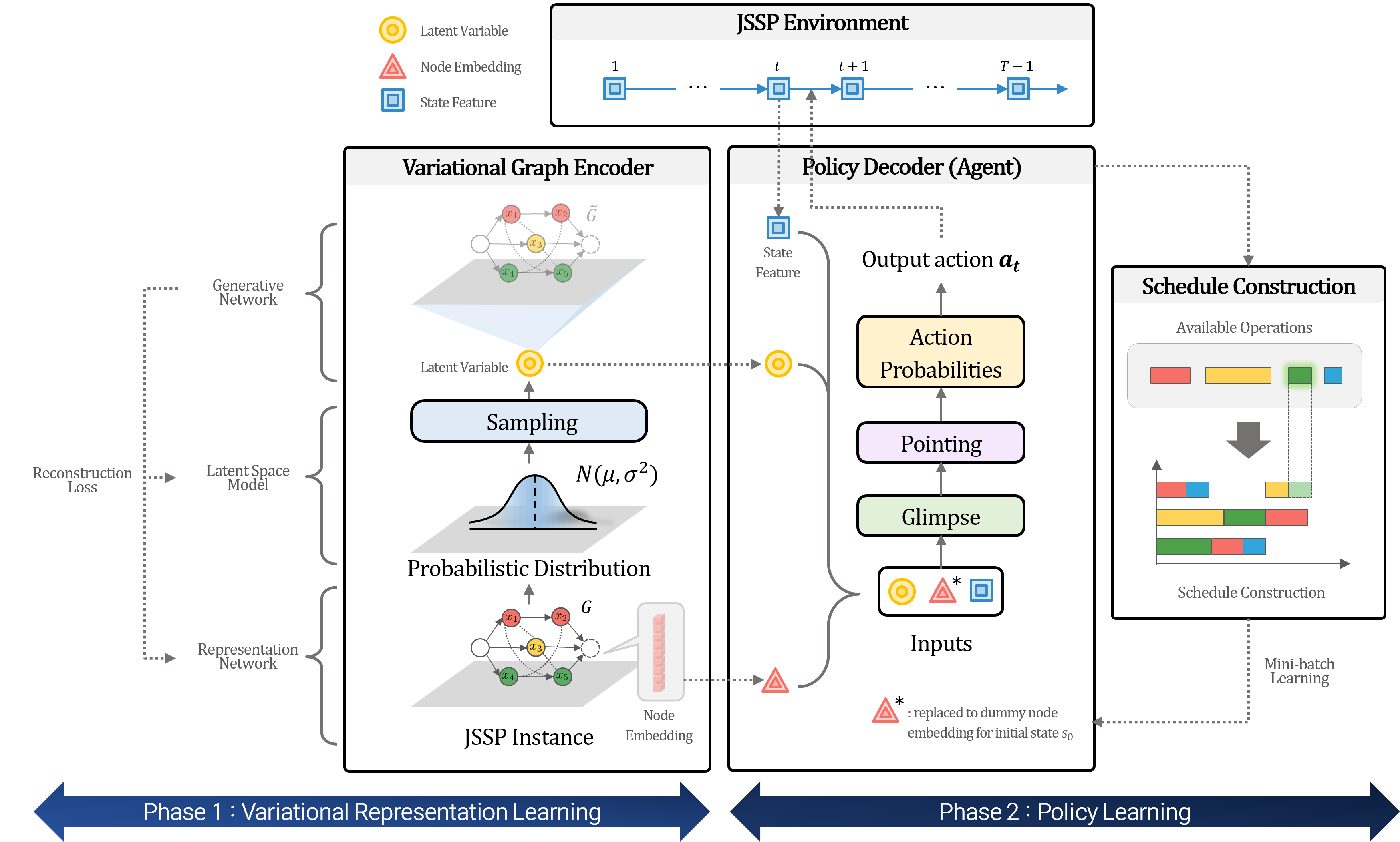}
    \caption{Overview of the proposed VG2S framework}
    \label{framework}
\end{figure}

We propose a Variational Graph-to-Scheduler (VG2S) framework for the JSSP, which integrates the aforementioned theoretical insights with a Variational Autoencoder (VAE) approach \cite{kingma2013auto} and a graph-to-sequence architecture \cite{chen2022deep, oh2022distributional}. The proposed framework consists of \textcolor{black}{a} variational graph encoder (Section~\ref{Variational graph encoder}) and \textcolor{black}{a} policy decoder (Section~\ref{Operation selecting decoder}), as illustrated in Fig.~\ref{framework}. The training process is divided into two stages: variational representation learning and policy learning. During the variational representation learning stage, by optimizing the ELBO—specifically the reconstruction loss —the model is compelled to distill the complex structural dependencies and constraints of the JSSP into a highly informative and compact latent space. This process yields a more robust and expressive representation of the problem instances, which ultimately \textcolor{black}{leads to higher-quality scheduling decisions by} the policy decoder. \textcolor{black}{In the policy learning stage, the policy loss is optimized using rollout samples $(G, a, C_{\max})$, where $G$ is the graph-represented JSSP instance, $a$ is the action sequence, and $C_{\max}$ is the resulting makespan.}
\subsection{Variational graph encoder}
\label{Variational graph encoder}
The variational graph encoder is composed of a representation network (Section~\ref{Representation network}) for inferring latent representations, a latent space model (Section~\ref{Latent space model}) for capturing the distributional properties of the learned representations, and a generative network (Section~\ref{Generative network}) for graph reconstruction. \textcolor{black}{We emphasize that while the framework incorporates a generative architecture, its primary objective is ELBO-driven representation learning, not the direct utilization of reconstructed graph instances.} The overall network architecture is illustrated in Fig.~\ref{vge}.
\subsubsection{Representation network}
\label{Representation network}
\begin{figure}
    \centering
    \includegraphics[width=\textwidth]{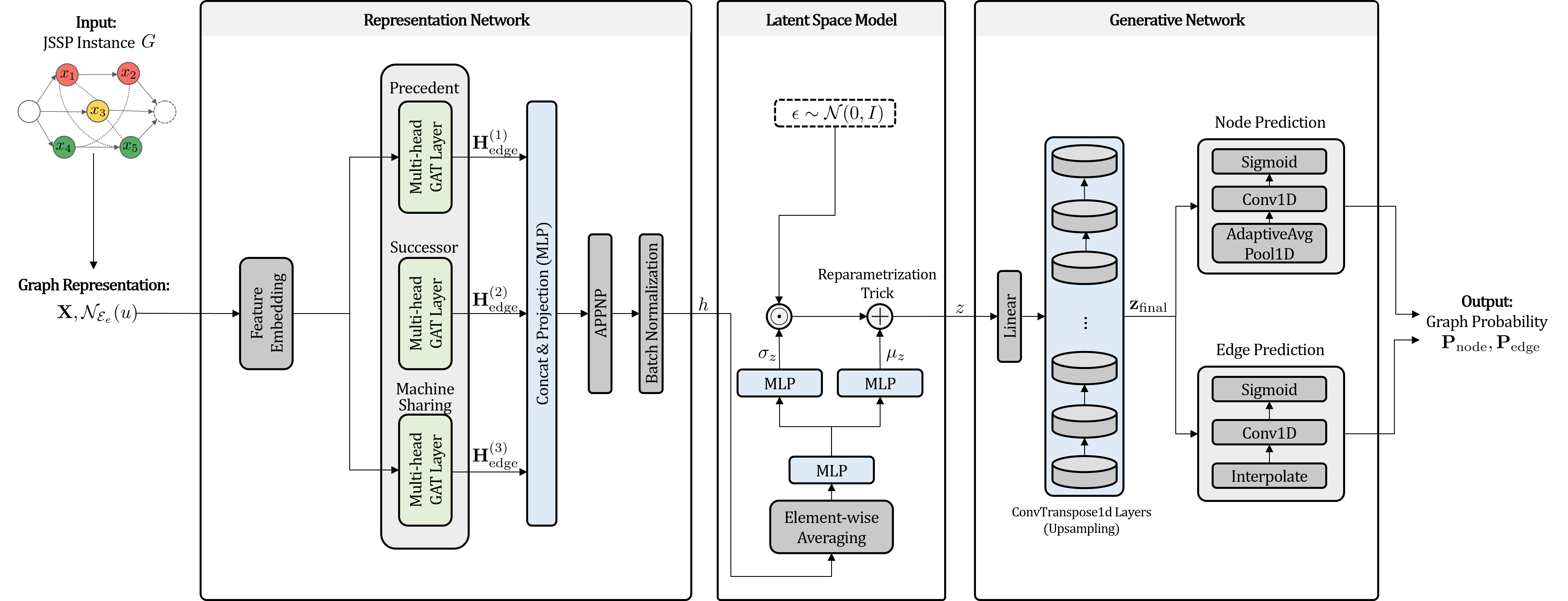}
    \caption{Architecture of variational graph encoder}
    \label{vge}
\end{figure}

In our architecture, \textcolor{black}{a} JSSP instance $G$ is fed into the representation network $f_{\phi_1}$ parametrized by $\phi_1$, which outputs node embeddings $h$ as follows:
\begin{equation*}
h = f_{\phi_1}(G)
\end{equation*}
\noindent
The representation network consists of two main components: a feature embedding layer and GNN layers. The feature embedding layer applies MLP to map input node features $\mathbf{X} \in \mathbb{R}^{N_O \times 6}$ from dimension $6$ to $d_{\text{graph}}$:
\begin{equation*}
H_{\text{embed}} = \operatorname{MLP}_{\text{embed}} (\mathbf{X})
\end{equation*}
where $H_{\text{embed}}\in \mathbb{R}^{N_O \times d_{\text{graph}}}$ is the embedded feature matrix whose $u$-th row represents the node embedding of $O_u$. GNN layers employ multi-head GAT \cite{RN68}, which processes heterogeneous graphs with multiple edge types through multi-head attention mechanisms. For each edge type $e \in \{1, 2,3\}$ and attention head $m \in \{1, 2, ..., M\}$, the layer computes:

\begin{equation*}
H_{\operatorname{linear}}^{(e,m)} = H_{\operatorname{embed}} W_s^{(e,m)}
\end{equation*}
$W_s^{(e,m)} \in \mathbb{R}^{d_{\operatorname{graph}} \times d_{\text{graph}}}$ is the learnable weight matrix for edge type $e$ and head $m$.
The attention mechanism computes edge-specific attention coefficients:
\begin{equation*}
a^{(e,m)}_{uv} = \text{LeakyReLU}(a_1^{(e,m)\top} H^{(e,m)}_{\text{linear},u} + a_2^{(e,m)\top} H^{(e,m)}_{\text{linear},v}), \quad \alpha^{(e,m)}_{uv} = \frac{\exp(a^{(e,m)}_{uv})}{\sum_{v' \in \mathcal{N}_{\mathcal{E}_{e}}(u)} \exp(a^{(e,m)}_{uv'})}
\end{equation*}
where $H^{(e,m)}_{\text{linear},u}$ is the element corresponding to node $u$ in $H_{\operatorname{linear}}^{(e,m)}$ and $a_1^{(e,m)}, a_2^{(e,m)} \in \mathbb{R}^{d_\text{graph}}$ are learnable attention parameters. The output of node $u$ for edge type $e$ and head $m$, denoted as $H^{(e,m)}_{\text{edge}, u}  \in \mathbb{R}^{d_{\textcolor{black}{\text{graph}}}}$, is computed as follows:
\begin{equation*}
H^{(e,m)}_{\text{edge},u} = \text{ELU}\left(\sum_{v' \in \mathcal{N}_{\mathcal{E}_{e}}(u)} \alpha^{(e,m)}_{uv'} H^{(e,m)}_{\text{linear},v'} \right)
\end{equation*}
 and then concatenated to form $H^{(e)}_{\text{edge},u} \in \mathbb{R}^{Md_{\textcolor{black}{\text{graph}}}}$ as:
\begin{equation*}
H^{(e)}_{\text{edge},u} = \text{Concat}(H^{(e,1)}_{\text{edge},u}, ..., H^{(e,M)}_{\text{edge},u})
\end{equation*}
The embeddings $H^{(e)}_{\text{edge},u}$ computed for each edge type are concatenated into $H_{\text{combined},u} \in \mathbb{R}^{3Md_{\textcolor{black}{\text{graph}}}}$ as follows:
\begin{equation*}
\label{combined}
H_{\text{combined},u} = \text{Concat}(H^{(1)}_{\text{edge},u}, H^{(2)}_{\text{edge},u}, H^{(3)}_{\text{edge},u}) 
\end{equation*}
Note that $H^{(e)}_{\text{edge}}$ in Fig.~\ref{vge} denotes the collection of $H^{(e)}_{\text{edge},u}$ for all $u$. We then employ a linear transformation to project the concatenated features back to the latent dimension $d_\text{latent}$:
\begin{equation*}
\label{proj}
H_{\text{proj},u} =\operatorname{Linear}( H_{\text{combined},u} )
\end{equation*}
 where $H_{\text{proj},u} \in \mathbb{R}^{d_\text{latent}}$. The aggregation function, denoted as $\operatorname{Aggr}(\cdot)$\textcolor{black}{,} computes the node embedding $h_u$ by combining Approximate Personalized Propagation of Neural Predictions (APPNP), residual connections, and batch normalization:
\begin{equation*}
h_u=\operatorname{Aggr}(H_{\text{proj},u})
\end{equation*}
\textcolor{black}
{The aggregation function $\operatorname{Aggr}(\cdot)$ operates as follows:}
\textcolor{black}
{\begin{align*}
H_{\text{aggr},u} &= \operatorname{MLP}_{\text{aggr1}}(H_{\text{proj},u}) \\
H_{\text{APPNP},u} &= \operatorname{BN}\!\left((1-\alpha_{\text{APPNP}})\, H_{\text{aggr},u} + \alpha_{\text{APPNP}}\, \mathbf{X}_u\right) \\
H_{\text{aggr2},u} &= \operatorname{MLP}_{\text{aggr2}}(H_{\text{APPNP},u}) \\
h_u &= \operatorname{BN}(H_{\text{aggr2},u} + H_{\text{APPNP},u})
\end{align*}}
where $\alpha_{\text{APPNP}} \in [0,1]$ is the teleport probability controlling the balance 
between the propagated representation and the original node feature $\mathbf{X}_u$, 
and $\operatorname{BN}(\cdot)$ denotes batch normalization. \textcolor{black}{The teleport term $\alpha_{\text{APPNP}}  \mathbf{X}_u$ explicitly reintroduces the original node features during aggregation, thereby reinforcing identity preservation. An ablation study on the APPNP is provided in Supplementary Material F.4.} The node embeddings $h\in \mathbb{R}^{N_O \times d_\text{latent}}$ are the collection of $h_u \in \mathbb{R}^{d_\text{latent}}$ for all $u$.

\subsubsection{Latent space model}
\label{Latent space model}

The latent space model implements a variational framework with a Gaussian prior distribution having mean $0$ and standard deviation $\sigma_p$:
\begin{equation*}
p(z) = N(0, \sigma_p^2 I)
\end{equation*}
 where $\sigma_p$ is $1$ and $I$ is identity matrix. The latent space model implements a variational framework where the posterior distribution $q_\phi(z|G)$ is modeled as a diagonal Gaussian with instance-dependent mean and variance. Our architecture employs a specialized network to jointly learn both parameters:
\begin{equation*}
q_\phi(z|G) = \mathcal{N}(\mu_z, \operatorname{diag}(\sigma_z^2))
\end{equation*}
\textcolor{black}{The element-wise mean node embeddings across all nodes, $h_{\text{mean}}$, are first processed by a shared feature extractor, followed by two separate heads.} Specifically, a shared MLP first maps the input to an intermediate latent representation, which is then split into two distinct feature vectors, {\color{black}$h_{\mu} ,h_{\sigma} \in \mathbb{R}^{d_{\text{latent}}}$}:
\begin{equation}
\label{latent1}
[h_{\mu}, h_{\sigma}] = \operatorname{MLP}_{\text{shared}}(\textcolor{black}{h_{\text{mean}}})
\end{equation}
 where $[\cdot, \cdot]$ denotes the splitting operation along the feature dimension. Subsequently, these features are passed through specific MLP heads to predict the mean and standard deviation:
\begin{align}
\label{latent2}
\mu_z &= \operatorname{MLP}_{\mu}(h_{\mu}) \\
\sigma_z &= \operatorname{Softplus}(\operatorname{MLP}_{\sigma}(h_{\sigma})) + \epsilon_{\text{min}}
\end{align}
where $\operatorname{Softplus}(x) = \log(1 + e^x)$ ensures positive standard deviation, 
and $\epsilon_{\text{min}}$ (set to $10^{-5}$) is a small constant for numerical stability, 
and {\color{black}$\mu_z, \sigma_z \in \mathbb{R}^{d_{\text{latent}}}$}. Finally, the latent variable $z$ is sampled using the reparameterization trick \textcolor{black}{to enable gradient-based optimization:}
\begin{equation}
\label{latent3}
z = \mu_z + \epsilon \odot \sigma_z, \quad \epsilon \sim \mathcal{N}(0, I)
\end{equation}
\textcolor{black}{where $\odot$ indicates the element-wise product.}

\subsubsection{Generative network}
\label{Generative network}
Generative component $p_\psi(G|z)$, parametrized with $\psi$\textcolor{black}{,} serves as a generative model for graph reconstruction. In practice, this is implemented via a deterministic function $g_\psi(z)$ that outputs node probabilities $\mathbf{P}_{\text{node}}$ and edge connection probabilities $\mathbf{P}_{\text{edge}}$, where $\mathbf{P}_{\text{node}} \in [0,1]^{\textcolor{black}{N_*} \times 6}$ and $\mathbf{P}_{\text{edge}} \in [0,1]^{\textcolor{black}{N_*} \times \textcolor{black}{N_*}\textcolor{black}{\times 3}}$\textcolor{black}{, with $6$ denoting the number of columns of $\mathbf{X}$, $3$ denoting the number of edge types, and $N_* = n^{*}m^{*}$ where $n^{*}$ and $m^{*}$ represent the maximum number of jobs and machines that can be encountered across problem instances}:
\begin{equation*}
G \sim p_\psi(G|z) \text{ via } \mathbf{P}_{\text{node}}, \mathbf{P}_{\text{edge}} = g_\psi(z)
\end{equation*}
 The generative network employs 1D transposed convolutions to generate graph structures from latent representations. The architecture consists of:
\begin{equation*}
\mathbf{z}^{(p)} = \begin{cases}
\text{Linear}(z) & \text{if } p = 0 \\
\text{ConvTranspose}_{d_{p-1} \rightarrow d_{p}}(\mathbf{z}^{(p-1)}) & \text{otherwise}
\end{cases}
\end{equation*}
For $p=0$, the linear transformation maps $z$ to an initial feature representation of dimension $d_0$ and reshapes it to $\mathbf{z}^{(0)} \in \mathbb{R}^{d_0 \times 1}$ with sequence length 1. Subsequently, the $p$-th transposed convolution operation $\text{ConvTranspose}_{d_{p-1} \rightarrow d_p}$ (with kernel size 4, stride 2, and padding 1) progressively reduces the channel dimension from $d_{p-1}$ to $d_p$ while doubling the sequence length at each layer. After $P$ transposed convolution layers, the sequence length exponentially grows to $2^P$. This hierarchical upsampling process generates increasingly detailed structural representations from the compressed latent space. The final output $\mathbf{z}_{\text{final}} \in \mathbb{R}^{d_P \times N_*}$ has channel dimension $d_P$. For node prediction, adaptive average pooling resizes $\mathbf{z}_{\text{final}}$ to match the desired node count before convolution:
\begin{align*}
\mathbf{P}_{\text{node}} &= \sigma(\text{Conv1D}(\text{AdaptiveAvgPool1D}(\mathbf{z}_{\text{final}}, N_*)))
\end{align*}
where $\sigma$ is the sigmoid activation function and $\mathbf{P}_{\text{node}} \in [0,1]^{\textcolor{black}{N_*} \times 6}$ represents the prediction of node features. For edge prediction, the sequence is interpolated to match the size $N_*^2$:
\begin{equation*}
\mathbf{P}_{\text{edge}} = \sigma(\text{Conv1D}(\text{Interpolate}(\mathbf{z}_{\text{final}}, \text{size}=N_*^2)))
\end{equation*}
where $\mathbf{P}_{\text{edge}} \in [0,1]^{\textcolor{black}{N_*} \times \textcolor{black}{N_*} \times 3}$ represents edge probabilities reshaped from the $\textcolor{black}{3}N_*^2$-length sequence.

\subsection{Policy decoder}
\label{Operation selecting decoder}
\textcolor{black}{The policy decoder constructively generates the schedule using $z$ and $h$ by interacting with the JSSP environment under the semi-active transition dynamics defined by $g_0$ and $g$.}
The action is selected from a policy function $\pi$, which is defined as follows:
\begin{equation*}
\pi_\theta (a|z, G) = \prod_{t=1}^{nm} p_\theta(a_t|z, s_t,a_{t-1})
\end{equation*}
where action sequence $a = (a_1, a_2, \ldots, a_{nm})$ consists of actions $a_t$ at each timestep $t$. The policy decoder representing the policy interacts with the JSSP environment iteratively by observing the state and selecting actions. The policy decoder inputs for selecting action $a_t$ at timestep $t$ consist of the context embedding ($\mathbf{c}_t$) and node-specific state embeddings ($\mathbf{k}_t^u$) as follows:
\begin{equation*}
\label{decoder input}
\mathbf{c}_t = \operatorname{Concat}(z, h_{a_{t-1}}), \quad \mathbf{k}_t^u = \operatorname{Concat}(h_{u},\operatorname{MLP}(s_t^u))
\end{equation*}
where $h_{a_{t-1}}$ represents the node embedding corresponding to the selected operation $a_{t-1}$ at the previous timestep $t-1$. To handle \textcolor{black}{the initial timestep}, where no previous action exists, the policy decoder uses \textcolor{black}{a} learnable parameter. Since $\operatorname{MLP}(s_t^u) \in \mathbb{R}^{d_{\text{latent}}}$, we have $\mathbf{c}_t, \mathbf{k}_t^u \in \mathbb{R}^{2d_{\text{latent}}}$. $\mathbf{c}_t$ and $\mathbf{k}_t^u$ are fed into a glimpse network, which employs multi-head attention to iteratively refine queries over the context (i.e., action history in our context) and state. \textcolor{black}{At each glimpse layer}, the query is \textcolor{black}{computed} as follows:
\begin{equation*}
\label{query}
\mathbf{q}_t^{l} = \sum_{m=1}^{M'}\operatorname{Attention}(\mathbf{q}_t^{l-1}W_{m,l}^Q, \mathbf{K}_tW_{m,l}^K, \mathbf{K}_tW_{m,l}^V)
\end{equation*}
where $\mathbf{q}_t^{0} = \mathbf{c}_t$ and $l$ denotes the $l$-th glimpse layer, $\mathbf{K}_t$ is the state embedding matrix with its $u$-th row being $\mathbf{k}_t^u$. $W_{m,l}^Q, W_{m,l}^K\in \mathbb{R}^{2d_{\text{latent}} \times (d_{\text{latent}}+d_{\text{glimpse}})}$, $W_{m,l}^V\in \mathbb{R}^{2d_{\text{latent}} \times 2d_{\text{latent}}}$ are learnable parameter matrix. $\operatorname{Attention}(Q, K, V)$ \textcolor{black}{is defined as the scaled dot-product attention with masking:}
\begin{equation*}
\operatorname{Attention}(Q, K, V) =\operatorname{Softmax}\left(  \frac{QK^\top}{\sqrt{d}}+\operatorname{mask}_t^1\right)V
\label{attention}
\end{equation*}
where $d=d_{\text{latent}}/\textcolor{black}{M'}$ and $\operatorname{mask}_t^1$ is a masking vector with size $nm$, whose $u$-th element indicates whether $O_u$ is scheduled (if already selected as an action at the previous timestep, a very small number, e.g., $-10^{8}$, is used, otherwise $0$). After $L$ iterations of glimpse refinement, the logit calculator \textcolor{black}{yields} a logit for each operation $O_u$ as follows:
\begin{equation*}
\label{pn}
\mathbf{z}_{u,t} = \operatorname{LC}(\mathbf{q}_t^{L} W_{\operatorname{LC}}^Q, \mathbf{k}_t^u W_{\operatorname{LC}}^K ) 
\end{equation*}
where $W_{\operatorname{LC}}^Q, W_{\operatorname{LC}}^K\in \mathbb{R}^{2d_{\text{latent}} \times (d_{\text{latent}}+d_{\text{logit}})}$ and logit calculating function $\operatorname{LC}$ with $\operatorname{tanh}$ clipping is defined as follows:
\begin{equation*}
\label{masking2}
\operatorname{LC}(q, k_u)
=\begin{cases}
      C \cdot \operatorname{tanh}(\frac{{q}^\top k_u}{d_{\text{latent}}}) & \text{if } O_u \in \mathcal{A}(s_t)\\
     -\infty & \text{otherwise}
   \end{cases}
\end{equation*}
Finally, the probability of selecting $O_u$ in $s_t$ is determined by the softmax function as follows:
\begin{equation*}
\label{softmax}
p_\theta(a_t=O_u|z, s_t,a_{t-1})=\frac{\operatorname{exp}{\textcolor{black}{\mathbf{z}_{u,t}}}}{\textcolor{black}{\sum_{u'=1}^{nm}} \operatorname{exp} {\textcolor{black}{\mathbf{z}_{u',t}}}}
\end{equation*}
where $\theta$ is the collection of parameters approximating $\pi$. During training, actions are sampled from this probability distribution to maintain exploration (stochastic policy), while during test \textcolor{black}{and validation}, the action with the highest probability is selected deterministically (greedy policy).
\textcolor{black}{\subsection{Design of the optimization objective}}
For the practical implementation of the first component of the reconstruction loss in \eqref{last elbo}, $\mathbb{E}_{z,a \sim q}[\log p(G|z)]$, we approximate the likelihood $p(G|z)$ as $p(G|\tilde{G})$, where $\tilde{G}$ represents the reconstructed instance generated by the decoder. 
This objective is then decomposed into two reconstruction losses for nodes and edges, denoted as $\mathcal{L}_{\text{node}} + \mathcal{L}_{\text{edge}}$. 
By assuming a Bernoulli distribution for the generation process, these terms are formulated as the Binary Cross-Entropy \textcolor{black}{(BCE)} loss:
\begin{align}
\mathcal{L}_{\text{edge}} &= -\frac{1}{\textcolor{black}{(nm)^2}} \sum_{e=1}^{3} \sum_{i=1}^{\textcolor{black}{nm}} \sum_{j=1}^{\textcolor{black}{nm}} \Big[ \mathbf{A}_{i,j}^{(e)} \log \mathbf{P}_{\text{edge},(e)}^{(i,j)}  + (1-\mathbf{A}_{i,j}^{(e)}) \log (1-\mathbf{P}_{\text{edge},(e)}^{(i,j)}) \Big] \nonumber
\end{align}
\begin{equation*}
\mathcal{L}_{\text{node}} = -\frac{1}{\textcolor{black}{nm}} \sum_{f=1}^{6} \sum_{i=1}^{\textcolor{black}{nm}} \left[ \mathbf{X}_{i,f} \log \mathbf{P}_{\text{node}}^{(i,f)} + (1-\mathbf{X}_{i,f}) \log (1-\mathbf{P}_{\text{node}}^{(i,f)}) \right]
\end{equation*}

\textcolor{black}{The standard BCE loss is originally formulated for discrete binary data. However, our node features are continuous and normalized, which, strictly speaking, does not satisfy this assumption. Nevertheless, applying BCE to such features is a widely accepted heuristic in the VAE literature \cite{kingma2013auto, larsen2016autoencoding, sonderby2016ladder}. We validated this choice through a preliminary comparison against Mean Squared Error (MSE) and the Continuous Bernoulli (CB) distribution, confirming BCE as the most effective option in terms of training stability and scalability. The experimental results are provided in Supplementary Material \textcolor{black}{F.5.}} The second loss component of the reconstruction loss, $\mathcal{L}_{\text{KL}}$, is defined as the KL divergence between the posterior $q_\phi(z|G)$ and the prior $p(z)$. 
To ensure computational tractability, we assume multivariate Gaussian distributions with diagonal covariance matrices, where the prior is centered at zero. 
Under these assumptions, the term can be computed analytically in closed form:
\begin{align*}
\mathcal{L}_{\text{KL}} &= D_{\text{KL}}[q_\phi(z|G) \| p(z)] \nonumber \\
&= \frac{1}{2} \sum_{i=1}^{d} \left[ 
\frac{\textcolor{black}{\sigma_{z,i}^{2}}}{\sigma_{p}^{2}} 
+ \frac{\textcolor{black}{\mu_{z,i}^{2}}}{\sigma_{p}^{2}} 
- 1 
- \log \frac{\textcolor{black}{\sigma_{z,i}^{2}}}{\sigma_{p}^{2}} 
\right] \nonumber 
\end{align*}
{\color{black}Consequently, the reconstruction loss $\mathcal{L}_{\text{repr}}$ is defined as:
\begin{equation*}
    \mathcal{L}_{\text{repr}} = \beta \cdot \mathcal{L}_{\text{KL}} + \mathcal{L}_{\text{node}} + \mathcal{L}_{\text{edge}}
\end{equation*}
where $\beta$ controls the regularization strength of the KL divergence term. A sensitivity analysis on $\beta$ is provided in Supplementary Material J.2.}
\par
The policy loss term $\mathcal{L}_{\text{policy}}$ is derived from \eqref{PG with maximum entropy}. 
Since the JSSP operates in a discrete action space, we can optimize the objective directly without relying on the reparameterization trick, unlike continuous control methods such as Soft Actor-Critic \cite{haarnoja2018soft}. 
By applying the policy gradient theorem, we reformulate the objective into the following gradient form:
\begin{equation*}
\label{policy gradient}
\mathcal{L}_{\text{policy}}= \mathbb{E}_{G \sim p, z,a \sim q}[\log \pi_\theta(a|z, G) A(z,a) - \alpha_{\text{entropy}} \log \pi_\theta(a|z, G)]\nonumber 
\end{equation*}
Here, $A(z,a) = Q(z,a) - V_\gamma(z)$ denotes the advantage function. 
$V_\gamma(z)$ represents the soft state value function estimating the expected return from the latent state $z$, while the Q-value is defined as the negative makespan, $Q(z,a) = -C_{\max}(z,a)$. 
The critic loss function for policy evaluation is defined as follows:
\begin{equation*}
    \mathcal{L}_{\text{critic}} = \mathbb{E}_{G \sim p, \, z,a \sim q} \left[ \left( V_\gamma(z) - \left( Q(z,a) - \alpha_{\text{entropy}} \log \pi_\theta(a|z, G) \right) \right)^2 \right]
\end{equation*}
This optimization process provides theoretical guarantees equivalent to those of Soft Policy Iteration (SPI) \cite{haarnoja2018soft}. Derivations are provided in Supplementary Material C.
\par
\begin{algorithm}[H]
\caption{Two-phase training algorithm}
\label{alg:two_phase_training}
\begin{algorithmic}[1]
\REQUIRE Initial parameters $\phi$, $\psi$, $\theta$,  $\gamma$ 
\REQUIRE Hyperparameters: representation epochs $E_r$, policy epochs $E_p$,\\
\textcolor{black}{KL divergence weight $\beta$},
learning rates $\eta_1, \eta_2, \eta_3$, mini-batch size $B$

\STATE \textbf{Phase 1: Variational Representation Learning}
\FOR{$\text{epoch} = 1$ to $E_r$}
    \STATE Sample random JSSP instance $G \sim p(G)$
    \STATE $z \sim q_\phi(z|G)$ \COMMENT{Sampling latent variable}
    \STATE $\mathbf{P}_{\text{node}}, \mathbf{P}_{\text{edge}} = g_\psi(z)$ \COMMENT{Reconstructing instance}
    \STATE Compute $\mathcal{L}_{\text{KL}}$, $\mathcal{L}_{\text{node}}$ and $\mathcal{L}_{\text{edge}}$
    \STATE {\color{black}$\mathcal{L}_{\text{repr}} \leftarrow \beta \cdot \mathcal{L}_{\text{KL}} + \mathcal{L}_{\text{node}} + \mathcal{L}_{\text{edge}}$}
    \STATE \textcolor{black}{$\phi, \psi \leftarrow \phi - \eta_1 \nabla_\phi \mathcal{L}_{\text{repr}}, \, \psi - \eta_1 \nabla_\psi \mathcal{L}_{\text{repr}}$}
\ENDFOR
\STATE \textbf{Phase 2: Policy Learning}
\STATE Fix $\phi, \textcolor{black}{\psi}$ from Phase 1
\FOR{$\text{epoch} = 1$ \TO $E_p$}
    \STATE Initialize $\mathcal{L}_{\text{policy}} = 0$ and $\mathcal{L}_{\text{critic}} = 0$
    \FOR{$\text{batch} = 1$ \TO $B$}
        \STATE Sample random JSSP instance $G \sim p(G)$
        \STATE $z \sim q_\phi(z|G)$ \COMMENT{Sampling latent variable}
        \FOR{$t = 1$ \TO $T-1$}
            \IF{$t = 1$}
                \STATE $s_1 \gets g_0(G)$ \COMMENT{Initialize state}
            \ENDIF
            \STATE $a_t \sim p_\theta(a_t|z, s_t, a_{t-1})$ \COMMENT{Sample action}
            \STATE $s_{t+1} \gets \textcolor{black}{g}(s_t, a_t, G)$ \COMMENT{Update state}
        \ENDFOR
        \STATE $a = (a_1, a_2, \ldots, a_{T-1})$ \COMMENT{Construct action sequence}
        \STATE $Q(z,a) \leftarrow -C_{\max}(z,a)$ \COMMENT{Negative makespan}
        \STATE $V_\gamma(z) \leftarrow$ critic network output
        \STATE $A(z,a) \leftarrow Q(z,a) - V_\gamma(z)$ \COMMENT{Advantage}
        \STATE Compute \textcolor{black}{batch} losses $\mathcal{L}_{\text{policy}}^{\textcolor{black}{\text{mini-batch}}}$, $\mathcal{L}_{\text{critic}}^{\textcolor{black}{\text{mini-batch}}}$
        \STATE $\mathcal{L}_{\text{policy}} \leftarrow \mathcal{L}_{\text{policy}}+\frac{1}{B} \mathcal{L}_{\text{policy}}^{\textcolor{black}{\text{mini-batch}}}$
        \STATE $\mathcal{L}_{\text{critic}} \leftarrow \mathcal{L}_{\text{critic}}+\frac{1}{B} \mathcal{L}_{\text{critic}}^{\textcolor{black}{\text{mini-batch}}}$
    \ENDFOR
    \STATE $\theta \leftarrow \theta + \eta_2 \nabla_\theta \mathcal{L}_{\text{policy}}$ \COMMENT{Update policy}
    \STATE $\gamma \leftarrow \gamma - \eta_3 \nabla_\gamma \mathcal{L}_{\text{critic}}$ \COMMENT{Update critic}
\ENDFOR
\end{algorithmic}
\end{algorithm}
\subsection{Two-phase training strategy}
We adopt a two-phase training strategy, as outlined in Algorithm 1, to operationalize the architecture and objective functions detailed in Sections \ref{ELBO with maximum entropy RL for JSSP} and \ref{VG2S framework for jssp}. \textcolor{black}{While the ELBO with maximum entropy RL objective in Section~\ref{ELBO with maximum entropy RL for JSSP} inherently assumes joint optimization of the variational graph encoder and policy decoder, the two-phase strategy is an empirically motivated design choice, as demonstrated in Section~\ref{Ablation on variational representation learning}: joint optimization yields inferior performance and training stability compared to the proposed two-phase approach. Furthermore, this separation can prevent overfitting to the source domain typically induced by standard end-to-end learning, thereby achieving robust zero-shot domain adaptation~\cite{higgins2017darla}.}
\par
Phase 1 trains the variational graph encoder. For each epoch, a JSSP instance is sampled (line 3), followed by sampling of its corresponding latent variable (line 4). The reconstruction loss $\mathcal{L}_{\text{repr}}$ is computed (lines 5-7) and used to update the variational graph encoder parameters (line 8). Upon completion of Phase 1 (\textcolor{black}{variational} representation learning), the variational graph encoder parameters are frozen (line 11) to proceed to Phase 2. Phase 2 trains the policy decoder using mini-batch learning. A JSSP instance is randomly sampled for training (line 15), and the latent variable for that instance is sampled accordingly (line 16). The system then performs rollouts to construct action sequences (lines 17-20) by interacting with the JSSP environment. Both $\mathcal{L}_{\text{policy}}$ and $\mathcal{L}_{\text{critic}}$ are computed and used to update the policy and critic parameters\textcolor{black}{,} respectively. \textcolor{black}{A flowchart is provided in Supplementary Material D.}

\section{Experiment}
\label{sec:experiment}

This section provides a multifaceted validation of the proposed VG2S framework. In Section~\ref{Test}, we demonstrate the zero-shot generalization and scalability of the proposed model using various benchmark datasets, conduct a statistical analysis under controlled experimental settings, \textcolor{black}{and compare VG2S against representative methods from fundamentally different learning paradigms.} Section~\ref{Effectiveness of variational representation learning} presents an ablation study to analyze the impact of the \textcolor{black}{variational approach} on policy learning and training stability, \textcolor{black}{examining} its robustness across different problem scales. In Section~\ref{Visual analysis of latent representation via UMAP}, we utilize a \textcolor{black}{dimensionality reduction technique} to visualize the evolution of the latent space during training, illustrating how the model captures the structural features of JSSP instances. \textcolor{black}{Section~\ref{Analysis of agent strategy via PDR similarity} provides an in-depth analysis of the learned policy behavior, interpreted through the lens of traditional heuristics.} Section~\ref{Computational complexity} analyzes the computational complexity and efficiency of the proposed framework. \textcolor{black}{Additional analyses—including implementation details and detailed results of the main experiments, preliminary studies on architectural design choices, sensitivity analyses, and semantic analysis of latent dimensions—are provided in the Supplementary Material.} Implementation details for all experiments are provided in Supplementary Material E.
\par
For each experiment, the proposed VG2S is trained using a dynamic data generation strategy rather than a fixed training set. Specifically, training instances are regenerated every five epochs according to the following configurations:
\begin{itemize}
    \item \textbf{Processing time}: $p_{ij} \sim \mathcal{DU}(1, 99)$ for each operation.
    \item \textbf{Number of machines}: $m \sim \mathcal{DU}(5, 9)$.
    \item \textbf{Number of jobs}: $n \sim \mathcal{DU}(m, 9)$.
    \item \textbf{Machine sequence}: A random permutation for each job, ensuring no recirculation.
\end{itemize}
where $\mathcal{DU}(a, b)$ denotes the discrete uniform distribution between $a$ and $b$. 
\textcolor{black}{All experiments are conducted on a system equipped with an NVIDIA GeForce RTX 3070 Ti GPU, 
a 12th Gen Intel Core i7-12700 processor (2.10 GHz), and 64 GB of RAM.}

\subsection{Test}
\label{Test}
To evaluate the robustness of VG2S, we assess its performance across a wide range of standard JSSP benchmark instances. \textcolor{black}{The key hyperparameters are tuned over six predefined configurations, each evaluated across five representation learning epochs ($E_r \in \{20{,}000, 40{,}000, 60{,}000, 80{,}000, 100{,}000\}$), resulting in a total of 30 hyperparameter combinations. As a result, Sections~\ref{Scalability test}, \ref{Generalization and scalability test} and \textcolor{black}{\ref{sec:paradigm_comparison}} use the model at epoch 6,001 with hyperparameter ID 4 ($E_r = 80,000$), selected based on average validation performance on TA61, TA62, DMU76, and DMU77 dataset; detailed hyperparameter configurations are provided in Supplementary Material E. Section~\ref{Statistical analysis under controlled settings} presents additional controlled experiments.} We evaluate the performance using the optimality gap metric, calculated as the percentage deviation from the best-known solution: $100 \% \times \frac{C^{\text{method}}_{\max}- C^{\text{UB}}_{\max}}{C^{\text{UB}}_{\max}}$, where $C^{\text{method}}_{\max}$ and $C^{\text{UB}}_{\max}$ are the makespan of the each methodology and the best-known, respectively.
\subsubsection{Scalability test}
\label{Scalability test}
The proposed method is evaluated on the Taillard (TA) benchmark dataset, a standard benchmark for classical JSSP. The TA01-80 instances were originally proposed in \cite{taillard1993benchmarks}. With processing time and machine sequence distributions similar to our training dataset but larger problem sizes, this dataset serves as an effective benchmark for evaluating scalability\footnote{The dataset characteristics are as follows: the number of machines $m$ and jobs $n$ are drawn from $DU[15, 20]$ and $DU[15, 100]$, respectively, while processing times are sampled from $U[1, 99]$. The machine operation sequences are generated through the following procedure: (1) operations are initially assigned to machines sequentially without perturbation, where the first machine processes the first operation; (2) starting from the first operation, the assigned machine is swapped with a randomly selected machine from the subsequent operations, and this procedure is repeated until the last operation.}.
\textcolor{black}{For comparison, we evaluate four priority dispatching rules (PDRs) commonly applied in the JSSP domain: Shortest Processing Time (SPT), Most Work Remaining (MWKR), Most Operations Remaining (MOR), and First In First Out (FIFO).} To assess representation learning effectiveness, we select \textcolor{black}{GNN-}DRL approaches sharing a similar graph representation foundation as baselines for the proposed VG2S framework: Zhang [A] \cite{zhang2020learning}, Park \textcolor{black}{[A]} \cite{park2021learning}, Yuan \cite{yuan2023solving}, Liu \cite{Dynamic_JSSP}, and Zhang [B] \cite{SAC_JSSP}. \textcolor{black}{These methods are not re-implemented under identical conditions. Optimality gaps are computed from the makespan values reported in the respective original papers using a single unified upper bound.}
\par
Table~\ref{tab:jssp_comparison} presents the performance comparison on the TA benchmark dataset. The proposed VG2S demonstrates superior performance across the TA benchmark instances, achieving the best results in 12 out of 16 test cases (75\%). Compared to PDRs, the proposed approach consistently outperforms across all datasets except TA32. Furthermore, it surpasses GNN-DRL baselines in most cases. Notably, a distinct pattern emerges when examining performance across different problem sizes. On smaller instances ($30 \times 15$ or below), the method achieves best performance on 5 out of 8 datasets, whereas on larger instances ($30 \times 20$ or above), it achieves best performance on 7 out of 8 datasets. This trend demonstrates that the method exhibits robust scalability, maintaining and even improving its competitive advantage as problem complexity increases.

\begin{table}
    \centering
    \caption{Performance comparison of VG2S and baseline methods on the Taillard benchmark instances}
    \label{tab:jssp_comparison}
    \renewcommand{\arraystretch}{1.2}
    \begin{tabular}{@{}cc|cccccccccc@{}}
    \toprule
        \textbf{Instance} & \textbf{Size} & \textcolor{black} {\textbf{SPT}} & \textcolor{black} {\textbf{MWKR}}& \textcolor{black} {\textbf{MOR}}&\textcolor{black} {\textbf{FIFO}}&  \makecell{\textbf{Zhang} \\
\textbf{[A]}\cite{zhang2020learning}}&  \makecell{\textbf{Park} \\ \textbf{[A]} \cite{park2021learning}}&  \makecell{\textbf{Yuan} \\ \cite{yuan2023solving}}&  \makecell{\textbf{Liu} \\ \cite{Dynamic_JSSP}}& \makecell{\textbf{Zhang} \\ \textbf{ [B]}\cite{SAC_JSSP}} & \textbf{VG2S}\\
        \midrule
        TA01 & $15\times15$ & 22.10& 23.56& 24.05&38.99
& 17.22& \textbf{12.84}& 18.36& 21.20& 16.57& 15.52
\\ 
        TA02 & $15\times15$ & 30.87& 26.69& 12.62&18.49
& 24.12& 22.11& 17.28& 14.55& 16.24& \textbf{11.09}\\ 
        \midrule
        TA11 & $20\times15$ & 37.73& 27.27& 34.93&35.81
& 32.20& 19.82& 22.33& 29.11& 25.35& \textbf{13.26}\\
        TA12 & $20\times15$ & 24.14& 27.21& 31.82&34.02
& 32.04& 22.02& 20.41& 23.77& 27.72& \textbf{15.29}\\
        \midrule
        TA21 & $20\times20$ & 30.45& 25.70& 22.53&35.38
& 37.15& 33.92& 20.83& 27.71& 29.96& \textbf{12.61}\\ 
        TA22 & $20\times20$ & 30.44& 23.88& 25.31&27.50
& 31.38& 28.06& \textbf{14.25}& 20.25& 23.13& 15.00\\ 
        \midrule
        TA31 & $30\times15$ & 41.72& 28.80& 22.00&32.88
& 45.41& 27.61& 26.13& 29.08& 24.55& \textbf{21.43}\\ 
        TA32 & $30\times15$ & 35.03& 23.93& 25.95&37.95
& 33.86& 33.30& 23.54& \textbf{23.49}& 33.52& 25.78
\\ 
        \midrule
        TA41 & $30\times20$ & 34.81& 30.32& 30.82&31.92
& 33.02& 32.37& 25.34& 34.56& 34.06& \textbf{22.89}\\ 
        TA42 & $30\times20$ & 52.97& 23.44& 34.38&39.39
& 37.53& 33.14& 21.89& 35.42& 32.78& \textbf{17.04}\\ 
        \midrule
        TA51 & $50\times15$ & 20.07& 28.26& 31.05&33.12
& 30.40& \textbf{13.95}& 25.91& 30.72& 27.79& 19.86
\\
        TA52 & $50\times15$ & 16.55& 23.66& 21.92&30.15
& 21.23& 14.55& 23.77& 27.87& 25.73& \textbf{12.37}\\
        \midrule
        TA61 & $50\times20$ & 41.74& 19.00& 21.34&29.36
& 27.41& 19.42& 17.26& 23.71& 24.48& \textbf{10.91}\\
        TA62 & $50\times20$ & 38.79& 20.18& 23.74&26.39
& 26.07& 26.39& 23.70& 23.98& 26.49& \textbf{14.88}\\
        \midrule
        TA71 & $100\times20$ & 20.64& 12.17& 9.33&12.98
& 18.08& 9.11& 11.62& 15.10& 12.32& \textbf{7.08}\\ 
        TA72 & $100\times20$ & 21.56& 8.22& 10.54&12.28
& 9.92& 6.58& 10.23& 15.85& 9.05& \textbf{5.66}\\ 
         \bottomrule
    \end{tabular}
\end{table}

\subsubsection{Generalization and scalability test}
\label{Generalization and scalability test}
To extend the evaluation scope beyond the TA dataset, which shares similar processing time and machine sequence distributions with our training data but with larger problem sizes, we additionally tested our method on widely-used benchmark datasets: DMU \cite{demirkol1998benchmarks}, SWV \cite{storer1992new}, LA \cite{lawrance1984resource}, ORB \cite{applegate1991computational}, ABZ \cite{adams1988shifting}, FT \cite{fisher1963probabilistic}, and YN \cite{Yamada1992AGA}. To assess representation learning effectiveness, we select GNN-DRL approaches sharing a similar graph representation foundation as baselines for the proposed VG2S framework: Chen \cite{chen2022deep}, Zhang \textcolor{black}{[A]} \cite{zhang2020learning}, Park [A] \cite{park2021learning}, Park [B] \cite{park2021schedulenet}, Yuan \cite{yuan2023solving}, Oh \cite{oh2025framework}, and \textcolor{black}{Zhang [B] \cite{SAC_JSSP}}. \textcolor{black}{These methods are not re-implemented under identical conditions. Optimality gaps are computed from the makespan values reported in the respective original papers using a single unified upper bound.}
\par
Table~\ref{tab:jssp_comparison2} presents a comprehensive performance comparison of VG2S against the baselines, covering a total of 242 benchmark datasets grouped by instance and size. Individual test results are provided in the Supplementary Material G. The proposed VG2S achieves the best performance on 52.9\% (18 out of 34) of the total benchmark dataset groups. We analyze the results from two perspectives: generalization and scalability.
\par
First, to validate the generalization capability, we focus on the performance on the DMU and SWV datasets, which \textcolor{black}{include} 2-set JSSP\footnote{The 2-set JSSP instances are defined by partitioning the set of machines into two distinct groups. All jobs are required to visit every machine in the first set in a random order before they can proceed to any machine in the second set. This specific routing structure has been found to be more difficult to solve than standard problems where the routing are a simple random permutation of all machines.} instances with distributions different from the training dataset, particularly in terms of machine sequence ordering. Notably, many instances in the DMU and SWV datasets remain unsolved optimally \cite{xie2022hybrid} and are widely recognized as particularly challenging problems \cite{demirkol1998benchmarks, storer1992new}. Despite this difficulty, the proposed method demonstrates superior performance on 90.9\% (10 out of 11) of the DMU and SWV groups. These results strongly suggest that the proposed methodology exhibits excellent generalization capability, even when applied to problem distributions that differ significantly from the training data.
\par
\textcolor{black}{To more comprehensively assess scalability beyond the TA benchmark analyzed in Section~\ref{Scalability test}, we analyze performance across the 242 benchmark datasets based on problem size by dividing instances into two categories:} those with fewer than 40 jobs and those with 40 or more jobs. For small-scale problems (fewer than 40 jobs), VG2S achieves the best performance in \textcolor{black}{11} out of 26 instance groups (\textcolor{black}{42.3}\%). \textcolor{black}{VG2S maintains a moderate advantage over the baselines across this scale. For large-scale problems \textcolor{black}{ (40 or more jobs)}, VG2S demonstrates superior performance in 7 of the 8 instance groups (87.5\%), confirming its robust scalability.} The results show that VG2S is effective for challenging and large-scale JSSP tasks, where standard DRL models often show limited performance.
\textcolor{black}{
Meanwhile, as baseline results are taken directly from the respective original papers under non-identical experimental conditions, these comparisons are intended to situate the generalization and scalability of VG2S relative to existing GNN-DRL methods rather than to claim definitive performance superiority. A controlled statistical comparison under identical conditions is provided in Section~\ref{Statistical analysis under controlled settings}.}

\begin{table}
    \centering
    \caption{Comprehensive performance evaluation of VG2S across diverse JSSP benchmarks: Generalization results on multiple instances of varying scales. Blank cells indicate benchmarks not reported in the respective original papers.}
    \label{tab:jssp_comparison2}
    \renewcommand{\arraystretch}{1.2}
    \begin{tabular}{@{}cc|cccccccccc@{}}
    \toprule
        \textbf{Instance} & \textbf{Size} &\textbf{FIFO}&\textbf{MWKR}
& \makecell{\textbf{Chen} \\\cite{chen2022deep}}&\makecell{\textbf{Zhang}  \\ \textbf{[A]}\cite{zhang2020learning}}& \makecell{\textbf{Park} \\ \textbf{[A]}\cite{park2021learning}}  & \makecell{\textbf{Park} \\ \textbf{[B]}\cite{park2021schedulenet}}& \makecell{\textbf{Yuan} \\ \cite{yuan2023solving}} & \makecell{\textbf{Oh} \\ \cite{oh2025framework}}&  \makecell{\textbf{Zhang} \\ \textbf{[B]}\cite{SAC_JSSP}}&\textbf{VG2S}\\
        \midrule
        TA& $15\times15$&30.72&22.58
& 35.38&25.96& 20.13&15.30& 21.32& \textbf{14.70}&  &15.66
\\
 TA& $20\times15$&35.06&25.08
& 32.09&30.03& 24.95&19.43& 22.39& 16.58& &\textbf{15.37}\\
 TA& $20\times20$&32.22&22.50
& 28.33&31.61& 29.25&17.25& 20.92& 16.69& &\textbf{16.30}\\
 TA& $30\times15$&31.43&28.35
& 36.43&33.00& 24.70&19.09& 23.28& \textbf{19.07}& &19.32
\\
 TA& $30\times20$&36.76&27.45
& 34.67&33.62& 32.00&23.75& 26.33& \textbf{19.69}& &20.62
\\
 TA& $50\times15$&22.95&18.47
& 31.86&20.86& 15.92&13.86& 16.03& 13.35& &\textbf{11.76}\\
 TA& $50\times20$&24.88&19.99
& 28.04&23.14& 21.30&13.53& 17.61& 13.23& &\textbf{13.01}\\
 TA& $100\times20$&13.82&9.53
& 17.98&13.52& 9.24&\textbf{6.66}& 8.91& 7.17& &6.95
\\
\midrule
 DMU & $20\times15$&40.44&30.26
& &38.95& 
 &
& 27.74& 23.14& 33.52
&\textbf{19.73}\\ 
        DMU & $20\times20$&36.25&29.51
& &37.74& 
 &
& 23.56& \textbf{18.72}&  28.37
&20.31
\\ 
        DMU & $30\times15$ &42.65&36.07
& &41.86& 
 &
& 28.70& 28.04&  38.24
&\textbf{22.75}\\ 
        DMU & $30\times20$ &38.72&36.23
& &39.48& 
 &
& 29.21& 26.55&  35.64
&\textbf{23.44}\\ 
        DMU & $40\times15$ &38.58&28.90
& &34.50& 
 &
& 25.32& 26.83&  33.79
&\textbf{20.96}\\ 
        DMU & $40\times20$ &42.20&35.49
& &39.00& 
 &
& 32.47& 28.28&  36.45
&\textbf{26.10}\\ 
        DMU & $50\times15$ &36.97&28.71
& &36.20& 
 &
& 24.96& 25.75&  32.67
&\textbf{17.52}\\ 
        DMU & $50\times20$ &41.38&33.73
& &38.40& 
 &
& 33.42& 29.89&  36.85
&\textbf{26.88}\\ 
        \midrule
        SWV & $20\times10$ &44.43&36.88
& && 28.42&34.39& 31.42& 28.28&  &\textbf{21.03}\\
        SWV & $20\times15$ &46.43&35.32
& && 29.39&30.51& 30.89& 29.94&  &\textbf{26.43}\\
        SWV & $50\times10$ &31.79&24.54
& && 16.80&25.33& 14.10& 20.68&  &\textbf{12.42}\\
        \midrule
        LA & $10\times5$ &21.15&16.63
& && 16.06&12.12& 14.84& \textbf{9.93}&  &10.24
\\
        LA & $15\times5$ &9.60&6.93
& && 1.09&2.65& 7.52& \textbf{1.02}&  &3.32
\\
        LA & $20\times5$ &10.05&5.61
& && 2.13&3.64& 4.78& 2.97&  &\textbf{1.80}\\
        LA & $10\times10$ &31.08&14.69
& && 17.06&11.95& 11.65& 9.99&  &\textbf{8.54}\\
        LA & $15\times10$ &30.37&19.58
& && 21.97&14.60& 13.01& \textbf{11.19}&  &16.02
\\
        LA & $20\times10$ &32.27&21.36
& && 27.26&15.72& 17.02& \textbf{13.87}&  &18.00
\\
        LA & $30\times10$ &13.37&9.77
& && 6.27&\textbf{3.10}& 7.92& 3.61&  &4.21
\\
        LA & $15\times15$ &28.95&19.66
& && 21.40&16.07& 16.22& \textbf{10.94}&  &14.94
\\
        \midrule
        ORB & $10\times10$ &33.13&25.85
& && 21.83&19.98& 23.08& \textbf{17.77}&  &18.04
\\
        \midrule
        ABZ & $10\times10$ &12.89&15.49
& && 10.12&\textbf{6.15}& 7.07& 6.92&  &9.14
\\
        ABZ & $20\times15$ &35.43&24.60
& && 29.02&20.55& 22.29& 19.48&  &\textbf{18.90}\\
        \midrule
        FT & $6\times6$ &9.09&9.09
& && 29.09&7.27& 9.09& \textbf{1.82}&  &7.27
\\
        FT & $10\times10$ &25.27&16.88
& && 22.80&19.46& 18.49& \textbf{11.61}&  &13.98
\\
        FT & $20\times5$ &42.66&35.54
& && 14.85&28.58& 14.33& \textbf{8.15}&  &11.76
\\
         \midrule
         YN & $20\times20$&27.24&25.71
& && 24.80&18.44& 20.28& 17.29&  &\textbf{15.81}\\ \hline
    \end{tabular}
\end{table}
\clearpage
\subsubsection{\textcolor{black}{Statistical analysis under controlled settings}}
\label{Statistical analysis under controlled settings}
\textcolor{black}{To provide a more rigorous and fair assessment of VG2S, we conduct an experiment in which all models are trained from scratch under controlled conditions (full details in Supplementary Material H). Given the strictly controlled nature of this experiment, baselines are selected from well-established GNN-DRL approaches with publicly available source code (Zhang [A] \cite{zhang2020learning} and Yuan \cite{yuan2023solving}) to ensure exact replication. In particular, unlike many state-of-the-art DRL methods \cite{SAC_JSSP, echeverria2025offline,zhang2024deep, iklassov2023study, huang2023novel, distributed_JSSP, ho2024residual, JSSP_uncertain_processing_time} that are evaluated solely on classical benchmarks, both baselines have been rigorously evaluated on both classical (TA) and challenging (DMU) benchmarks, making them consistent with our evaluation scope. All models, including the proposed VG2S, are trained from scratch under identical computing hardware and training data distributions. To further ensure fairness in hyperparameter tuning, a budget of 12 hyperparameter configurations—twice that allocated to VG2S—is assigned to each baseline. For all methods, the checkpoint is selected based on the best average makespan across four validation instances (TA61, TA62, DMU76, and DMU77), all of size $50 \times 20$, with validation performed every 200 training epochs, and all test results are obtained from this model.}
\par
\textcolor{black}{We focus on Yuan's model as it consistently shows the most competitive performance among the baselines, and thus serves as the primary subject of the two-tailed paired t-test analysis.} \textcolor{black}{Detailed comparison results for Zhang [A], 
Yuan, and VG2S are provided in Supplementary Material H.} \textcolor{black}{We evaluate Yuan at two training states, each corresponding to the best-performing checkpoint selected from the 12 hyperparameter configurations: Yuan (early best), the best-performing checkpoint in the early stage of training, and Yuan (converging best), the best-performing checkpoint in the converged stage\footnote{\textcolor{black}{Two checkpoints of Yuan are selected to provide the most favorable evaluation conditions for the baseline. Yuan's novel IAM mechanism functions as a powerful rule-based heuristic for narrowing the action space, which yields strong initial solutions as evidenced by both the original paper~\cite{yuan2023solving} and our experimental results. Accordingly, both the early-stage and converged checkpoints are considered: Yuan (early best), the best-performing checkpoint within the first 600 epochs, and Yuan (\textcolor{black}{converging best}), the best-performing checkpoint thereafter. Detailed training behavior of Yuan is discussed in Supplementary Material H.}}.}
For VG2S, results for each $E_r \in \{20{,}000, 40{,}000, 60{,}000, 80{,}000, 100{,}000\}$ reflect the best checkpoint across six hyperparameter configurations. Evaluating across multiple $E_r$ settings reveals how the length of the representation learning phase affects relative performance.
\par
\textcolor{black}{As detailed in Table~\ref{t-test}, we conduct a two-tailed paired t-test across three scopes: the overall dataset (all 242 benchmark instances), the DMU and SWV datasets (to evaluate zero-shot generalization), and large-scale instances with $n \times m \ge 500$ (to assess scalability). Across all scopes, VG2S consistently achieves lower optimality gaps compared to both Yuan (\textcolor{black}{early best}) and Yuan (\textcolor{black}{converging best}).} 
\par
\textcolor{black}{Across the overall dataset, VG2S achieves significant superiority over both states of Yuan's model regardless of the $E_r$ settings 
(all p-values at most $0.002$).} Furthermore, the results validate the zero-shot generalization of VG2S on the out-of-distribution (OOD) DMU and SWV datasets. Compared against the early best of Yuan's model, with the \textcolor{black}{exception of $E_r=20{,}000$ ($p = 0.013$) and 
$E_r=100{,}000$ ($p = 0.010$),} all other setups yield significant results ($p < 0.001$). When compared against Yuan (\textcolor{black}{converging best}), VG2S shows significant superiority ($p < 0.001$) across all $E_r$ settings. Finally, for large-scale problems ($n \times m \ge 500$), the performance 
superiority of VG2S remains statistically significant regardless of the 
$E_r$ settings \textcolor{black}{(all p-values at most $0.012$)}. \textcolor{black}{These statistical results confirm that the proposed VG2S framework significantly outperforms the baseline in terms of comprehensive performance, zero-shot generalization, and scalability. Evaluating across multiple $E_r$ settings further ensures fairness, 
confirming that the observed superiority of VG2S is not attributable 
to a specific choice of $E_r$ but holds consistently across the 
entire representation learning budget.}
\begin{table}[htbp]
\centering
\caption{\textcolor{black}{Two-tailed paired t-test results comparing the mean optimality gap between the baseline (Yuan) and the proposed VG2S across different dataset groups. Statistical significance levels are denoted as follows: $^{***}p < 0.001$.}}
\label{t-test}
\renewcommand{\arraystretch}{1.2}
\begin{tabular}{c | cc | cc | c}
\hline
\makecell{\textbf{Dataset} \\ \textbf{Group}} & \makecell{\textbf{Baseline} \\ \textbf{(Yuan)}} & \makecell{\textbf{Mean of} \\ \textbf{Optimality Gap}} & \makecell{\textbf{Target} \\ \textbf{(VG2S)}} & \makecell{\textbf{Mean of} \\ \textbf{Optimality Gap}} & \textbf{P-value} \\ 
\hline
\multirow{10}{*}{Overall} & \multirow{5}{*}{\textcolor{black}{Early best}}& \multirow{5}{*}{\textcolor{black}{21.88}}& $E_r = 20{,}000$ & 20.58
& \textcolor{black}{$0.002$} \\
& & & $E_r = 40{,}000$ & 18.92
& $^{***}$ \\
& & & $E_r = 60{,}000$ & 18.93
& $^{***}$ \\
& & & $E_r = 80{,}000$ & 16.80
& $^{***}$ \\
& & & $E_r = 100{,}000$ & 20.47
& $^{***}$ \\
\cline{2-6}
& \multirow{5}{*}{\textcolor{black}{Converging best}}& \multirow{5}{*}{\textcolor{black}{24.11}}& $E_r = 20{,}000$ & 20.58
& $^{***}$ \\
& & & $E_r = 40{,}000$ & 18.92
& $^{***}$ \\
& & & $E_r = 60{,}000$ & 18.93
& $^{***}$ \\
& & & $E_r = 80{,}000$ & 16.80
& $^{***}$ \\
& & & $E_r = 100{,}000$ & 20.47
& $^{***}$ \\
\hline
\multirow{10}{*}{\makecell{DMU and \\ SWV}} & \multirow{5}{*}{\textcolor{black}{Early best}}& \multirow{5}{*}{\textcolor{black}{27.74}} 
  & $E_r = 20{,}000$ & 26.20
& \textcolor{black}{0.013} \\
& & & $E_r = 40{,}000$ & 24.14
& $^{***}$ \\
& & & $E_r = 60{,}000$ & 24.33
& $^{***}$ \\
& & & $E_r = 80{,}000$ & 21.38
& $^{***}$ \\
& & & $E_r = 100{,}000$ & 26.08
& \textcolor{black}{0.010} \\
\cline{2-6}
& \multirow{5}{*}{\textcolor{black}{Converging best}}& \multirow{5}{*}{\textcolor{black}{32.23}}& $E_r = 20{,}000$ & 26.20
& $^{***}$ \\
& & & $E_r = 40{,}000$ & 24.14
& $^{***}$ \\
& & & $E_r = 60{,}000$ & 24.33
& $^{***}$ \\
& & & $E_r = 80{,}000$ & 21.38
& $^{***}$ \\
& & & $E_r = 100{,}000$ & 26.08
& $^{***}$ \\
\hline
\multirow{10}{*}{$n \times m \geq 500$} & \multirow{5}{*}{\textcolor{black}{Early best}}& \multirow{5}{*}{\textcolor{black}{23.33}} 
  & $E_r = 20{,}000$ & 22.18
& \textcolor{black}{$0.012$} \\
& & & $E_r = 40{,}000$ & 20.44
& $^{***}$ \\
& & & $E_r = 60{,}000$ & 20.89
& $^{***}$ \\
& & & $E_r = 80{,}000$ & 17.97
& $^{***}$ \\
& & & $E_r = 100{,}000$ & 21.67
& $^{***}$ \\
\cline{2-6}
& \multirow{5}{*}{\textcolor{black}{Converging best}}& \multirow{5}{*}{\textcolor{black}{27.74}}& $E_r = 20{,}000$ & 22.18
& $^{***}$ \\
& & & $E_r = 40{,}000$ & 20.44
& $^{***}$ \\
& & & $E_r = 60{,}000$ & 20.89
& $^{***}$ \\
& & & $E_r = 80{,}000$ & 17.97
& $^{***}$ \\
& & & $E_r = 100{,}000$ & 21.67
& $^{***}$ \\
\hline
\end{tabular}
\end{table}

\subsubsection{Comparison with recent learning-based paradigms}
\label{sec:paradigm_comparison}
\textcolor{black}{
While the preceding comparisons focus on methods within the GNN-DRL paradigm, 
recent advances in learning-based approaches have produced methods operating on 
fundamentally different paradigmatic axes: Self-Labeling Improvement Method (SLIM)~\cite{corsini2024self} replaces reward-driven training with supervised pseudo-labels; 
Reinforced Adaptive Staircase Curriculum Learning (RASCL)~\cite{iklassov2023study} progressively trains on instances of increasing difficulty; DRL-guided local search combines neural 
policies with iterative improvement heuristics~\cite{zhang2024deep}; Self-Evolutionary (SeEvo)~\cite{huang2026automatic} leverages large language models to automatically generate and evolve dispatching rules.}
\par
\textcolor{black}{
To situate VG2S within this broader landscape, Table~\ref{tab:paradigm_comparison} presents a comparison with SLIM, RASCL, DRL-guided local search, and SeEvo, each representing a fundamentally different learning paradigm. Each method reflects a distinct trade-off relative to VG2S.
SLIM~\cite{corsini2024self} yields high solution quality, it discards all but the single best trajectory from each sampled rollout, in contrast to 
VG2S which fully exploits all sampled trajectories for policy learning.
RASCL~\cite{iklassov2023study} retains reward-driven training but structures 
it through a curriculum, additionally compensating at inference time via 
128-sample augmentation; unlike VG2S which deploys as a single forward pass, 
this requires substantially higher inference cost. Furthermore, this approach achieves the highest performance among the compared methods.
DRL-guided local search~\cite{zhang2024deep} forgoes constructive scheduling altogether, instead learning to iteratively refine complete solutions via local search; this fundamentally differs from VG2S's single forward pass deployment and incurs repeated inference cost per instance.
SeEvo~\cite{huang2026automatic} replaces learned neural representations with 
Large Language Model (LLM)-generated heuristic rules, requiring no dedicated training but operating 
without instance-specific structural encoding, resulting in substantially 
lower solution quality compared to VG2S across all evaluated benchmarks.
This comparison serves to characterize the performance landscape across 
fundamentally distinct learning paradigms, each operating under different 
inference regimes and supervision assumptions. A detailed discussion of the paradigmatic distinctions between VG2S and the other methods, and their implications, is provided in Supplementary Material I.}

\begin{table}[!ht]
\centering
\renewcommand{\arraystretch}{1.3}
\caption{\textcolor{black}{Optimality gap comparison across paradigms on TA and DMU benchmarks. GNN-DRL baselines (Yuan, Oh) and VG2S share the same paradigm; SLIM, RASCL, and SeEvo represent different paradigmatic axes. Blank cells indicate benchmarks not reported in the respective original papers.}}

\label{tab:paradigm_comparison}
{\color{black}\begin{tabular}{ccccccccc}
\hline
\multirow{2}{*}{Instance Size} & \multicolumn{3}{c}{GNN-DRL paradigm} & \multicolumn{4}{c}{Other paradigms} \\
\cmidrule(lr){2-4} \cmidrule(lr){5-8}
 & Yuan~\cite{yuan2023solving} & Oh~\cite{oh2025framework} & \textbf{VG2S} & \makecell{SLIM} & \makecell{RASCL} & \makecell{DRL-guided \\ local search } & \makecell{SeEvo} \\
\hline
TA $15 \times 15$  & 21.32 & 14.70 & 15.66 & 13.77 & \textbf{9.02}  &9.29& 17.27 \\
TA $20 \times 15$  & 22.39 & 16.58 & 15.37 & 14.96 & \textbf{10.58} &11.67& 21.44 \\
TA $20 \times 20$  & 20.92 & 16.69 & 16.30 & 15.17 & \textbf{10.87} & 12.03&20.43 \\
TA $30 \times 15$  & 23.28 & 19.07 & 19.32 & 17.09 & \textbf{13.98} & 14.94&23.50 \\
TA $30 \times 20$  & 26.33 & 19.69 & 20.62 & 18.49 & \textbf{16.09} & 17.52&25.71 \\
TA $50 \times 15$  & 16.03 & 13.35 & 11.76 & 10.06 & \textbf{9.32}  & 10.19&15.92 \\
TA $50 \times 20$  & 17.61 & 13.23 & 13.01 & 11.55 & \textbf{9.89}  & 12.45&18.26 \\
TA $100 \times 20$ & 8.91  & 7.17  & 6.95  & 5.85  & \textbf{3.96}  & 7.29&9.59  \\
\hline
DMU $20 \times 15$ & 27.74 & 23.14 & 19.73 & \textbf{17.92} & 19.36 & - & 27.35 \\
DMU $20 \times 20$ & 23.56 & 18.72 & 20.31 & 19.37          & \textbf{15.98} &-& 24.22 \\
DMU $30 \times 15$ & 28.70 & 28.04 & 22.75 & 21.61          & \textbf{16.35} &-& 32.40 \\
DMU $30 \times 20$ & 29.21 & 26.55 & 23.44 & 25.53 & \textbf{20.00}          &-& 31.39 \\
DMU $40 \times 15$ & 25.32 & 26.83 & 20.96 & \textbf{17.37} & 17.49          &-& 29.49 \\
DMU $40 \times 20$ & 32.47 & 28.28 & 26.10 & \textbf{21.96} & 25.42          &-& 33.07 \\
DMU $50 \times 15$ & 24.96 & 25.75 & 17.52 & \textbf{15.45} & 21.54          &-& 30.89 \\
DMU $50 \times 20$ & 33.42 & 29.89 & 26.88 & 19.11          & \textbf{14.66} &-& 34.09 \\
\hline
\end{tabular}}
\end{table}

\subsection{Effectiveness of variational representation learning}
\label{Effectiveness of variational representation learning}
\textcolor{black}{To verify that the observed performance gains are attributable to the variational representation learning process rather than to incidental design choices, we conduct ablation studies on the variational components (Section~\ref{Ablation on variational representation learning}), examine their effect across problem scales (Section~\ref{Analysis across problem scales}), and isolate the contribution of look-ahead state features (Section~\ref{look_ahead}).}
\subsubsection{Ablation on variational representation learning}
\label{Ablation on variational representation learning}
\textcolor{black}{We conduct ablation experiments to validate the contribution of each component in the proposed VG2S. Specifically, we consider three configurations in a progressive manner: (1) End-to-end (E2E) without recon-loss, an end-to-end training baseline without the reconstruction loss term, which serves as the most \textcolor{black}{ablated reference}; (2) E2E with recon-loss, which introduces the reconstruction loss into end-to-end training with joint optimization of policy learning and 
representation learning, isolating the effect of the loss term itself; and (3) the proposed VG2S, which further incorporates a two-phase training strategy with three different $E_r$ values (20,000, 40,000, and 60,000), where the variational representation learning phase is explicitly separated from the policy learning phase. This progressive design allows us to disentangle the individual contributions of the reconstruction loss and the two-phase strategy. To ensure experimental robustness, we train each configuration with eight  \textcolor{black}{arbitrarily selected} hyperparameter combinations drawn from IDs 1--5 (Supplementary Material E): 
one variant from ID 1, two from ID 2 ($d_{\text{latent}} \in \{64, 96\}$), 
two from ID 3 ($d_{\text{latent}} \in \{96, 128\}$), two from ID 4 
($d_{\text{latent}} \in \{96, 128\}$), and one from ID 5, with identical hyperparameters applied across all configurations. All experiments are conducted over 15,000 policy learning epochs, with validation performed every 200 epochs across four benchmark datasets: TA61, TA62, DMU76, and DMU77.}
\par
\begin{figure}
    \centering
    \begin{subfigure}[b]{0.48\textwidth}
        \centering
        \includegraphics[width=\textwidth]{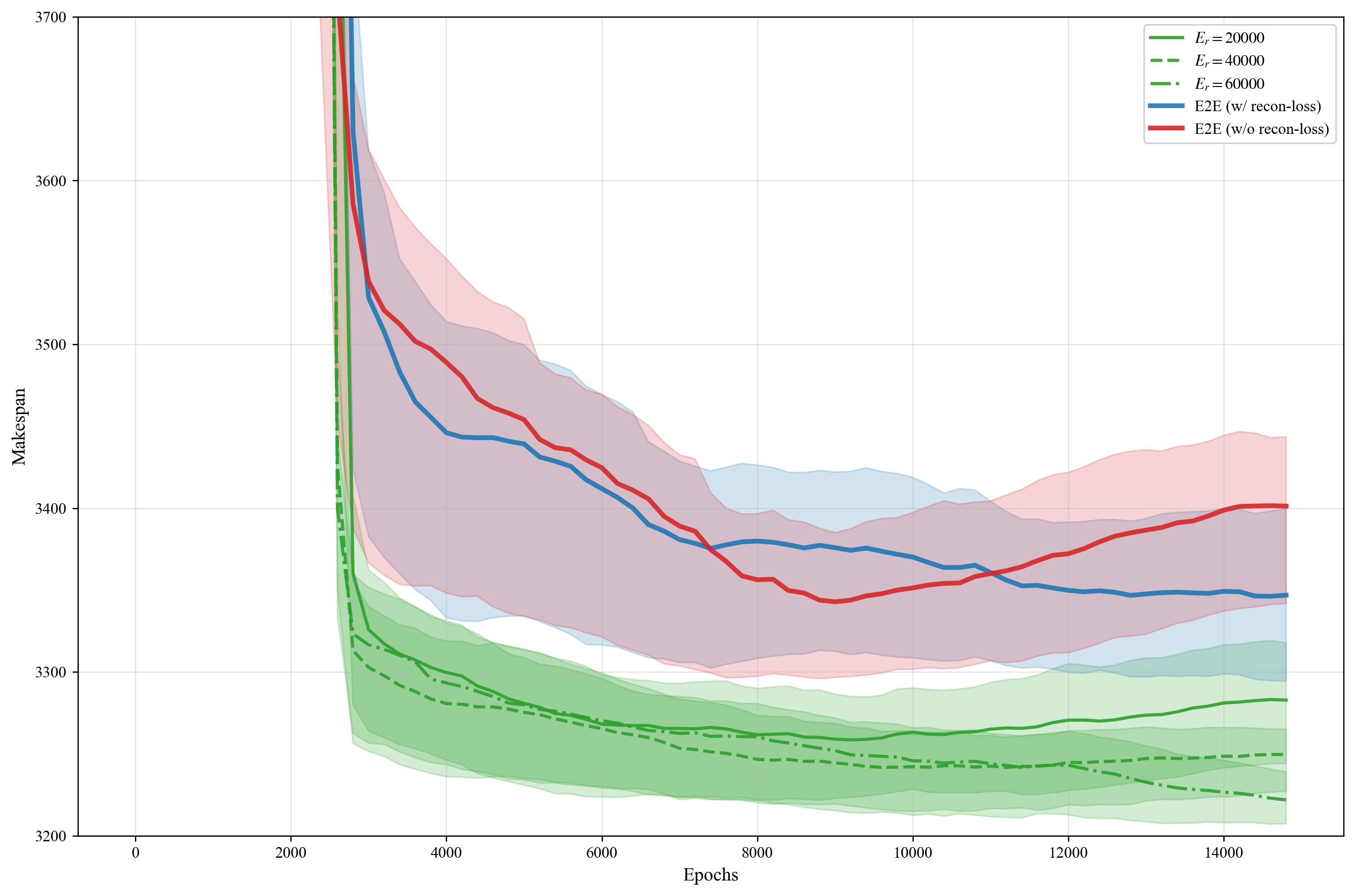}
        \caption{TA61}
        \label{TA61}
    \end{subfigure}
    \hfill
    \begin{subfigure}[b]{0.48\textwidth}
        \centering
        \includegraphics[width=\textwidth]{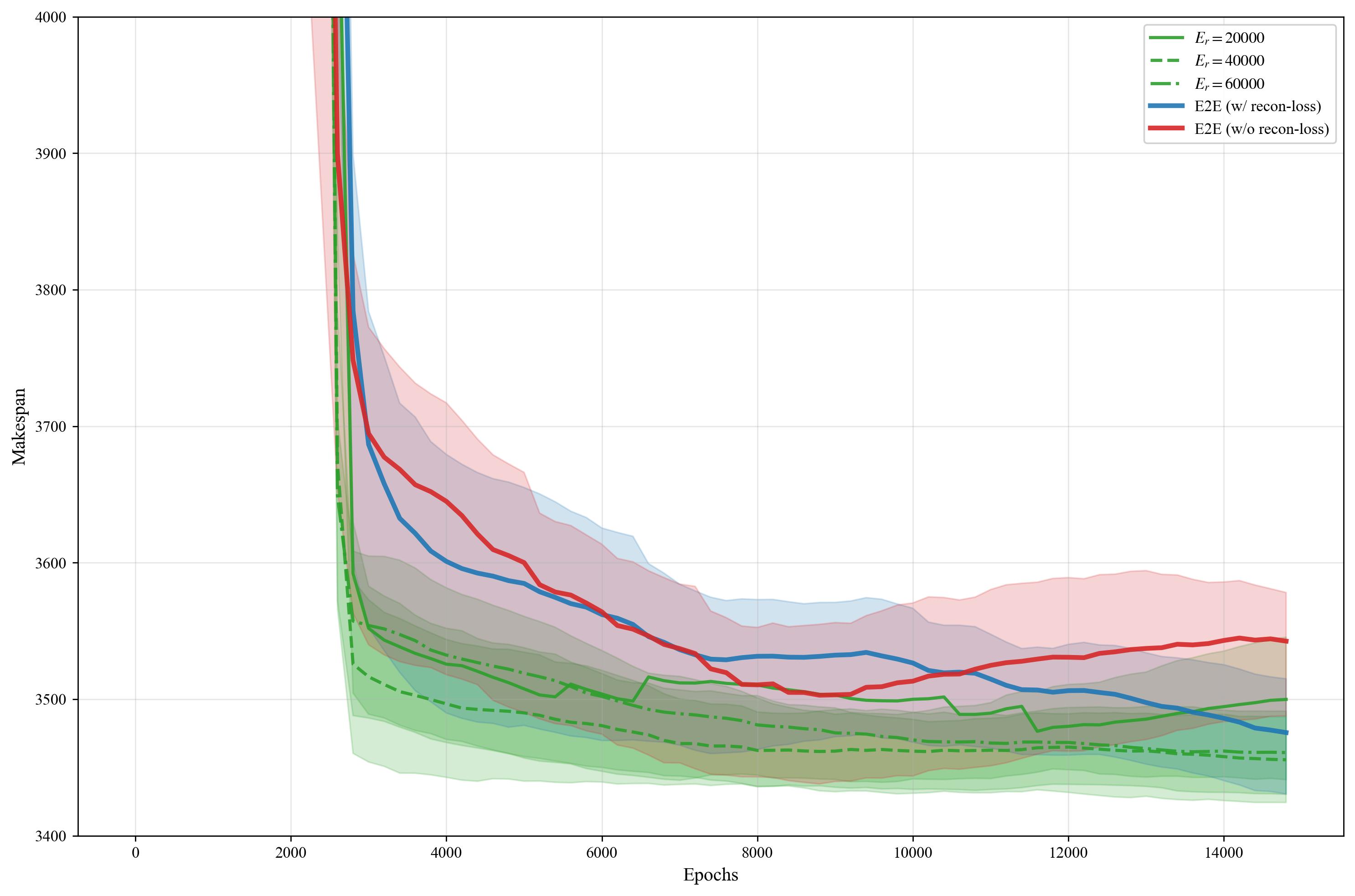}
        \caption{TA62}
        \label{TA62}
    \end{subfigure}
        \hfill
    \begin{subfigure}[b]{0.48\textwidth}
        \centering
        \includegraphics[width=\textwidth]{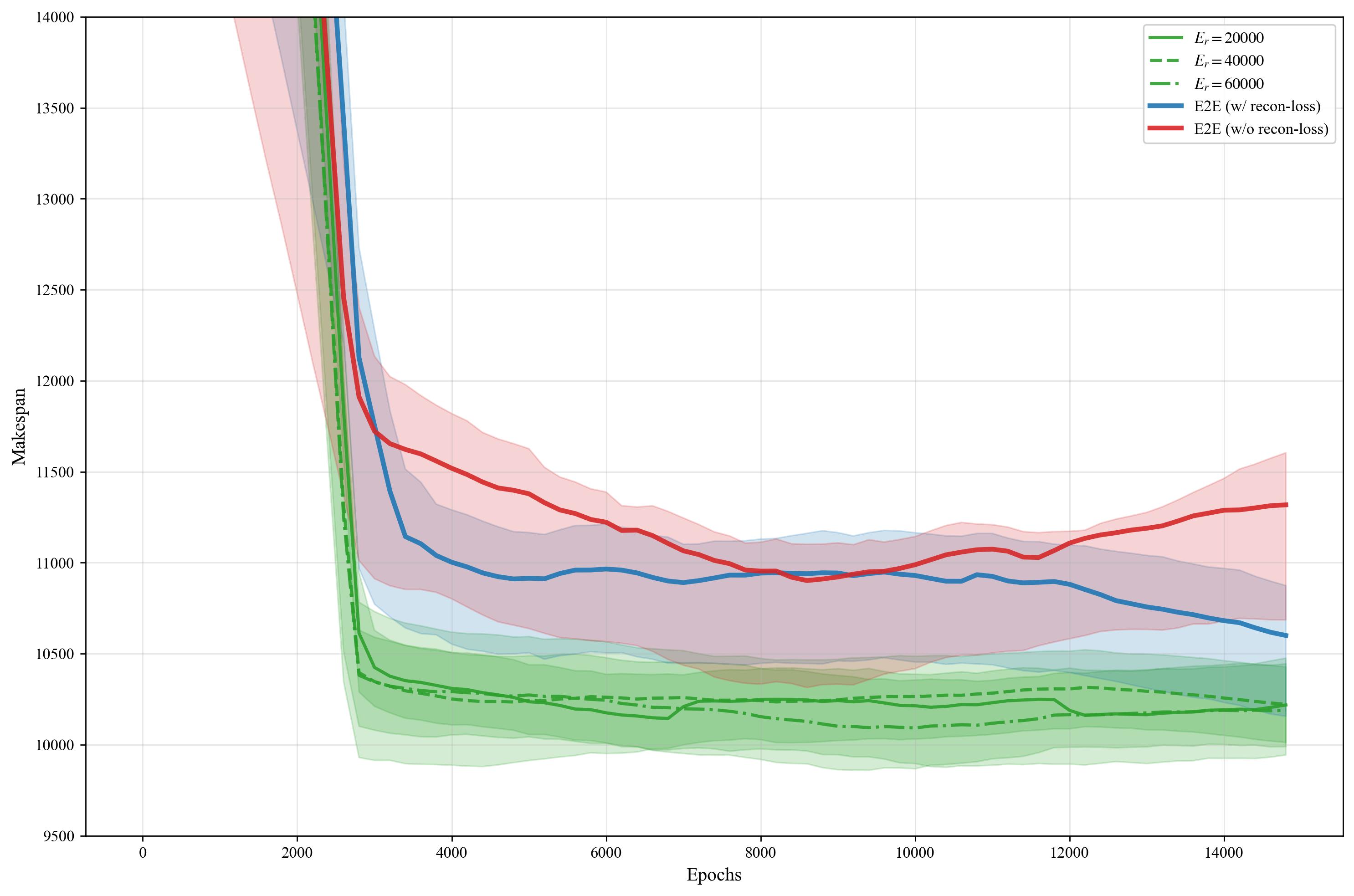}
        \caption{DMU76}
        \label{DMU76}
    \end{subfigure}
        \begin{subfigure}[b]{0.48\textwidth}
        \centering
        \includegraphics[width=\textwidth]{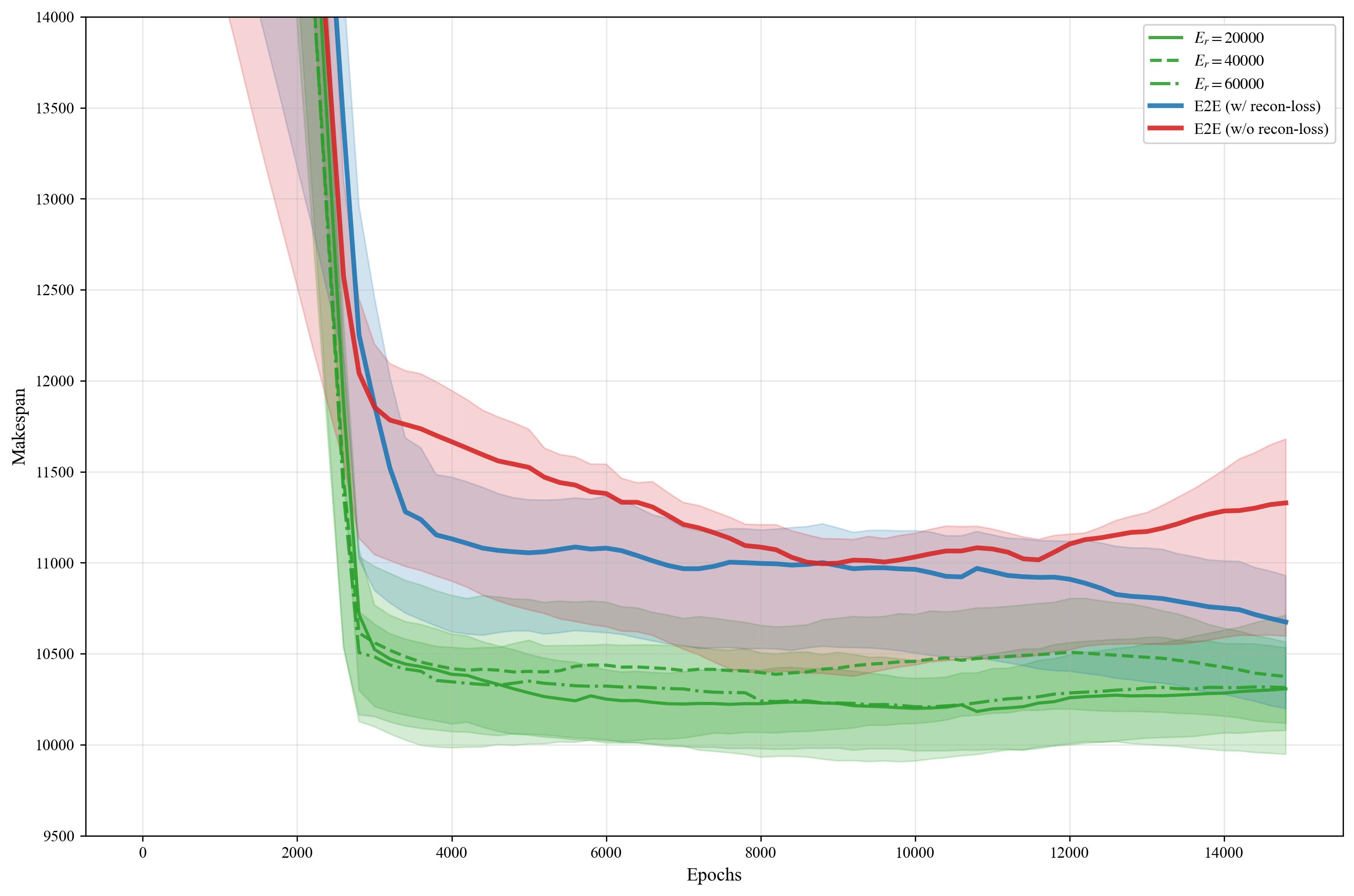}
        \caption{DMU77}
        \label{DMU77}
    \end{subfigure}
    \caption{Learning curves showing validation performance of average makespan on benchmark datasets}

    \label{learning curves}
\end{figure}

Fig.~\ref{learning curves} illustrates the results using a moving average filter (window: 25): the solid line represents the mean performance across hyperparameter combinations at each epoch, and the shaded area denotes the interquartile range (25th to 75th percentile). E2E without recon-loss exhibits persistent instability throughout training, with large interquartile ranges reflecting high sensitivity to hyperparameter choices. Introducing the reconstruction loss (E2E with recon-loss) yields a partial improvement: the learning curve becomes somewhat more stable and the final makespan is modestly reduced, suggesting that the reconstruction loss provides a regularizing effect even without a temporally decoupled representation learning phase. 
\par
However, the most significant gains are observed when the reconstruction loss is integrated with the two-phase training strategy. On TA61, all three VG2S configurations converge to an average range of 3,200--3,300, outperforming both E2E without recon-loss (approximately $3,350-3,400$) and E2E with recon-loss ($\approx 3,350$), while exhibiting significantly narrower interquartile ranges. On TA62, both VG2S configurations and E2E with recon-loss demonstrate superior final performance compared to E2E without recon-loss, converging to significantly lower makespan levels (approximately 3,450--3,520) with substantially narrower variance bands. In contrast, although the 25th percentile of E2E without recon-loss shows makespans comparable to VG2S, the model demonstrates high instability characterized by a broad interquartile range and a deteriorating trend after hitting its own local minimum.
\par
\textcolor{black}{The progressive improvement is even more pronounced on the DMU instances. E2E without recon-loss exhibits large variance, with its mean occasionally approaching the upper bound of its own interquartile range. This indicates a right-skewed performance distribution, where certain hyperparameter configurations yield substantially degraded solutions that pull the mean above the 75th percentile. This further corroborates the poor hyperparameter robustness of this baseline. Adding the reconstruction loss (E2E with recon-loss) alleviates this instability to some degree, reducing both the variance and the frequency of such right-skewed behavior. Nevertheless, a decisive improvement emerges with the proposed VG2S incorporating the two-phase strategy: the final mean makespan converges to approximately 10,000--10,300 on DMU76, compared to E2E without recon-loss's 10,500--10,700. A consistent pattern of relative superiority among the three configurations is also observed on DMU77. Notably, VG2S reaches stable convergence within approximately 8,000--10,000 epochs across all DMU benchmarks, whereas both baselines fail to reach a stable plateau within the same policy learning budget, further highlighting the efficiency of the two-phase strategy.}

\par
\textcolor{black}{A natural concern is whether the observed gains are primarily attributable to the two-phase training strategy itself, independent of the ELBO-based variational objective. To address this, we trained eight experimental configurations identical to those used in this section under a deterministic pre-training baseline, where $\mathcal{L}_{\text{KL}}$ is removed and the stochastic sampling in \eqref{latent1}--\eqref{latent3} is replaced with a deterministic encoding. Despite gradient clipping being applied, all eight configurations resulted in numerical divergence, confirming that the instability is fundamental to the training in the absence of $\mathcal{L}_{\text{KL}}$. This demonstrates that 
$\mathcal{L}_{\text{KL}}$ serves a dual role: not only as a probabilistic 
regularization term that structures the latent space, but also as a 
fundamental stability mechanism that bounds the latent variable during 
variational representation learning.}

\textcolor{black}{Overall, the ablation results demonstrate that both variational representation learning based on the reconstruction loss and the two-phase training strategy are jointly necessary for the stability and performance of VG2S. Furthermore, the failure of the deterministic pre-training baseline confirms that the two-phase strategy alone without the ELBO is insufficient. These components (ELBO-based objective and two-phase training) are jointly indispensable for effective variational representation learning in JSSP.}

\subsubsection{Analysis across problem scales}
\label{Analysis across problem scales}
\begin{figure}
    \centering
    \begin{subfigure}[b]{0.48\textwidth}
        \centering
        \includegraphics[width=\textwidth]{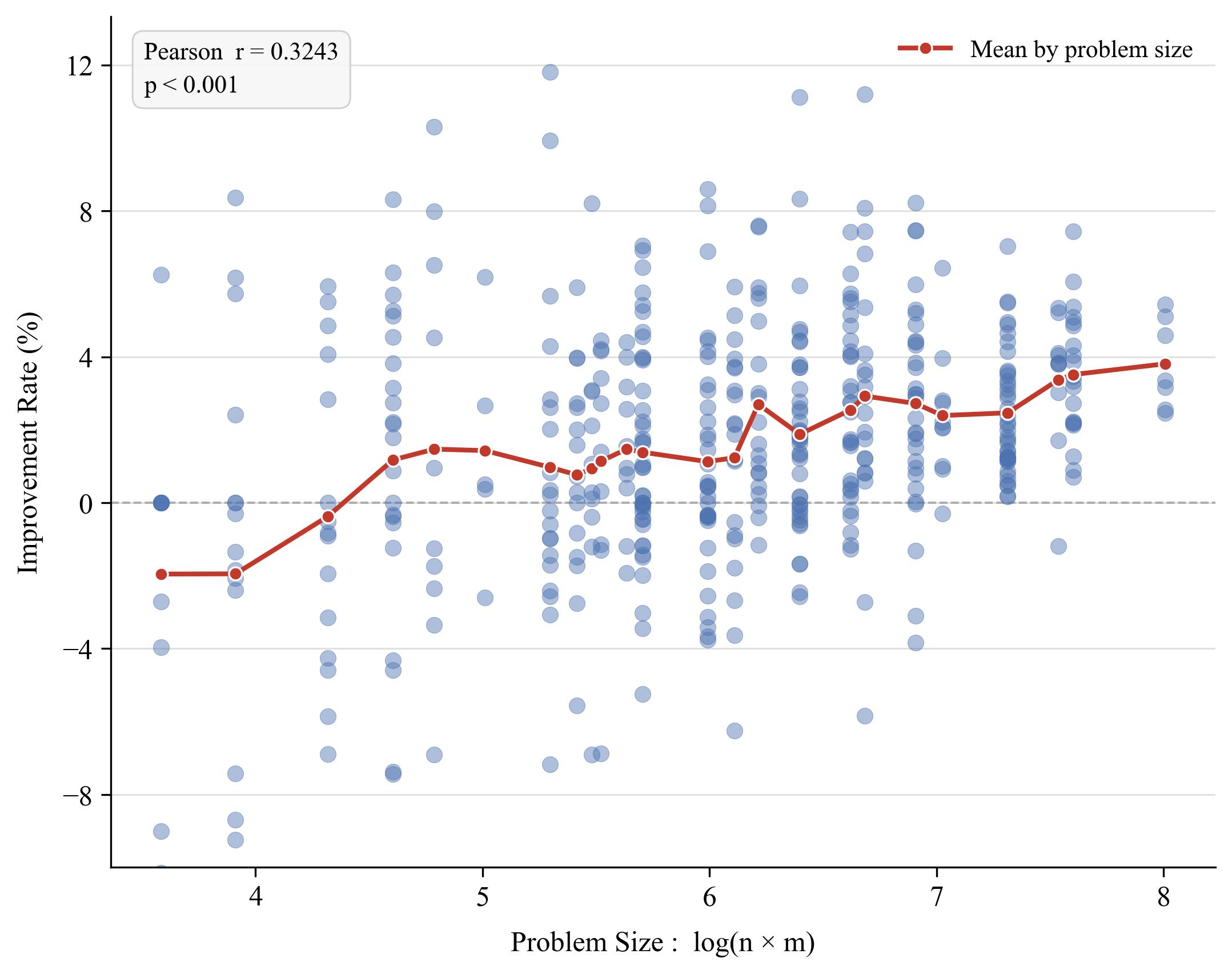}
        \caption{Improvement rate}
        \label{improvement_rate}
    \end{subfigure}
    \hfill
    \begin{subfigure}[b]{0.48\textwidth}
        \centering
        \includegraphics[width=\textwidth]{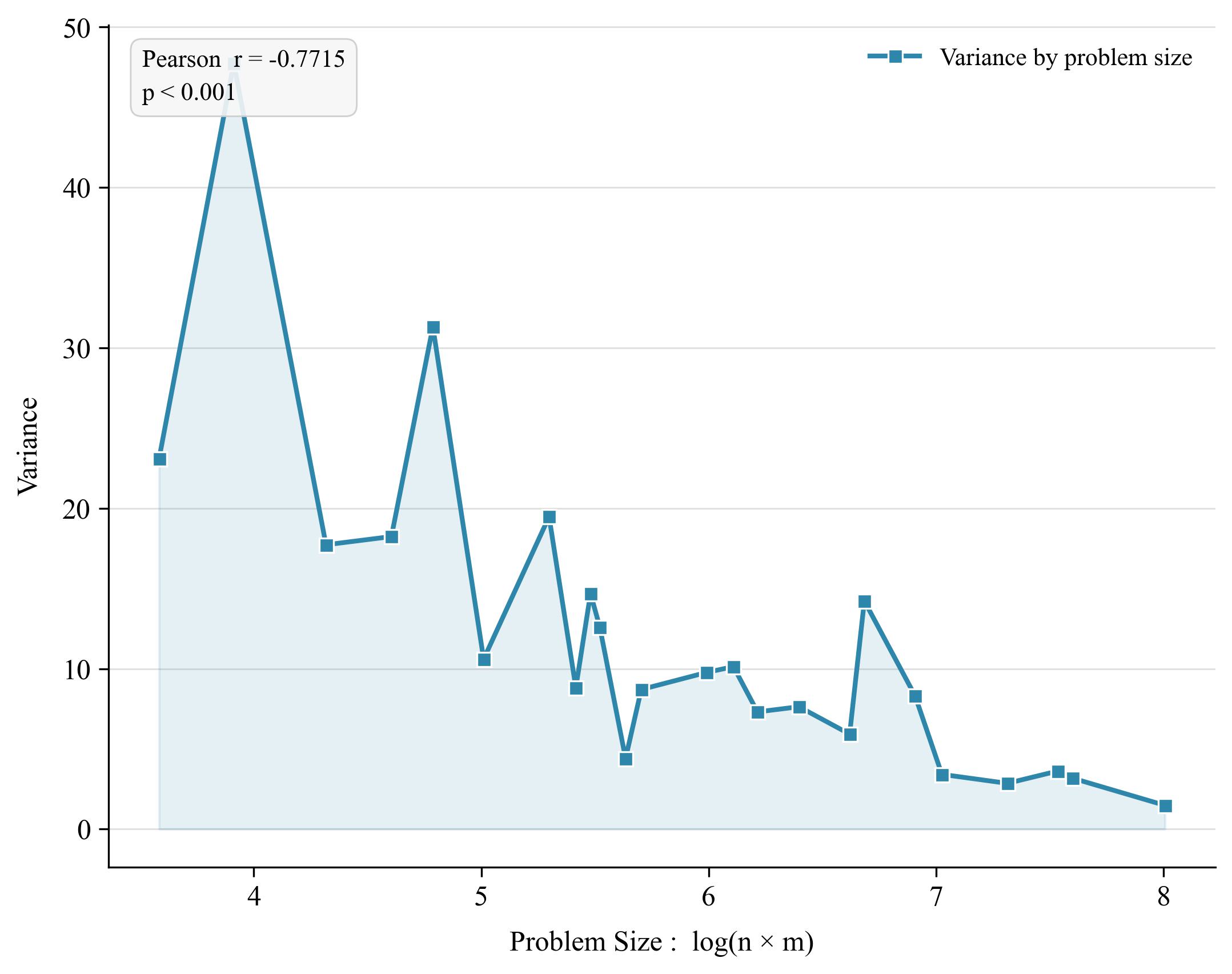}
        \caption{Variance}
        \label{variance}
    \end{subfigure}
    \caption{Trends in improvement rate and performance variance with respect to problem size}
\end{figure}

To further verify the utility of variational representation learning in relation to problem scale, we conduct comprehensive testing on 463 instances \textcolor{black}{(the 242 standard benchmark instances of Section 6.1.2, augmented with 221 randomly generated instances)}. We compare it against the \textcolor{black}{E2E without recon-loss.} \textcolor{black}{For this evaluation, we selected the best-performing models based on average validation performance on TA61, TA62, DMU76, and DMU77: the baseline at the 14,401st epoch (ID 3) and the proposed VG2S at the 6,001st policy learning epoch (ID 4, $E_r = 80,000$).} The performance metrics are the average improvement rate (red line in Fig.~\ref{improvement_rate}) and the variance (blue line in Fig.~\ref{variance}). The improvement rate is
calculated as: $100 \times \frac{C_{\max}^{\text{baseline}} - C_{\max}^{\text{VG2S}}}{C_{\max}^{\text{baseline}}}$,
\textcolor{black}{where $C_{\max}^{\text{VG2S}}$ and $C_{\max}^{\text{baseline}}$ are the makespans of VG2S and the E2E without recon-loss, respectively.}

\par
Fig.~\ref{improvement_rate} presents the relationship between problem size 
(measured as $\log(n \times m)$) and the improvement rate. 
The scatter plot reveals substantial variation in performance gains across individual 
instances, with improvement rates ranging from approximately -10\% to +12\%. However, 
the red line representing the mean improvement rate by problem size demonstrates a 
clear upward trend, with a statistically significant positive correlation 
(Pearson $r = 0.3243$, $p < 0.001$). This confirms that the performance superiority of the proposed method becomes clearly evident with increasing problem scale. Notably, 
for smaller problems ($\log(n \times m) < 4$), the mean improvement \textcolor{black}{rate} is negative 
or near-zero, suggesting that the baseline performs comparably or slightly better 
on simpler instances. However, for larger problems ($\log(n \times m) > 5$), the 
proposed method consistently achieves positive improvements, with mean gains 
reaching approximately 2-3\% for the largest problem sizes.
\par
Fig.~\ref{variance} provides complementary insights by examining the variance 
of improvement rates across problem sizes. The plot reveals a strong negative 
correlation (Pearson $r = -0.7715$, $p < 0.001$) between problem size and 
performance variance. For small problems ($\log(n \times m) \approx 3.5-4.5$), 
the variance is exceptionally high (reaching values near 50), indicating highly 
inconsistent performance across different instances. In contrast, as problem size 
increases, the variance decreases dramatically, stabilizing below 5 for problems 
with $\log(n \times m) > 6.5$. This pattern demonstrates that for smaller instances, neither method shows 
consistent superiority, with highly variable performance across different problems. 
However, as problem scale increases, the proposed VG2S exhibits increasingly 
stable and evident improvements over the baseline. \textcolor{black}{Collectively, these results confirm that variational representation learning plays a critical role in improving solution quality and stability, with its advantage becoming increasingly pronounced as problem scale grows.}

\subsubsection{Ablation on look-ahead state features}
\label{look_ahead}
\textcolor{black}{The preceding ablation in Section~\ref{Ablation on variational representation learning} isolates the contribution of variational representation learning under a fixed state-feature design. A complementary question, however, concerns the role of the look-ahead state features ($s_{4,t}^{ij}$--$s_{6,t}^{ij}$) themselves. This analysis is motivated by the following question: since look-ahead features have been shown to substantially improve scheduling performance~\cite{oh2025framework}, a natural concern is whether VG2S's gains derive primarily from this heuristic prior rather than from variational representation learning itself. To disentangle these effects, we conduct an ablation comparing the full VG2S, which retains $s_{1,t}^{ij}$--$s_{6,t}^{ij}$, against VG2S-lite, which excludes the look-ahead state features $s_{4,t}^{ij}$--$s_{6,t}^{ij}$. The two configurations share an identical architecture, except for the state feature input layer.}

\begin{figure}[htbp]
    \centering
    \begin{subfigure}[b]{0.48\textwidth}
        \centering
        \includegraphics[width=\textwidth]{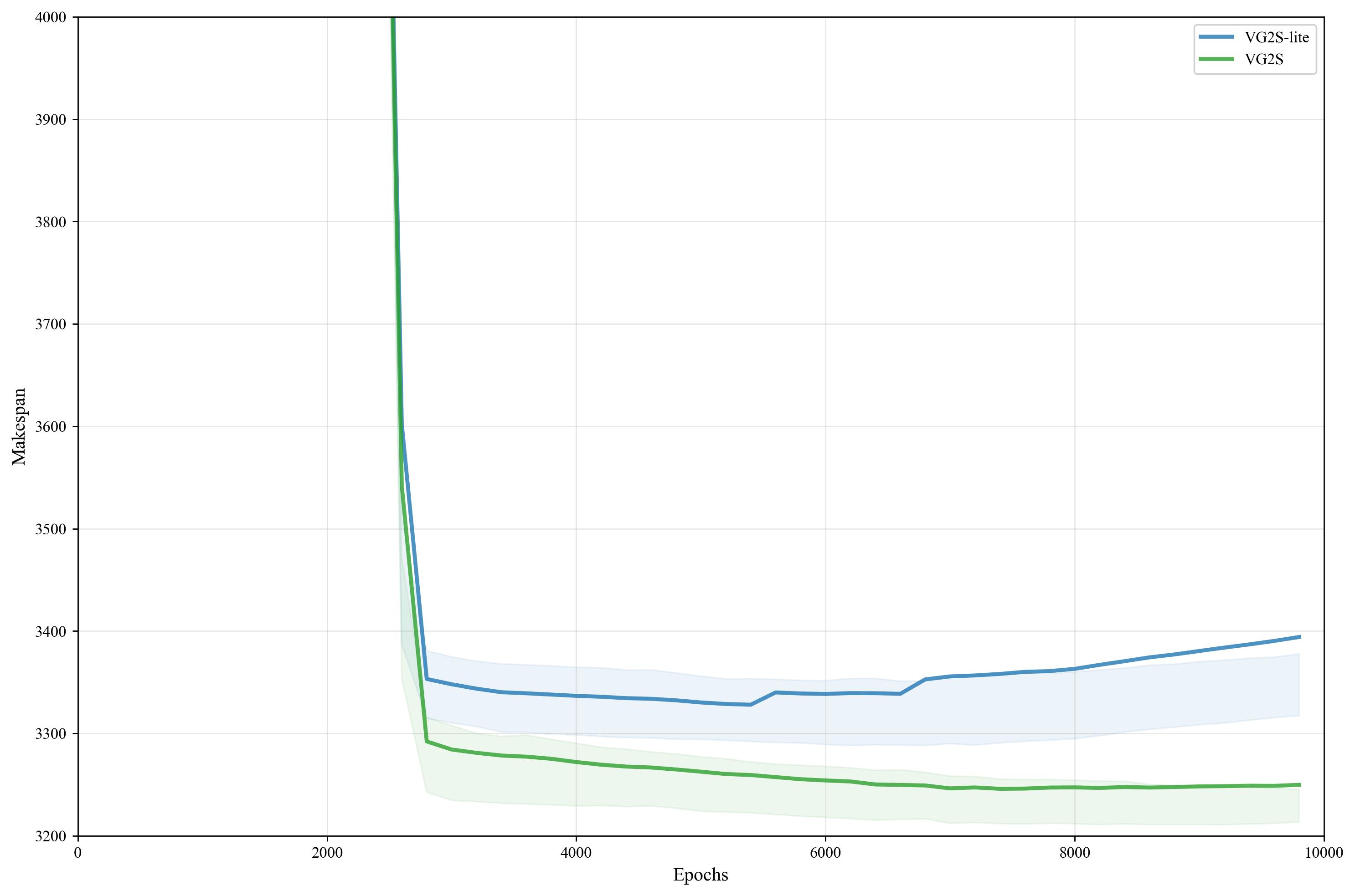}
        \caption{TA61}
    \end{subfigure}
    \begin{subfigure}[b]{0.48\textwidth}
        \centering
        \includegraphics[width=\textwidth]{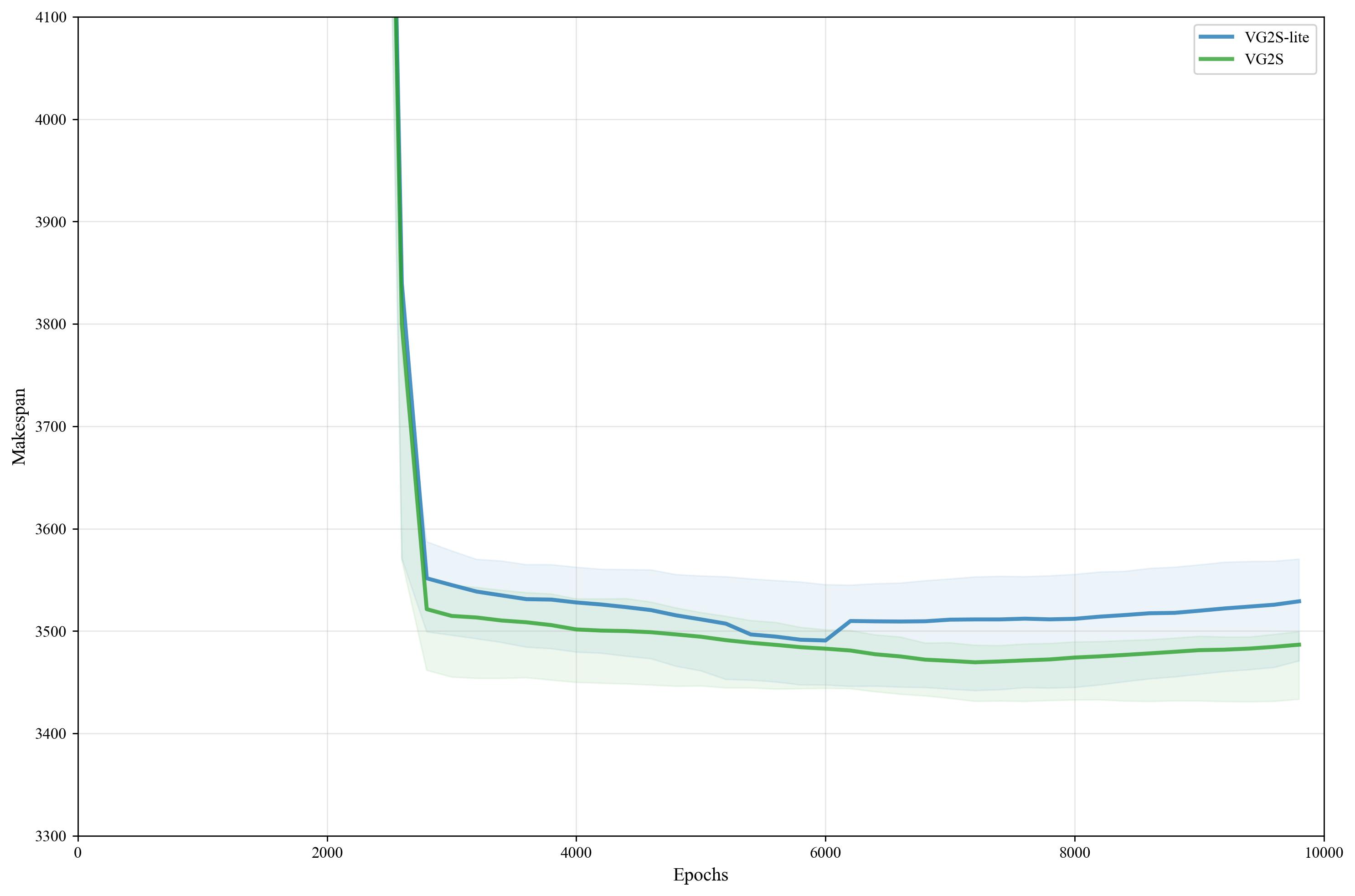}
        \caption{TA62}
    \end{subfigure}
    \\[0.5em]
    \begin{subfigure}[b]{0.48\textwidth}
        \centering
        \includegraphics[width=\textwidth]{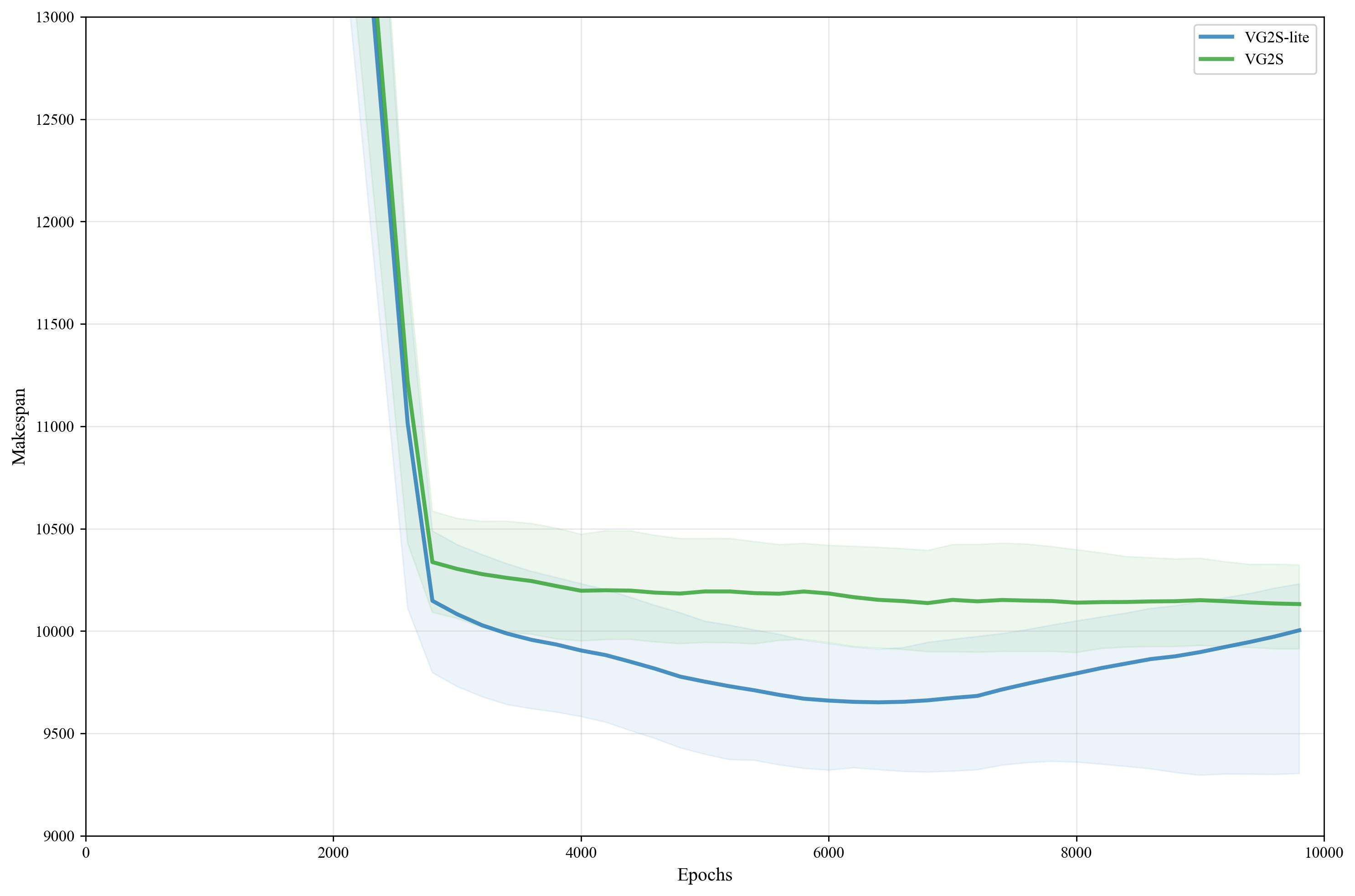}
        \caption{DMU76}
    \end{subfigure}
    \begin{subfigure}[b]{0.48\textwidth}
        \centering
        \includegraphics[width=\textwidth]{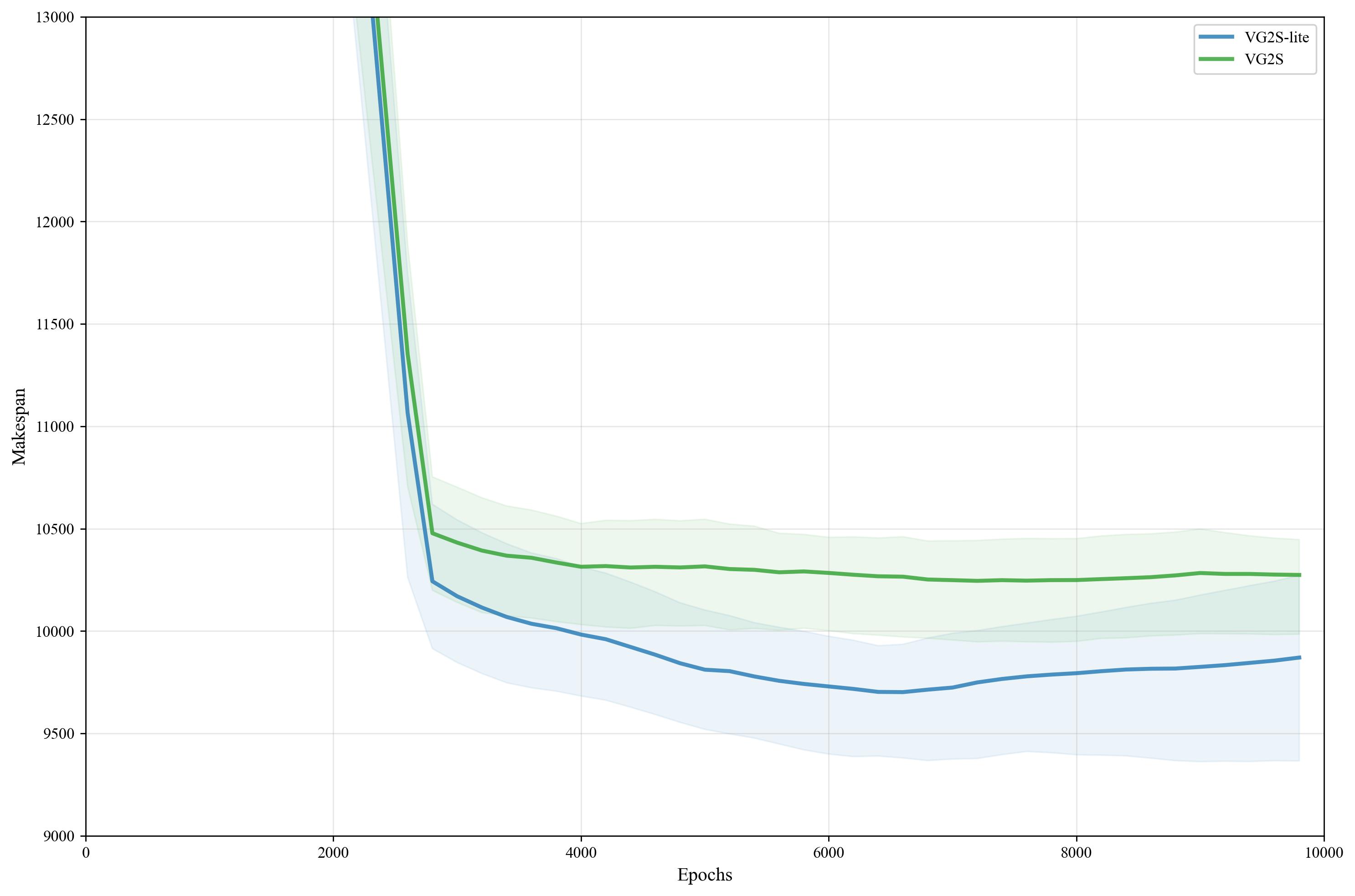}
        \caption{DMU77}
    \end{subfigure}
    \caption{Validation learning curves comparing VG2S (with look-ahead features $s^1$--$s^6$) and VG2S-lite (with $s^1$--$s^3$ only)}
    \label{fig:lookahead_ablation}
\end{figure}

\textcolor{black}{To ensure experimental robustness, we train each configuration with nine arbitrarily selected hyperparameter combinations drawn from Supplementary Material E: two variants for $E_r = 20{,}000$ (ID 1, ID 6), two for $E_r = 40{,}000$ (ID 6, ID 3 with $d_{\text{latent}} = 96$), two for $E_r = 60{,}000$ (ID 2 with $d_{\text{latent}} = 64$, ID 9), and three for $E_r = 80{,}000$ (ID 2, ID 3, ID 4), with identical hyperparameters applied across both VG2S and VG2S-lite. Validation is performed every 200 epochs on four large-sized benchmark instances (TA61, TA62, DMU76, and DMU77), each of size $50 \times 20$. Fig.~\ref{fig:lookahead_ablation} reports the resulting learning curves: the solid line denotes the mean validation makespan across the nine hyperparameter configurations at each evaluation point, and the shaded band represents the 25th--75th percentile range. To suppress short-term fluctuations and reveal the underlying convergence trends, all curves are smoothed with a moving-average filter of window size 25.}

\textcolor{black}{The results reveal a clear and consistent dichotomy depending on the benchmark. On the TA61 and TA62 datasets, VG2S converges to lower makespans than VG2S-lite throughout training. On the DMU76 and DMU77 datasets, this relationship is reversed: VG2S-lite consistently outperforms VG2S over the entire training trajectory. Two key findings follow directly from these pronounced differences. First, look-ahead features yield performance gains on instances with completely random machine sequences (TA), but are not unconditionally beneficial---on structurally partitioned instances (DMU), their inclusion actively degrades performance relative to their absence. Second, more importantly, had VG2S's gains derived primarily from the heuristic prior embedded in look-ahead features, their removal would be expected to yield a consistent and substantial performance degradation across all benchmarks. The performance inconsistency, however, provides empirical evidence suggesting that the look-ahead state features are not the primary driver of VG2S's generalization capability, and that variational representation learning potentially serves as the more fundamental source.}

\subsection{Visual analysis of latent representation via UMAP}
\label{Visual analysis of latent representation via UMAP}

\begin{figure}[H]
    \centering
    
    \textbf{Makespan}
    
    \begin{subfigure}[b]{0.32\textwidth}
        \centering
        \includegraphics[width=\textwidth]{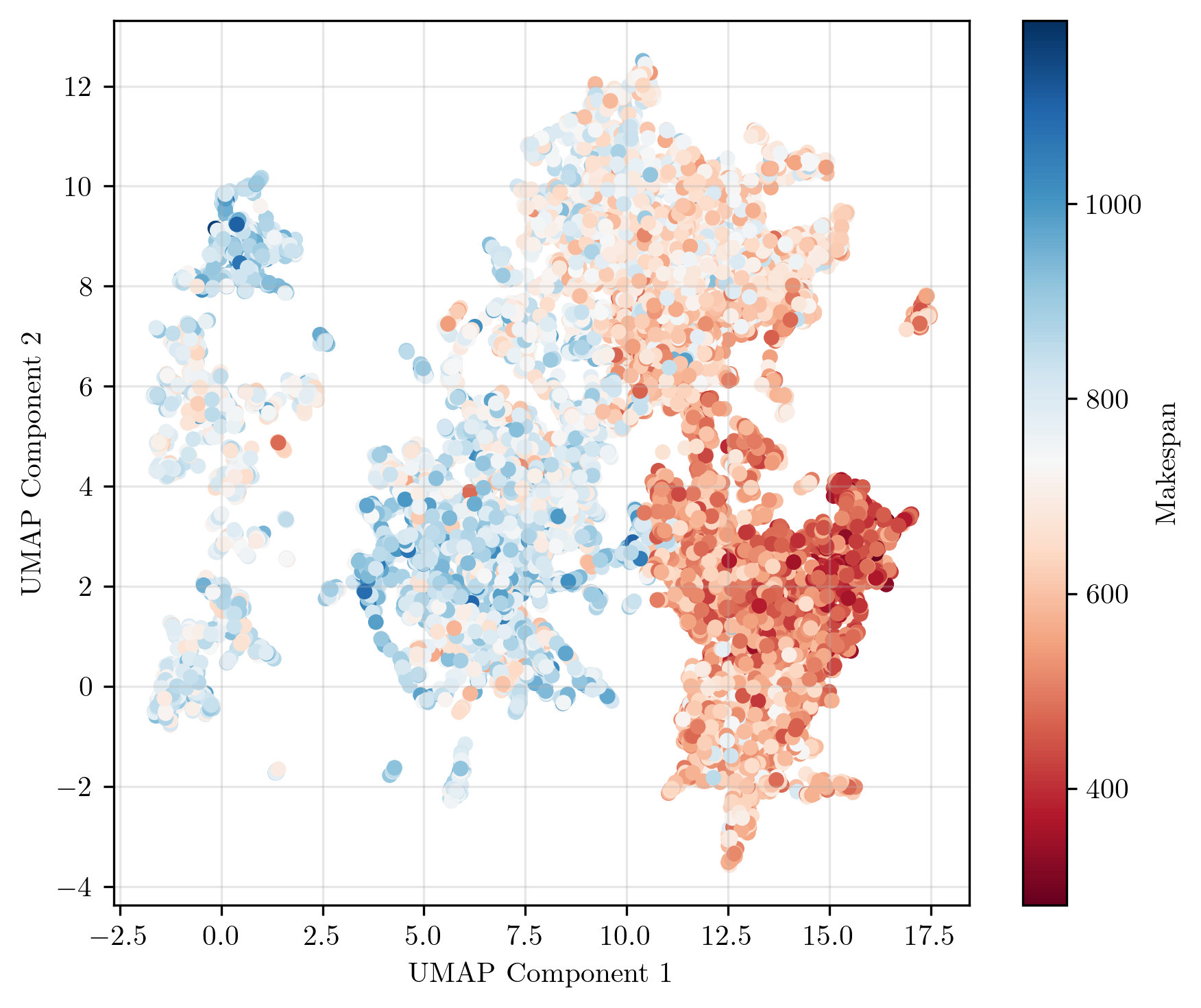}
        \caption{$E_r = 60,000$, $\text{epoch} = 0$}
        \label{a}
    \end{subfigure}
    \hfill
    \begin{subfigure}[b]{0.32\textwidth}
        \centering
        \includegraphics[width=\textwidth]{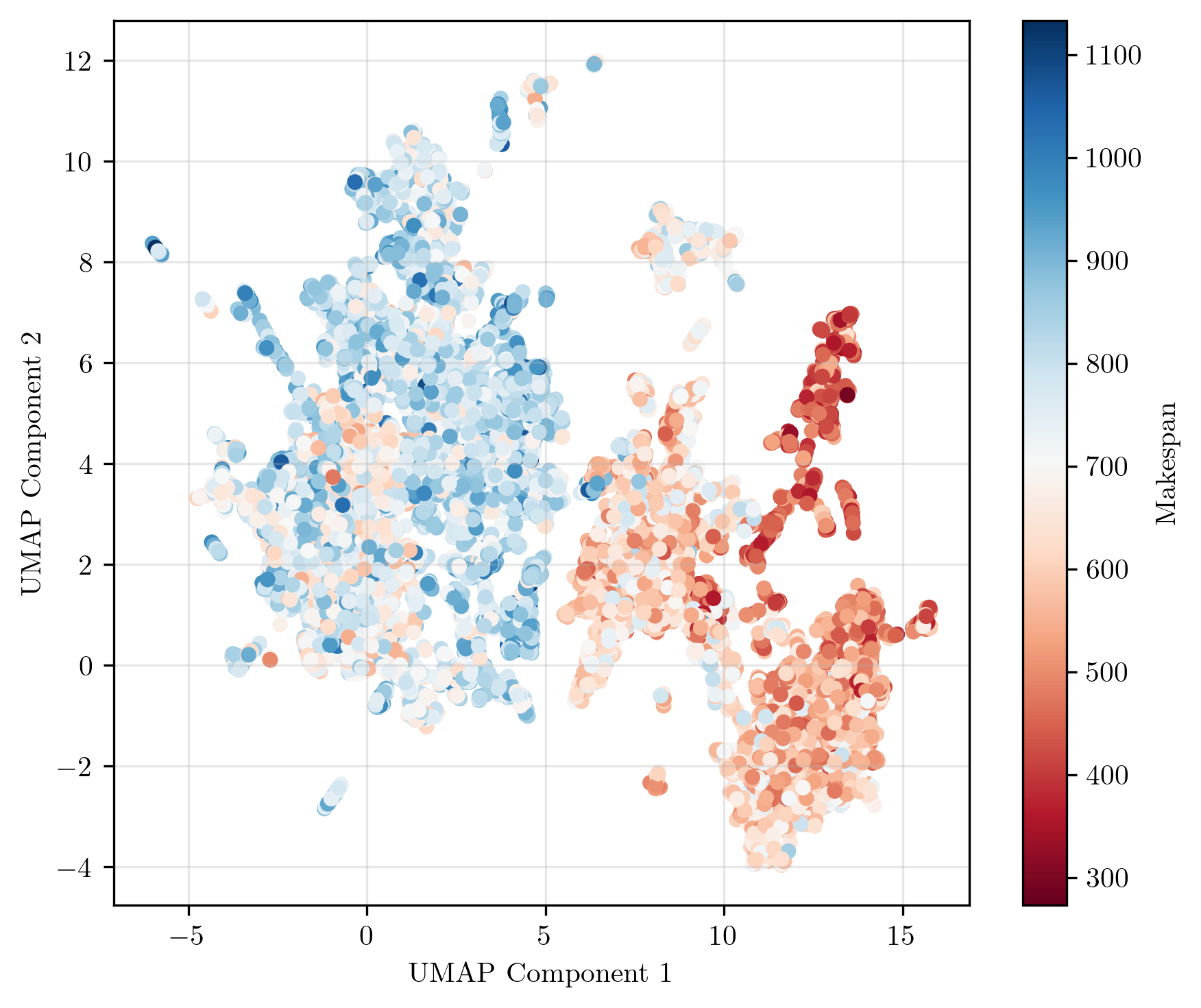}
        \caption{$E_r = 80,000$, $\text{epoch} = 6,000$}
        \label{b}
    \end{subfigure}
    \hfill
    \begin{subfigure}[b]{0.32\textwidth}
        \centering
        \includegraphics[width=\textwidth]{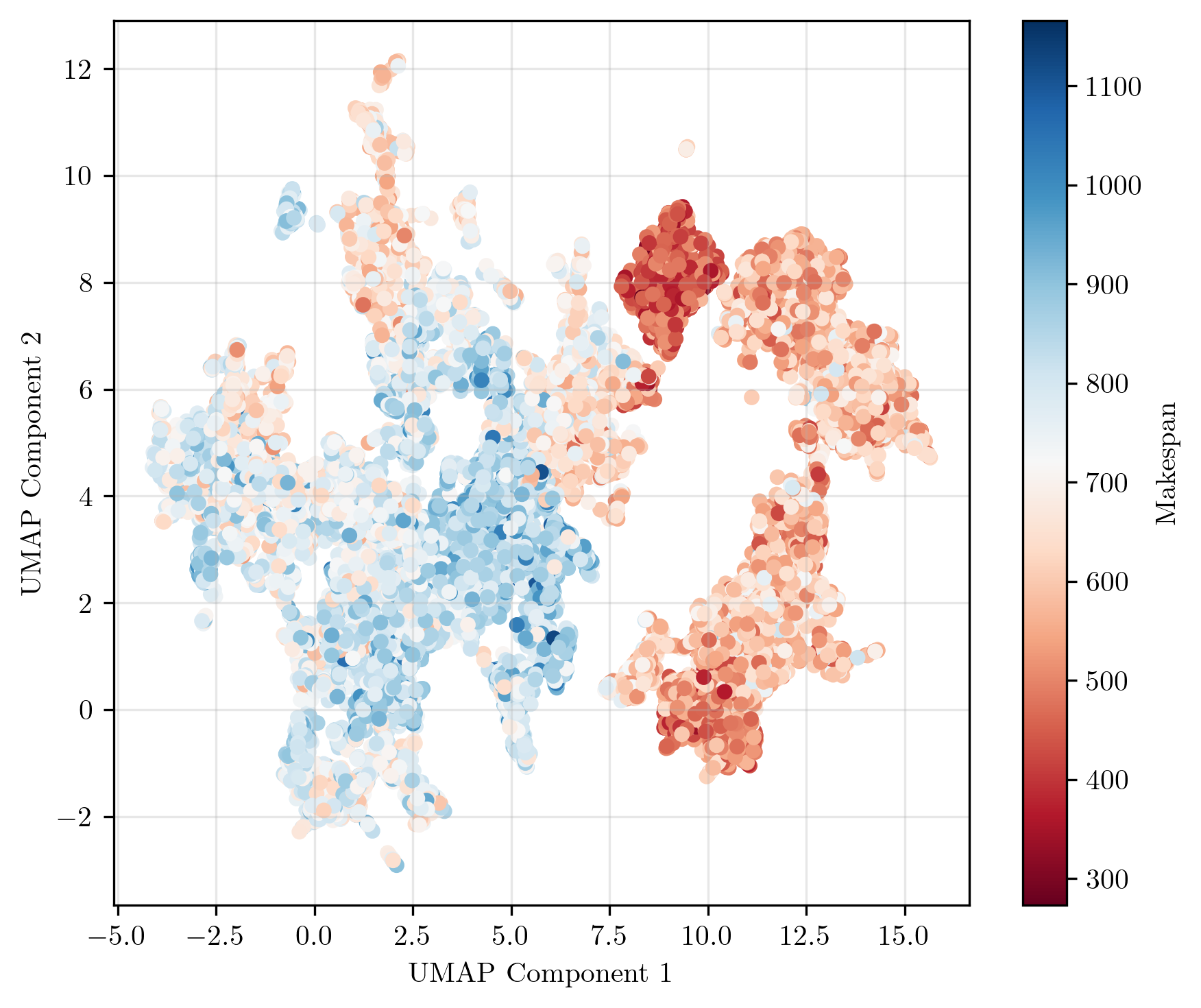}
        \caption{$E_r = 100,000$, $\text{epoch} = 9,000$}
        \label{c}
    \end{subfigure}
    
    \begin{subfigure}[b]{0.32\textwidth}
        \centering
        \includegraphics[width=\textwidth]{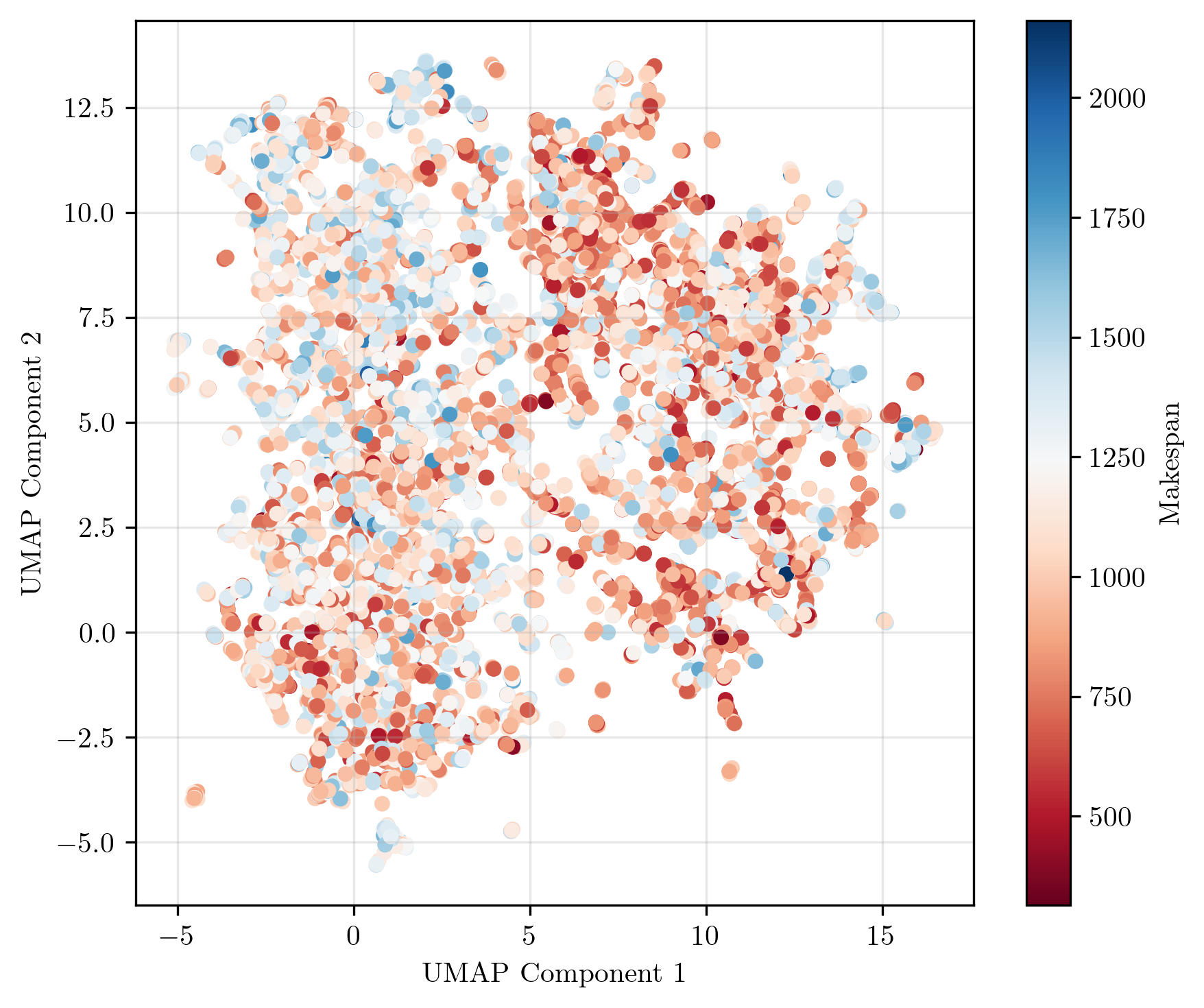}
        \caption{$\text{E2E w/o recon-loss}$, $\text{epoch} = 0$}
        \label{d}
    \end{subfigure}
    \hfill
    \begin{subfigure}[b]{0.32\textwidth}
        \centering
        \includegraphics[width=\textwidth]{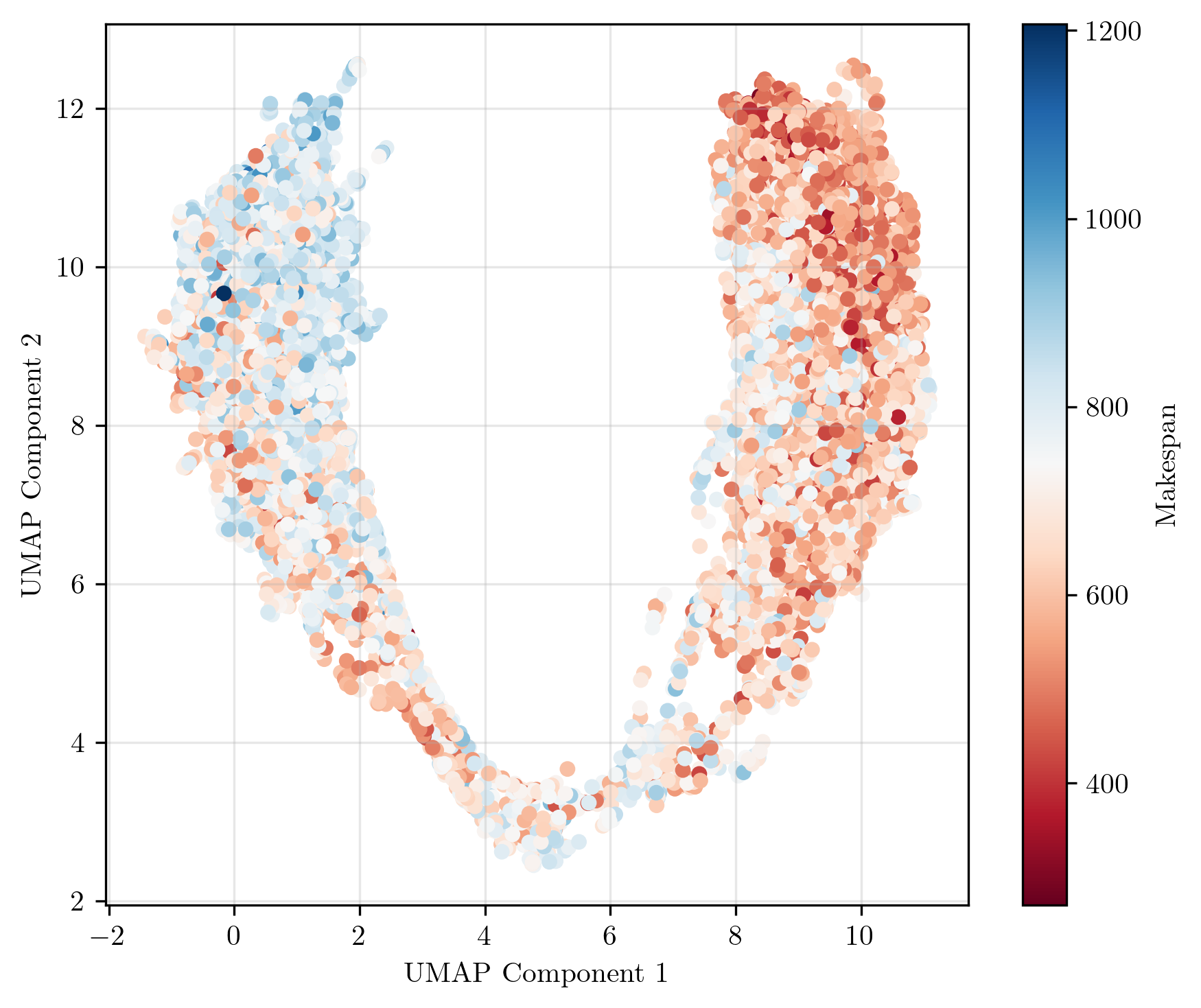}
        \caption{$\text{E2E w/o recon-loss}$, $\text{epoch} = 6,000$}
        \label{e}
    \end{subfigure}
    \hfill
    \begin{subfigure}[b]{0.32\textwidth}
        \centering
        \includegraphics[width=\textwidth]{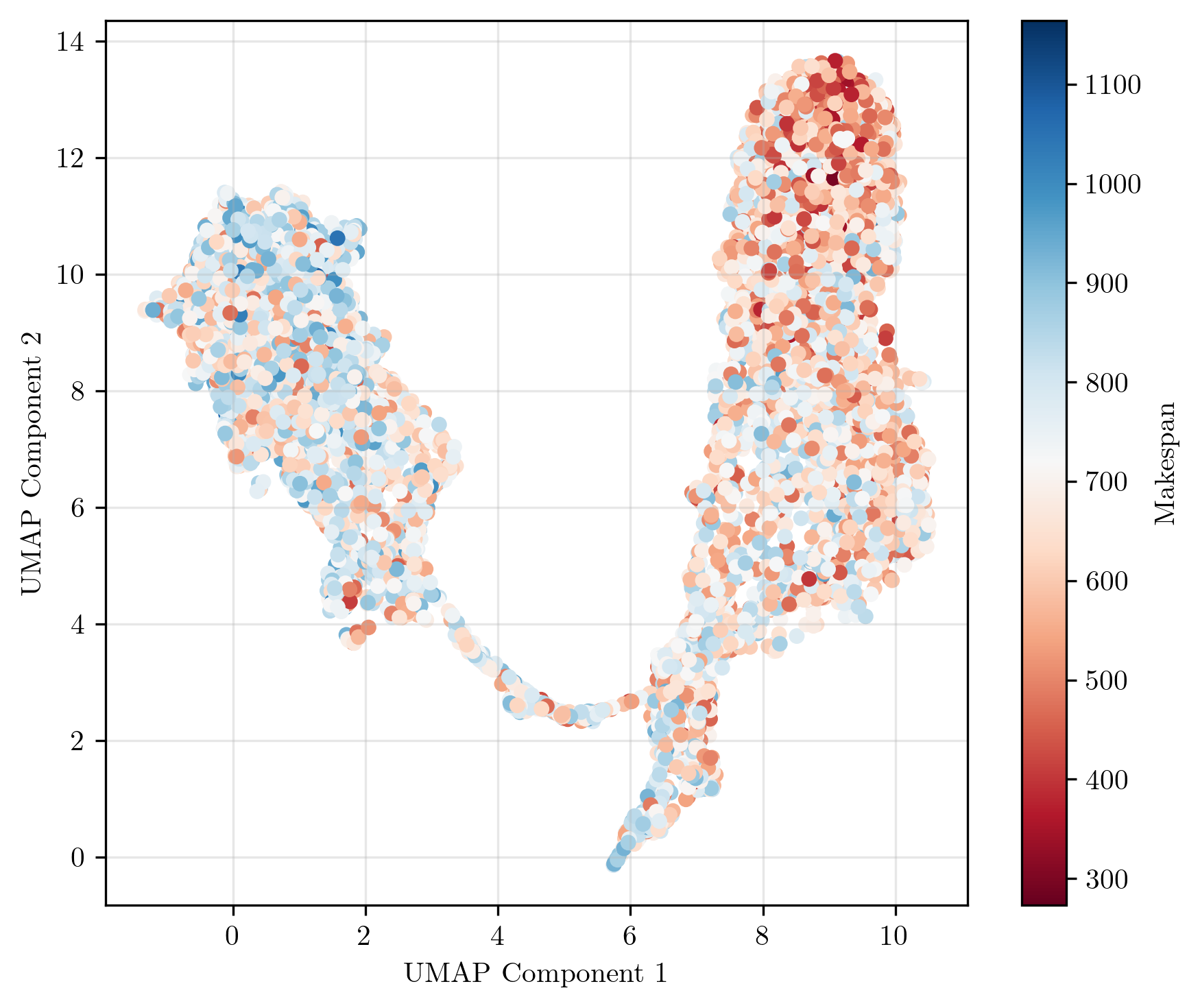}
        \caption{$\text{E2E w/o recon-loss}$, $\text{epoch} = 9,000$}
        \label{f}
    \end{subfigure}
    
    \vspace{1em}
    
    \textbf{Scaled Flowshop Index}
    
    \begin{subfigure}[b]{0.32\textwidth}
        \centering
        \includegraphics[width=\textwidth]{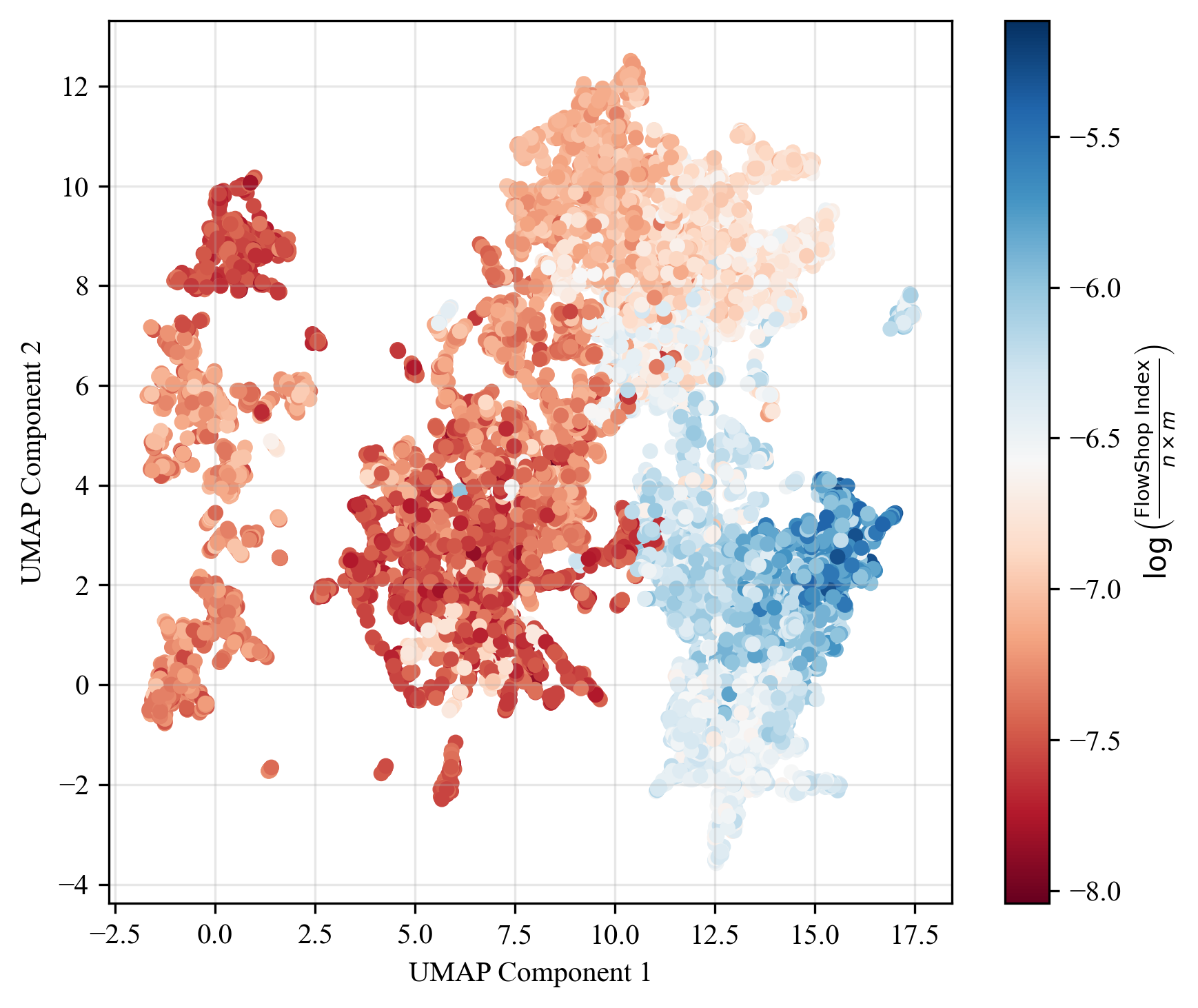}
        \caption{$E_r = 60,000$, $\text{epoch} = 0$}
        \label{g}
    \end{subfigure}
    \hfill
    \begin{subfigure}[b]{0.32\textwidth}
        \centering
        \includegraphics[width=\textwidth]{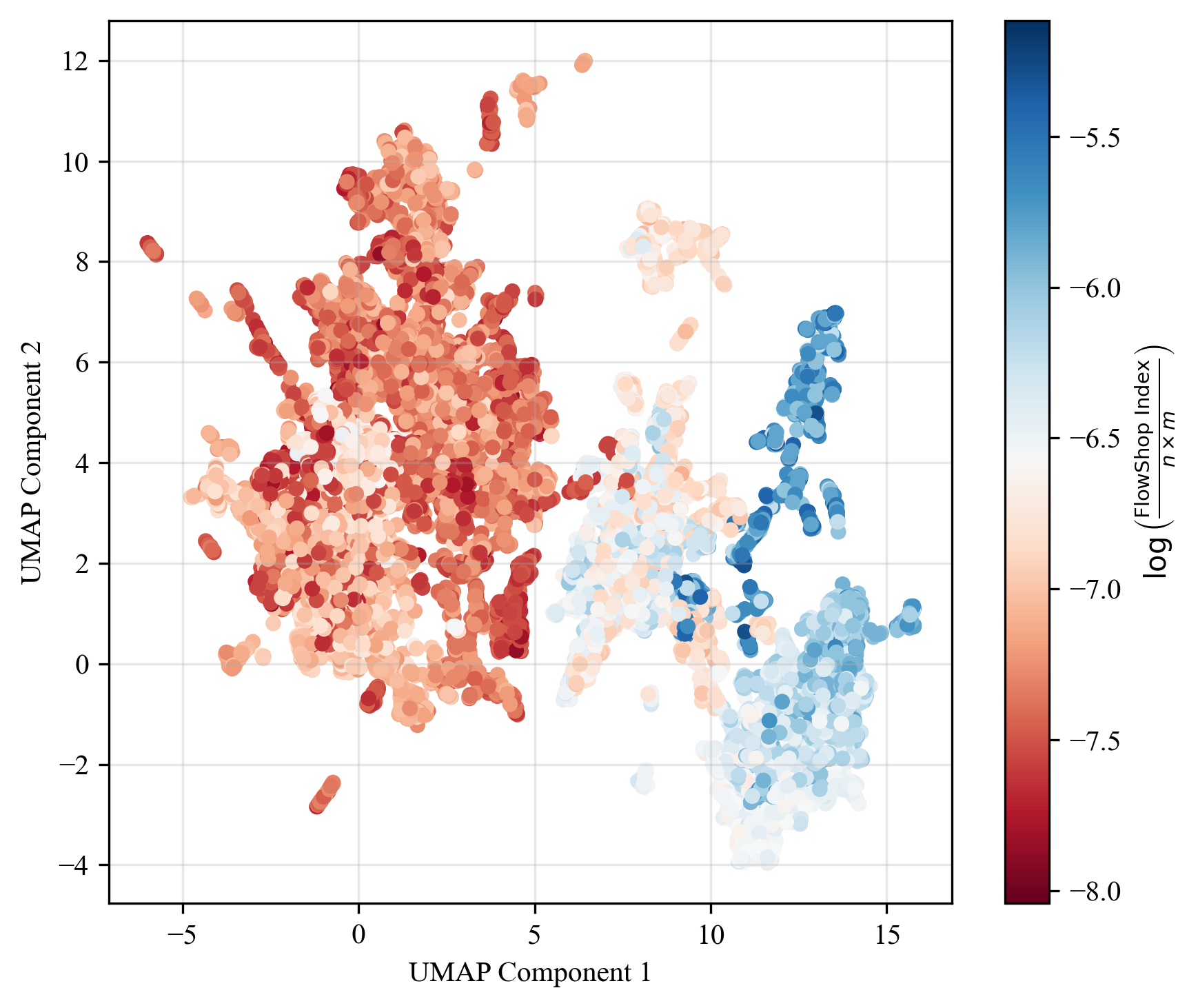}
        \caption{$E_r = 80,000$, $\text{epoch} = 6,000$}
        \label{h}
    \end{subfigure}
    \hfill
    \begin{subfigure}[b]{0.32\textwidth}
        \centering
        \includegraphics[width=\textwidth]{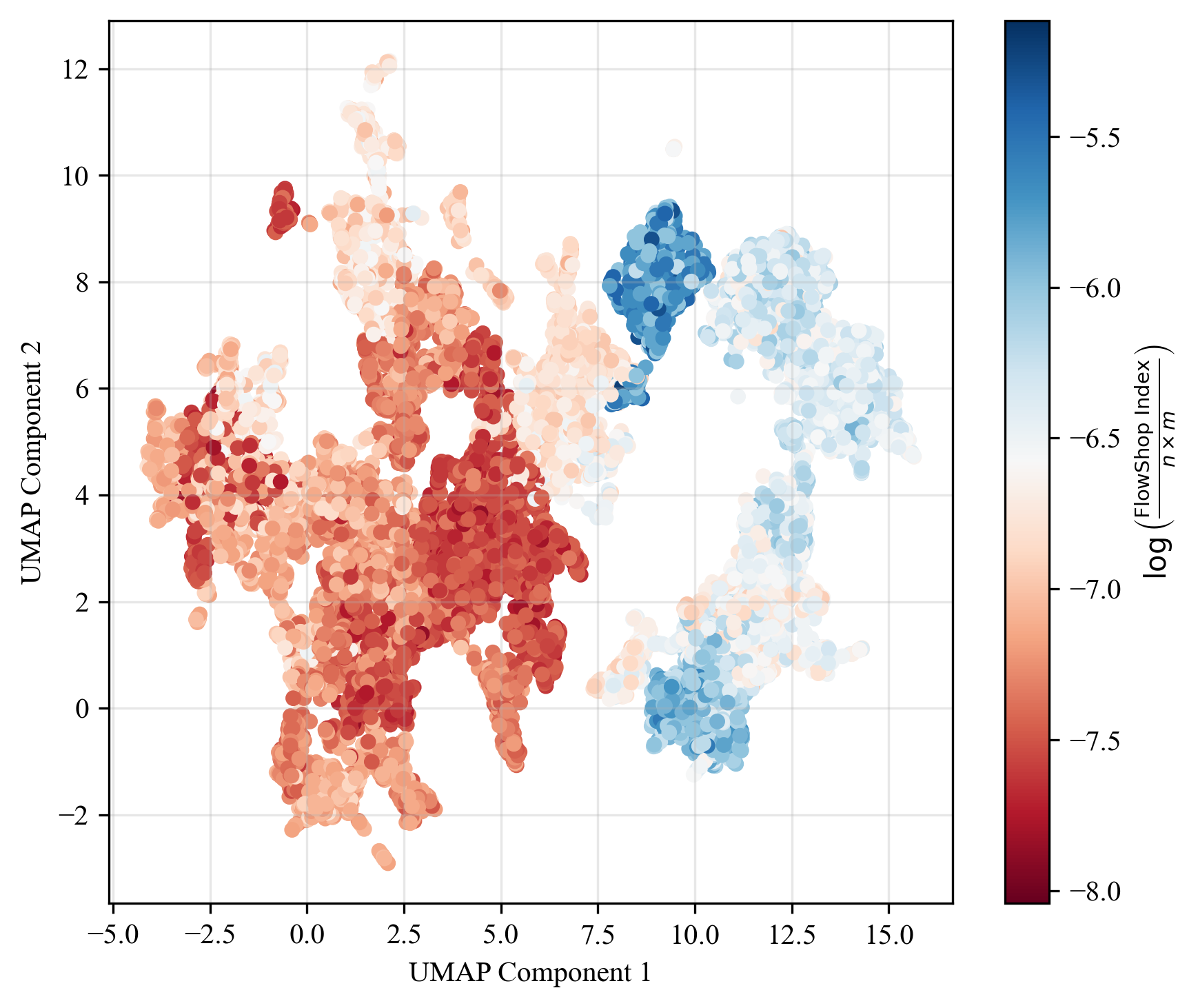}
        \caption{$E_r = 100,000$, $\text{epoch} = 9,000$}
        \label{i}
    \end{subfigure}
    
    \begin{subfigure}[b]{0.32\textwidth}
        \centering
        \includegraphics[width=\textwidth]{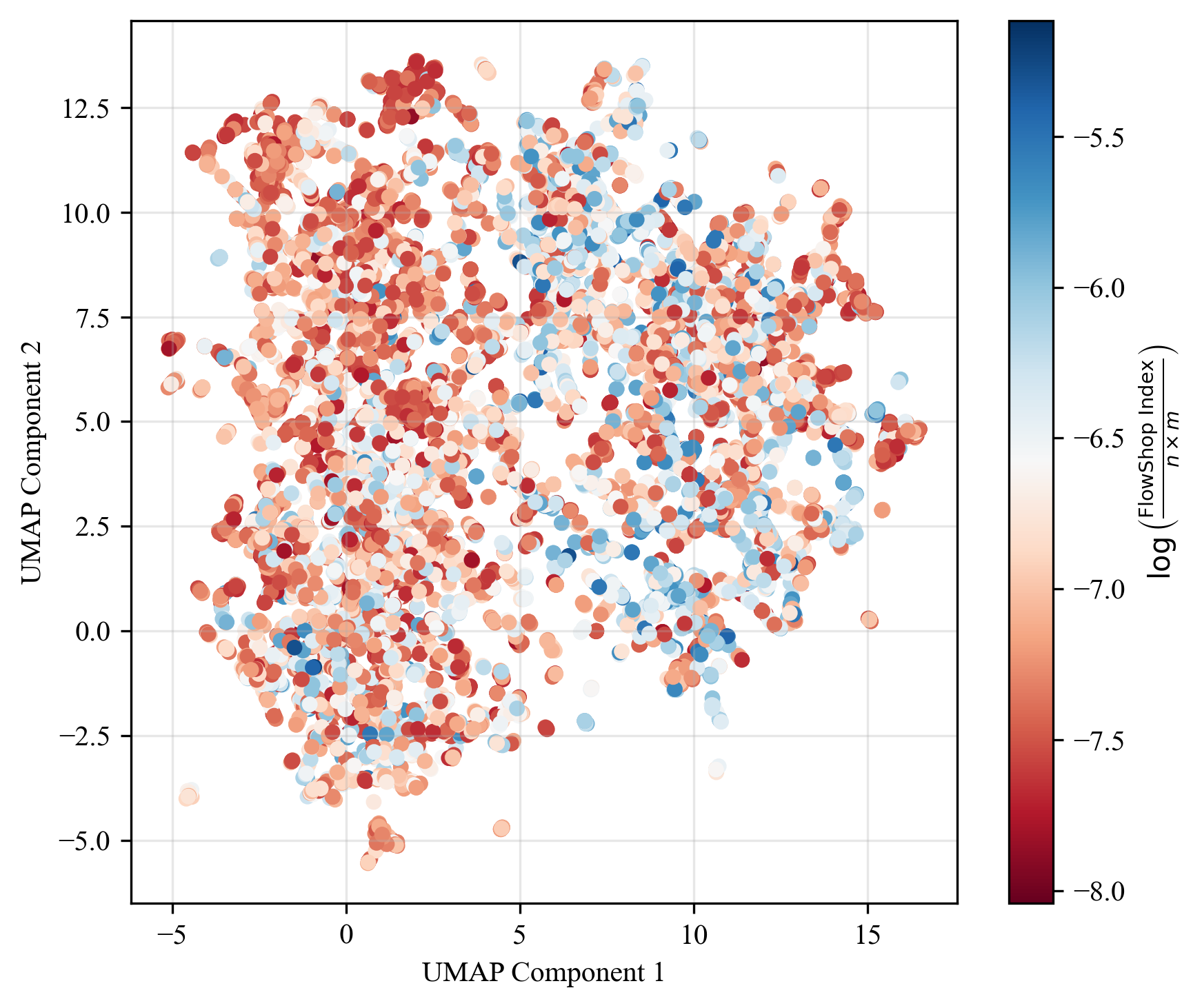}
        \caption{$\text{E2E w/o recon-loss}$, $\text{epoch} = 0$}
        \label{j}
    \end{subfigure}
    \hfill
    \begin{subfigure}[b]{0.32\textwidth}
        \centering
        \includegraphics[width=\textwidth]{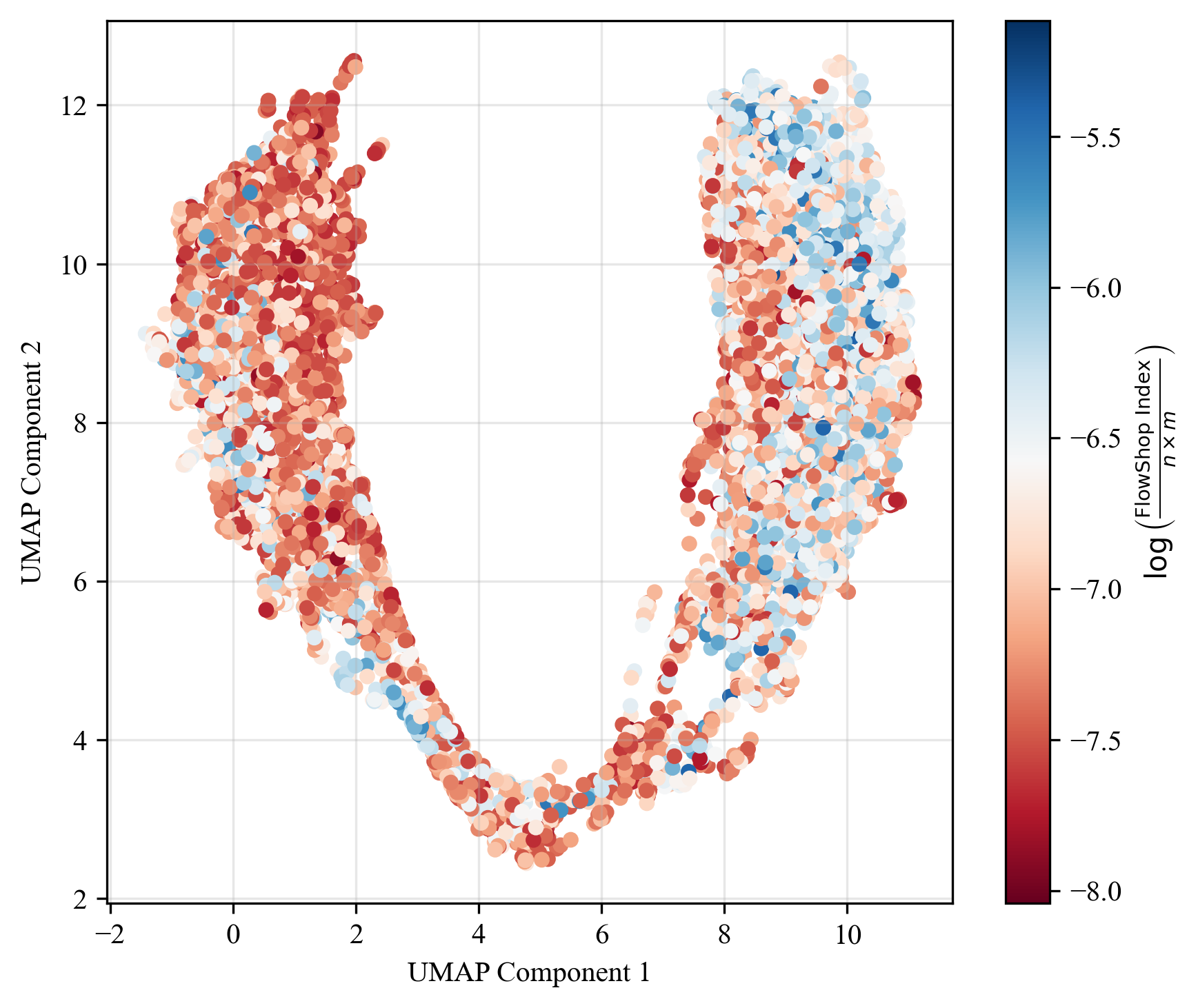}
        \caption{$\text{E2E w/o recon-loss}$, $\text{epoch} = 6,000$}
        \label{k}
    \end{subfigure}
    \hfill
    \begin{subfigure}[b]{0.32\textwidth}
        \centering
        \includegraphics[width=\textwidth]{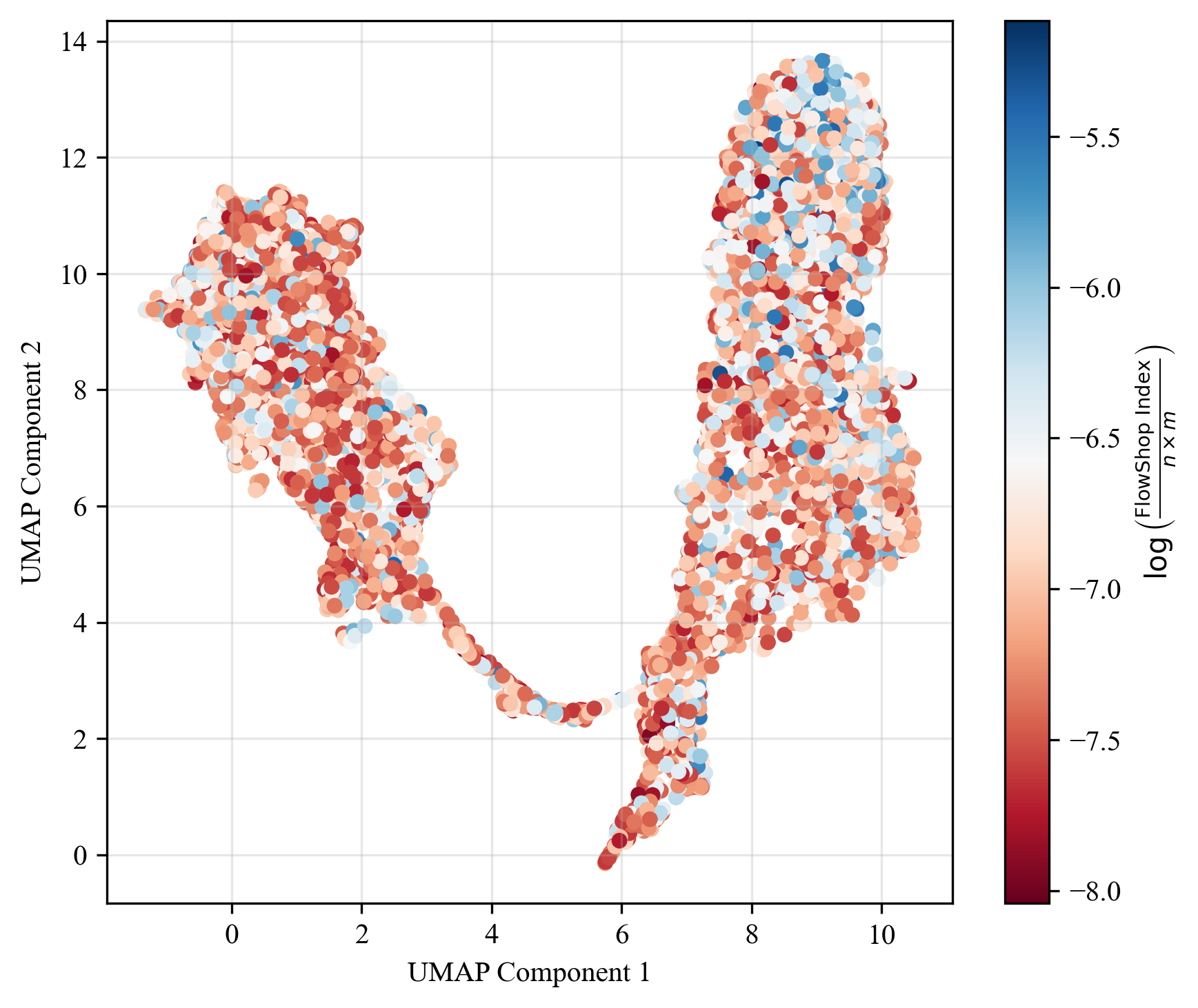}
        \caption{$\text{E2E w/o recon-loss}$, $\text{epoch} = 9,000$}
        \label{l}
    \end{subfigure}
    
    \caption{Latent space visualization using UMAP: Analysis of the learned manifold with points colored by makespan and scaled flowshop index}
    \label{umap_visualization}
\end{figure}

To further investigate how the proposed framework captures the structural \textcolor{black}{characteristics} of JSSP instances, we visualize the latent space using Uniform Manifold Approximation and Projection (UMAP). Fig.~\ref{umap_visualization} illustrates the 2D projection of $\mu_z$ at different training stages (epoch 0, 6,000, and 9,000), \textcolor{black}{where each of the 20,000 plotted points represents one JSSP instance, 
colored by its makespan or scaled flowshop index.} The formal definition of the scaled flowshop index is provided in Supplementary Material \textcolor{black}{K}. In this visualization, the makespan and scaled flowshop index serve as representative indicators of instance-specific characteristics. By coloring the points based on these metrics, we aim to demonstrate that the learned latent variables effectively capture and compress the essential structural features of each JSSP instance. \textcolor{black}{E2E without recon-loss is adopted as the baseline model, with identical hyperparameters applied as in VG2S.}

\textbf{Structural awareness at initial stage (epoch 0):} 
As shown in Fig.~\ref{a} and \ref{g}, the proposed VG2S (with $E_r = 60,000$) exhibits a highly organized pattern even at the beginning of policy training (epoch 0). Instances with similar makespans and scaled flowshop indices are positioned in close proximity within the latent space. In contrast, the baseline model (Fig.~\ref{d} and \ref{j}) displays a completely stochastic distribution without any discernible pattern. \textcolor{black}{This observation confirms that the variational graph encoder effectively captures the intrinsic topological properties of JSSP instances through reward-free representation learning, establishing a structural map prior to policy learning.}

\textbf{Evolution of latent space during training (epoch 6,000 and 9,000):} 
During the later stages of policy learning (epoch 6,000 and 9,000), VG2S \textcolor{black}{retains} these structural \textcolor{black}{patterns} (Fig.~\ref{b}, \ref{c}, \ref{h}, \ref{i}). The clear separation across different instance metrics demonstrates the representational power of the model for individual instances. \textcolor{black}{In particular, instances with similar makespans or scaled flowshop index are mapped to proximal regions in the latent space while those with dissimilar metrics are placed farther apart, and this geometric structure is retained throughout the progression of policy learning epochs and across different $E_r$ settings.} For the baseline model, a gradual transition in color becomes observable along UMAP Component 1 as training progresses (Fig.~\ref{e}, \ref{f}, \ref{k}), suggesting that the agent begins to extract some features from the \textcolor{black}{reward-based representation learning process}. However, this pattern is significantly weaker and more fragmented compared to the proposed VG2S. Notably, as shown in Fig.~\ref{l}, the baseline fails to form a clear structural representation of the scaled flowshop index even at epoch 9,000, suggesting that the latent space representation remains non-informative for JSSP instances. 
This suggests that the learning instability of the E2E without recon-loss, as discussed in Section~\ref{Ablation on variational representation learning}, may also stem from  \textcolor{black}{the failure of reward-driven representation learning to form a structured latent space, coupled with the continuous fluctuation of latent representations throughout training.}
\par
The superior \textcolor{black}{instance discriminability} of VG2S provides a fundamental explanation for its robust zero-shot generalization. By mapping instances into a structured latent space based on their structural resemblance, the agent can immediately identify the characteristics of unseen problems. This results in more stable and efficient policy learning compared to the baseline, which must attempt to \textcolor{black}{infer} structural information solely from sparse reward signals. To further substantiate the structural interpretability of the learned latent space, a quantitative analysis of the semantic structure of the learned latent dimensions via Spearman rank correlation is provided in Supplementary Material \textcolor{black}{M.} \textcolor{black}{Meanwhile, the consistent structured pattern is robustly observed across varying UMAP hyperparameter settings (Supplementary Material L).}

\clearpage
\clearpage

\subsection{\textcolor{black}{Interpretation} of agent strategy via PDR similarity}
\label{Analysis of agent strategy via PDR similarity}

\begin{figure}
    \centering
    \begin{subfigure}[b]{\textwidth}
        \centering
        \includegraphics[width=\textwidth]{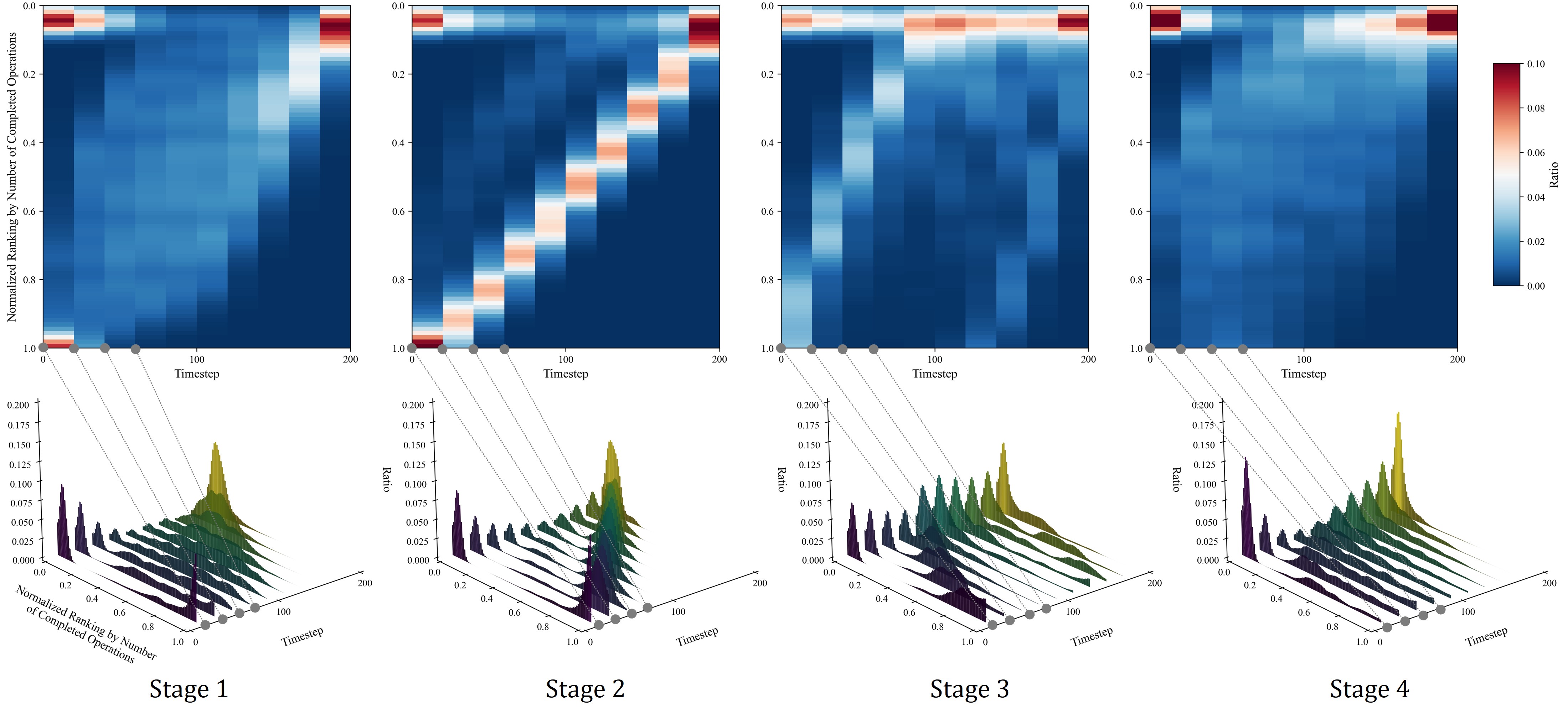}
        \caption{Rank by completed operations}
        \label{completed operations}
    \end{subfigure}
    \hfill
    \begin{subfigure}[b]{\textwidth}
        \centering
        \includegraphics[width=\textwidth]{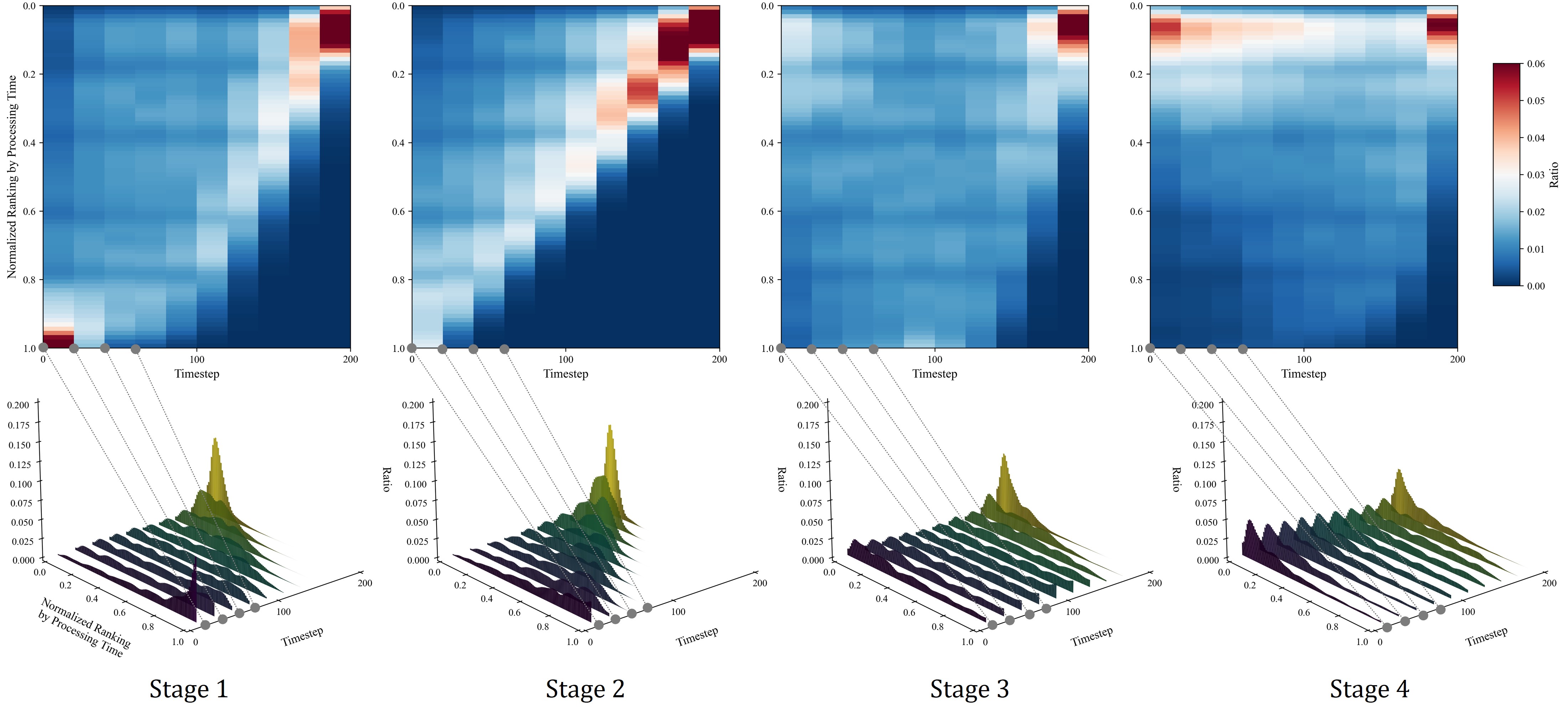}
        \caption{Rank by processing time}
        \label{processing time}
    \end{subfigure}
    \caption{Evolution of decision-making logic interpreted through normalized ranking metrics}
    \label{evolution}
\end{figure}

\textcolor{black}{To provide insight into the decision-making logic of the trained agent, we interpret its actions through the lens of representative PDRs}, specifically SPT and \textcolor{black}{MOR}. We track how the policy evolves across different training stages to reveal the underlying learning dynamics and the impact of variational representation. This analysis is conducted using the training process of hyperparameter ID 4 ($E_r = 80,000$) up to epoch 6,000. The evaluation is performed on 100 JSSP instances comprising $n \times m = 200$ operations, utilizing two normalized ranking metrics to quantify behavioral tendencies. The first metric, the normalized rank by the number of completed operations, measures the priority of an operation based on the progress of its parent job, where a rank closer to $0$ signifies a stronger resemblance to the \textcolor{black}{MOR} rule. The second metric, the normalized rank by processing time, identifies the relative standing of the chosen operation's duration, with a rank closer to $0$ representing alignment with the SPT rule. All rankings are normalized by the number of jobs $n$, as the maximum number of dispatchable operations at any given timestep is bounded by the total number of jobs. \textcolor{black}{To provide a comprehensive interpretation of the policy evolution, we utilize both 2D colormaps and 3D ridge plots, as illustrated in Fig.~\ref{evolution}. While the 2D colormaps effectively capture the temporal transitions and shifts in the agent’s strategy across the scheduling horizon, the 3D ridge plots offer a deeper insight into the modality and decision confidence at each individual timestep. To enhance visual clarity and mitigate local noise, the ranking metrics are averaged over 20-timestep intervals.}
\par
\textcolor{black}{\textbf{Stage 1 (before Phase 1): }At the very start, mixed patterns resembling \textcolor{black}{Least Operations Remaining (LOR), MOR}, and \textcolor{black}{Longest Processing Time} (LPT) intermittently emerge, driven primarily by the constrained action space rather than any understanding of the graph topology or resource constraints. As scheduling progresses into the early phase, \textcolor{black}{LOR} becomes momentarily more prominent while a faint LPT tendency persists. In the mid-phase, as the action space expands, the policy transitions into a near-random pattern. Toward the later phase, \textcolor{black}{MOR} and SPT patterns naturally emerge as a consequence of the contracting action space.
}
\par
\textcolor{black}{\textbf{Stage 2 (after Phase 1, before Phase 2): }A critical shift occurs in the second stage, immediately following the completion of Phase 1 (epoch 0). Although the policy decoder weights remain identical to the randomized state in Stage 1, the agent suddenly displays distinct and consistent tendencies toward LPT and \textcolor{black}{LOR}. The emergence of these organized patterns from a still-randomized decoder suggests that a meaningful structural change has occurred in $z$ during Phase 1 — a shift further corroborated by the UMAP analysis in Section~\ref{Visual analysis of latent representation via UMAP}. This demonstrates that the reorganization of the latent space through variational representation learning alone is sufficient to induce significant behavioral changes in the agent, even in the complete absence of any policy decoder adaptation.}
\par
\textcolor{black}{\textbf{Stage 3 (early Phase 2): }The third stage, corresponding to early Phase 2 (approximately epoch 200), is characterized by the agent beginning to explore the optimization landscape guided by reward signals. The previously dominant LPT tendency diminishes, resulting in a temporary increase in randomness with respect to processing time-based rules as the agent navigates toward the more efficient strategy that emerges in Stage 4. As demonstrated by the training stability of VG2S in Section~\ref{Ablation on variational representation learning}, this apparent randomness is not an unintended artifact but a guided exploration enabled by the informative latent representation.}
\par
\textcolor{black}{\textbf{Stage 4 (convergence of Phase 2): }In the final stage of mature Phase 2 (epoch 6,000+), the agent converges to a sophisticated, time-varying policy that dynamically shifts between behavioral tendencies resembling SPT and \textcolor{black}{MOR} depending on the specific scheduling state. For instance, the policy prioritizes job balancing in the initial and final stages while focusing on individual operation efficiency in the mid-phase to maximize machine turnover. Grounded in the stable and informative latent representation established in Phase 1, the agent ultimately arrives at a consistently effective strategy. A detailed analysis of the converged policy is provided in Supplementary Material N.}
\par
\textcolor{black}{These results demonstrate that the agent's capability extends far beyond the mere orchestration of explicit dispatching rules (e.g., SPT and MOR). Guided by the stationary latent map established in Phase 1, the agent's exploration is systematically steered toward increasingly refined policies rather than \textcolor{black}{aimless} trial-and-error. This interpretation does not suggest that the agent's behavior reduces to a simple combination of PDRs, as VG2S develops a sophisticated, context-aware strategy that dynamically shifts beyond SPT and MOR across different scheduling stages. Nevertheless, these interpretive lenses qualitatively characterize and explain the agent's decision-making logic, providing a foundation for understanding how VG2S arrives at its scheduling decisions.}

\subsection{Computational complexity}
\label{Computational complexity}
\textcolor{black}{
This section examines the computational complexity of VG2S from three complementary perspectives. We first characterize its theoretical and empirical complexity (Section~\ref{theo_complexity}), then compare its inference time against PDRs (Section~\ref{sec_complexity}), and finally assess its solution quality-to-computational cost relative to metaheuristic approaches (Section~\ref{comparison with meta}).}

\subsubsection{Theoretical and empirical complexity}
\label{theo_complexity}
The theoretical inference complexity of VG2S is $\mathcal{O}(n^3m)$, dominated by the look-ahead state feature computation ($s^{\cdot}_{4,\cdot}$, $s^{\cdot}_{5,\cdot}$, $s^{\cdot}_{6,\cdot}$); a detailed component-wise derivation is provided in  Supplementary Material \textcolor{black}{O}. Fig.~\ref{fig:empirical_complexity} empirically validates and refines this analysis. Measured inference times across a wide range of instance sizes are plotted against four candidate complexity curves: $\mathcal{O}(nm)$, $\mathcal{O}(n^2m)$, $\mathcal{O}(n^3m)$, and $\mathcal{O}(n^2m^2)$. The observed computation times align most closely with the $\mathcal{O}(n^2m)$ curve, which lies strictly below the theoretical worst case. This gap is explained by the shrinking action space during scheduling: while $|\mathcal{A}(s_t)|$ is bounded by $n$ in the worst case, the number of available operations decreases progressively as the schedule is constructed, substantially reducing the effective cost per timestep below the theoretical worst case.

\begin{figure}
    \centering
    \includegraphics[width=0.6\textwidth]{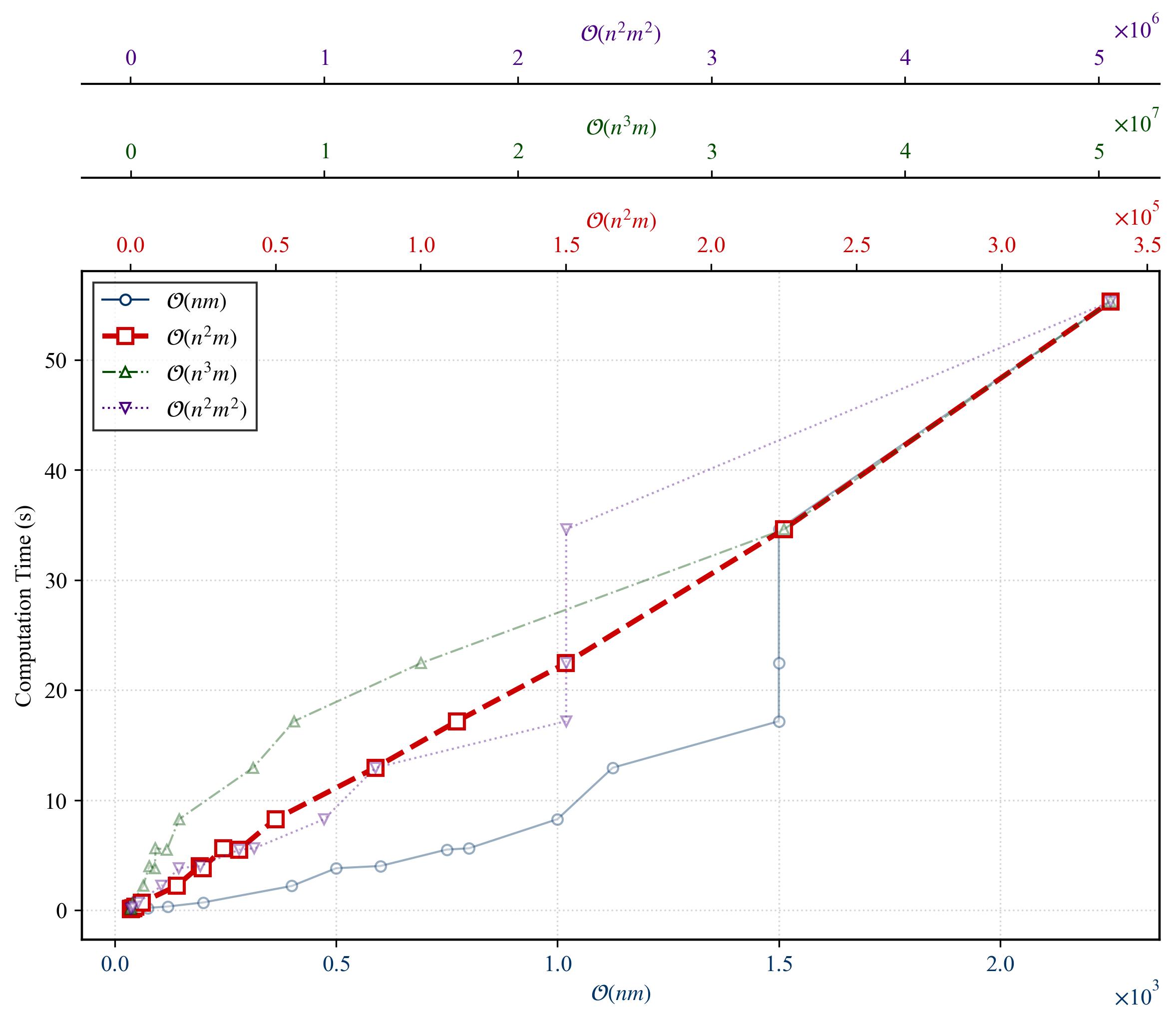}
    \caption{Empirical computation time across various problem sizes.}
    \label{fig:empirical_complexity}
\end{figure}

\subsubsection{Training and inference time comparison with PDRs}
\label{sec_complexity}
\textcolor{black}{
To compare inference time, we measure the average optimality gap ($\%$) and computation time (s) for VG2S, VG2S-lite, and three PDR baselines (SPT, MWKR, and MOR) across TA and DMU benchmark instances, as reported in Table~\ref{opt_gap_computation_time}. VG2S-lite (a configuration with reduced complexity of $\mathcal{O}(n^{2}m)$ introduced in Section~\ref{look_ahead}) is considered alongside VG2S to assess the practical deployment trade-off of the proposed framework. For both configurations, the model checkpoints selected based on the best average validation performance across TA61, TA62, DMU76, and DMU77 from the experiments in Section~\ref{look_ahead} are used. Computation times are averaged over five runs. The VG2S model used in this section requires 3,502\,s ($E_r = 80,000$) for Phase 1 and 11,695\,s for Phase 2, totaling approximately 15,197\,s for training. VG2S-lite requires 2,740\,s ($E_r = 60,000$) for Phase 1 and 8,805\,s for Phase 2, totaling approximately 11,545\,s for training. } 
\par
\textcolor{black}{As shown in Table~\ref{opt_gap_computation_time}, VG2S substantially outperforms all PDR baselines in solution quality. However, the more critical observation here concerns computation time. Unlike the PDR baselines, which operate in near-real-time regardless of instance size, VG2S incurs non-trivial overhead that grows with problem scale: at the $100\times20$ level, inference requires approximately 31 seconds per instance, while VG2S-lite reduces this to approximately 19 seconds (approximately $1.6\times$ faster, with a range of $1.3$--$1.7\times$ across all instance sizes), substantially alleviating the latency overhead, though still short of near-real-time PDR throughput.}

\textcolor{black}{
To contextualize these numbers in terms of real-world deployability, we examine them relative to the scheduling time horizon. In scheduling environments where decisions are made on a daily or hourly basis (such as static or shift-level planning), an inference time of up to 31 seconds for VG2S remains practically insignificant. In dynamic settings, however, where scheduling decisions must be made on a minute-by-minute basis, an inference time of 31 seconds may no longer be acceptable. While VG2S-lite offers a substantial improvement, it nevertheless remains insufficient for real-time scheduling requirements. Nevertheless, both configurations offer substantial solution-quality advantages over PDR baselines, achieved through zero-shot generalization without requiring additional training per instance. To further examine the real-world deployability of VG2S and VG2S-lite, the following section presents a direct comparison against metaheuristic approaches.}

\begin{table}[htbp]
  \centering
  \footnotesize
  \caption{\textcolor{black}{Average opt gap (\%) and computation time (s) by instance size}}
  \label{opt_gap_computation_time}
  {\color{black}\begin{tabular}{cccccc|cccccc}
    \toprule
    \multicolumn{6}{c|}{(a) TA Benchmark} & \multicolumn{6}{c}{(b) DMU Benchmark} \\
    \midrule
    Size & SPT & MWKR & MOR & VG2S & VG2S-lite & Size & SPT & MWKR & MOR & VG2S & VG2S-lite \\
    \midrule
    $15\times15$ & \makecell{30.93 \\ (0.019)} & \makecell{22.58 \\ (0.023)} & \makecell{23.71 \\ (0.019)} & \makecell{\textbf{15.66} \\ (0.759)} & \makecell{17.66 \\ (0.511)} & $20\times15$ & \makecell{34.23 \\ (0.029)} & \makecell{30.26 \\ (0.036)} & \makecell{36.32 \\ (0.029)} & \makecell{\textbf{19.73} \\ (1.101)} & \makecell{22.44 \\ (0.804)} \\[4pt]
    $20\times15$ & \makecell{38.50 \\ (0.027)} & \makecell{25.08 \\ (0.034)} & \makecell{29.67 \\ (0.028)} & \makecell{\textbf{15.37} \\ (1.113)} & \makecell{22.41 \\ (0.700)} & $20\times20$ & \makecell{36.66 \\ (0.042)} & \makecell{29.51 \\ (0.053)} & \makecell{30.94 \\ (0.042)} & \makecell{\textbf{20.31} \\ (1.447)} & \makecell{23.70 \\ (1.080)} \\[4pt]
    $20\times20$ & \makecell{30.41 \\ (0.041)} & \makecell{22.50 \\ (0.051)} & \makecell{24.44 \\ (0.039)} & \makecell{\textbf{16.30} \\ (1.520)} & \makecell{20.45 \\ (0.950)} & $30\times15$ & \makecell{33.05 \\ (0.052)} & \makecell{36.07 \\ (0.066)} & \makecell{39.57 \\ (0.052)} & \makecell{\textbf{22.75} \\ (2.092)} & \makecell{23.09 \\ (1.425)} \\[4pt]
    $30\times15$ & \makecell{36.76 \\ (0.051)} & \makecell{28.35 \\ (0.062)} & \makecell{23.48 \\ (0.051)} & \makecell{\textbf{19.32} \\ (2.131)} & \makecell{21.24 \\ (1.481)} & $30\times20$ & \makecell{38.73 \\ (0.079)} & \makecell{36.23 \\ (0.097)} & \makecell{37.85 \\ (0.079)} & \makecell{\textbf{23.44} \\ (2.932)} & \makecell{23.90 \\ (1.988)} \\[4pt]
    $30\times20$ & \makecell{38.43 \\ (0.075)} & \makecell{27.45 \\ (0.093)} & \makecell{30.00 \\ (0.074)} & \makecell{\textbf{20.62} \\ (2.975)} & \makecell{23.01 \\ (2.023)} & $40\times15$ & \makecell{31.08 \\ (0.079)} & \makecell{28.90 \\ (0.101)} & \makecell{38.21 \\ (0.080)} & \makecell{20.96 \\ (3.620)} & \makecell{\textbf{19.96} \\ (2.236)} \\[4pt]
    $50\times15$ & \makecell{25.16 \\ (0.107)} & \makecell{18.47 \\ (0.136)} & \makecell{19.58 \\ (0.109)} & \makecell{\textbf{11.76} \\ (5.818)} & \makecell{15.04 \\ (3.473)} & $40\times20$ & \makecell{36.93 \\ (0.118)} & \makecell{35.49 \\ (0.154)} & \makecell{40.28 \\ (0.121)} & \makecell{26.10 \\ (4.984)} & \makecell{\textbf{24.66} \\ (3.327)} \\[4pt]
    $50\times20$ & \makecell{32.68 \\ (0.160)} & \makecell{19.99 \\ (0.202)} & \makecell{19.90 \\ (0.162)} & \makecell{\textbf{13.01} \\ (7.906)} & \makecell{14.63 \\ (4.942)} & $50\times15$ & \makecell{27.95 \\ (0.110)} & \makecell{28.71 \\ (0.151)} & \makecell{36.23 \\ (0.115)} & \makecell{17.52 \\ (5.446)} & \makecell{\textbf{15.67} \\ (3.459)} \\[4pt]
    $100\times20$ & \makecell{17.33 \\ (0.435)} & \makecell{9.53 \\ (0.593)} & \makecell{11.10 \\ (0.459)} & \makecell{\textbf{6.95} \\ (31.317)} & \makecell{7.13 \\ (19.858)} & $50\times20$ & \makecell{33.60 \\ (0.167)} & \makecell{33.73 \\ (0.240)} & \makecell{38.40 \\ (0.170)} & \makecell{26.88 \\ (7.826)} & \makecell{\textbf{22.06} \\ (4.848)} \\
    \bottomrule
  \end{tabular}}
\end{table}
\subsubsection{Solution quality and computation time comparison with metaheuristic approaches}
\label{comparison with meta}
\textcolor{black}{
While PDRs offer near-real-time responsiveness, metaheuristic approaches are more commonly adopted for scheduling problems where solution quality is the primary concern. A comparison against such methods therefore provides a more rigorous assessment of the solution quality-time efficiency of VG2S. To this end, we examine whether VG2S and VG2S-lite can remain competitive against such methods given a generous time budget. We compare VG2S and VG2S-lite against three metaheuristic baselines: 
Genetic Algorithm and Iterative Local Search (GAILS)~\cite{shao2024gails}, GAILS-Relative-Urgency Based Initialization (RUBI)~\cite{shao2024gails,baek2025rubi}, and 
Tabu-RUBI~\cite{baek2025rubi}, which perform well in JSSP. Google OR-Tools CP-SAT, one of the most 
competitive exact solvers for combinatorial optimization, is additionally included 
as a solution quality reference. Two JSSP datasets sourced from real-world 
manufacturing environments are used, each comprising 30 instances: a Part Shop 
(PS)~\cite{smaili2025simulation} of size $140 \times 5$, and a shipyard Panel 
Block Assembly Line (PBAL)~\cite{cho2022minimize} of size $100 \times 6$\footnote{\textcolor{black}{The PBAL dataset is an instance of the permutation flow shop problem, 
in which all jobs share an identical machine visiting order. 
In this work, we interpret it as a JSSP instance in which 
all jobs share the same machine sequence.}}. 
All metaheuristic baselines and OR-Tools CP-SAT are given a time limit of 
$300 \times \bar{t}_{\text{VG2S-lite}}$, yielding $2{,}370\,\text{s}$ for PS 
and $1{,}140\,\text{s}$ for PBAL, with early stopping applied once optimality 
is certified. VG2S-lite produces its schedule in a single 
forward pass with no additional overhead. Solution quality is measured as the 
OR-tools gap, i.e., the percentage deviation from the CP-SAT reference. Further 
implementation details for metaheuristic baselines are provided in Supplementary Material P.}
\par
\textcolor{black}{Fig.~\ref{versus metaheuristic} summarizes the results. On PS, OR-Tools proves 
optimality in under 1 second for all instances. VG2S-lite achieves $0.60\%$ gap 
in just $7.9\,\text{s}$, and VG2S achieves $1.74\%$ gap in $11.7\,\text{s}$, 
with VG2S-lite marginally outperforming VG2S on this dataset. Among the metaheuristic baselines, GAILS attains the optimal solution 
($0.00\%$ gap) and GAILS-RUBI attains a near-optimal gap of $0.15\%$, 
both achieving superior solution quality to VG2S and VG2S-lite. However, 
these results are obtained by consuming $1{,}157\,\text{s}$ and 
$1{,}693\,\text{s}$, respectively (approximately $146\times$ and $214\times$ 
the computation budget of VG2S-lite), whereas VG2S and VG2S-lite achieve 
competitive solution quality in $11.7\,\text{s}$ and $7.9\,\text{s}$, 
respectively. Tabu-RUBI yields an inferior gap of $2.36\%$, requiring 
$2{,}368\,\text{s}$ (approximately $299\times$ the computation budget of 
VG2S-lite) to reach its solution quality.}

\textcolor{black}{On PBAL, OR-Tools fails to prove optimality within its time limit, reflecting 
the inherent difficulty of these instances. Unlike PS, VG2S now outperforms 
VG2S-lite, achieving a negative average gap of $-0.94\%$ (strictly 
outperforming even OR-Tools in solution quality) in only $4.8\,\text{s}$. 
VG2S-lite follows closely at $0.58\%$ gap in $3.8\,\text{s}$. The metaheuristic 
baselines, by contrast, deteriorate substantially compared to PS, with gaps of 
$41.08\%$, $45.10\%$, and $40.40\%$ for GAILS-RUBI, GAILS, and Tabu-RUBI, 
despite consuming $1{,}183\,\text{s}$ on average, confirming 
the exceptional difficulty of PBAL instances. These results highlight the 
computational efficiency of both VG2S configurations relative to their solution 
quality, particularly on challenging instances where metaheuristic approaches 
struggle most.}

\textcolor{black}{Collectively, these results reveal a compelling efficiency advantage of VG2S 
and VG2S-lite over metaheuristic baselines. On the relatively tractable PS 
dataset, metaheuristic baselines achieve comparable solution quality but require 
at least $146\times$ (GAILS), $214\times$ (GAILS-RUBI), to $299\times$ (Tabu-RUBI) more computation than 
VG2S-lite. On the more challenging PBAL dataset, despite consuming over 
$300\times$ the computation budget, metaheuristic baselines still fail to match 
the solution quality of VG2S. While neither configuration achieves real-time 
throughput, both VG2S and VG2S-lite demonstrate a substantially more favorable 
speed-quality trade-off compared to metaheuristic approaches, underscoring their 
practical deployability in real-world manufacturing environments.}

\begin{figure}[htbp]
    \centering
    \includegraphics[width=\textwidth]{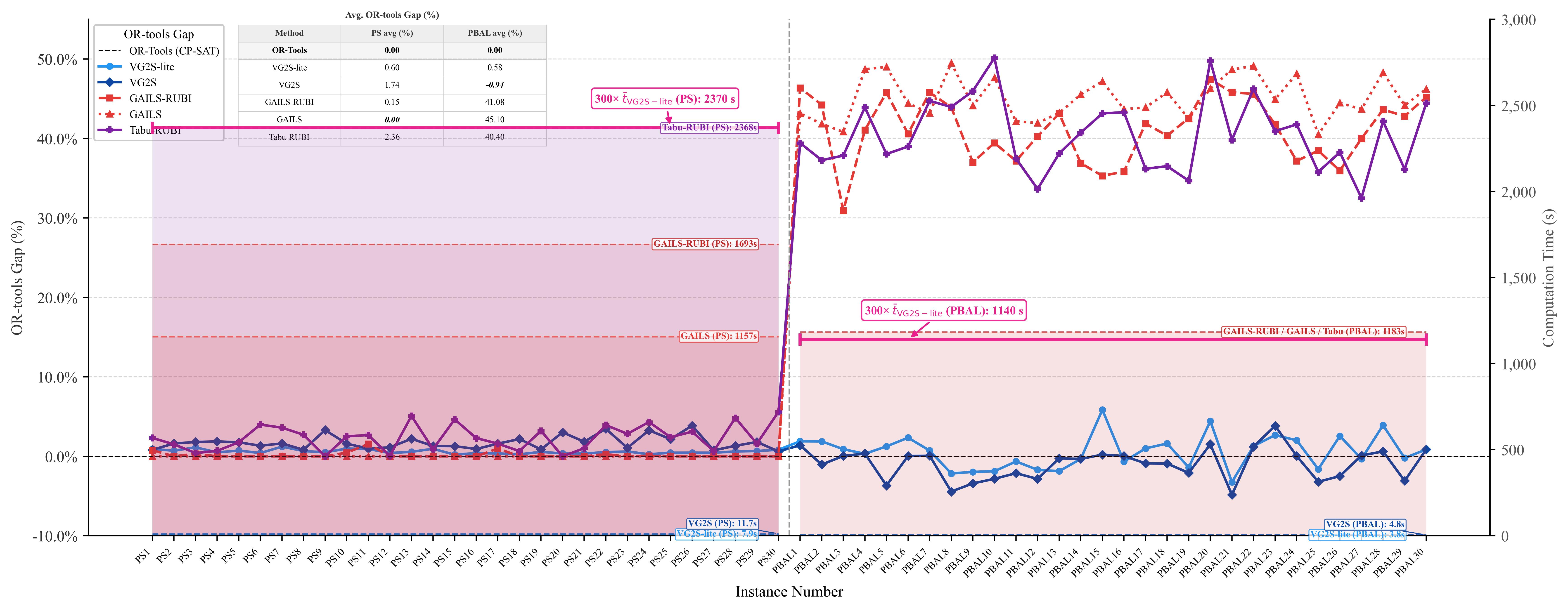}
    \caption{\textcolor{black}{Performance comparison of VG2S, VG2S-lite, and metaheuristic 
    baselines in terms of solution quality (OR-tools gap, left axis) and 
    computation time (right axis) across PS and PBAL instances.}}
    \label{versus metaheuristic}
\end{figure}

\clearpage 

\section{Discussion}

We discuss the key factors underlying the observed performance of VG2S:
\begin{enumerate}

\item \textcolor{black}{\textbf{Probabilistic robustness against distributional shifts}}: \textcolor{black}{Section~\ref{ELBO with maximum entropy RL for JSSP} explicitly incorporates the instance distribution $p(G)$ into the objective function, providing a principled probabilistic foundation for generalization across diverse problem instances. In end-to-end approaches, representation learning is absorbed into the reward-driven optimization process. Since the reward signal in JSSP carries only the objective function value (i.e., makespan), it conveys little about the structural properties of the problem instance itself, despite being informative for policy optimization. Consequently, the learned representations are confined to information sufficient for policy improvement, making them implicitly specialized to the training distribution and vulnerable to distributional shifts. In contrast, the ELBO explicitly encourages the latent space to encode the structural properties of JSSP instances by requiring the latent representation to preserve and reconstruct the original graph structure. This yields semantically meaningful representations that reflect the underlying topological characteristics of the problem, enabling generalization beyond the training distribution. The superior zero-shot performance on DMU and SWV, reported in Section~\ref{Generalization and scalability test}, provides direct empirical evidence of this robustness.}

\item \textcolor{black}{\textbf{Stable training via two-phase strategy}}: \textcolor{black}{The two-phase strategy freezes the learned latent representation, which may appear to underutilize the theoretical implications of Section~\ref{ELBO with maximum entropy RL for JSSP}, seemingly introducing a representational bottleneck. However, freezing the representation provides a stable foundation for the critic. Rather than simultaneously tracking a moving latent representation, the critic can dedicate its entire capacity to refining value prediction under the fixed representation, thereby progressively enhancing advantage estimation and promoting smooth convergence. This suggests that the benefit of allowing the latent space to adapt to the policy's needs through joint optimization is marginal compared to the stability gained by the two-phase strategy. The ablation results in Section~\ref{Effectiveness of variational representation learning} empirically corroborate this, demonstrating that freezing the representation consistently improves training stability and solution quality.}

\end{enumerate}

\textcolor{black}{Two questions merit careful examination: (1) whether the observed performance gains are primarily attributable to the two-phase strategy itself, independent of the ELBO-based objective, and (2) whether look-ahead features constitute the primary driver of improvements. Each is addressed in turn below.}

\begin{enumerate}

\item \textcolor{black}{\textbf{Necessity of ELBO-based two-phase strategy:}
While a two-phase strategy separating representation learning and policy learning is a widely adopted practice in RL, one might expect that it alone (without the ELBO-based variational objective) is sufficient to achieve the observed gains. However, as demonstrated in Section~\ref{Ablation on variational representation learning}, the deterministic pre-training baseline consistently resulted in numerical divergence across all eight experimental configurations. Since $\mathcal{L}_{\text{KL}}$ is removed, the magnitude of $\mu_z$ is no longer penalized, potentially allowing $\mu_z$ to grow unboundedly. This confirms that $\mathcal{L}_{\text{KL}}$ serves as a fundamental stability mechanism beyond probabilistic regularization. The two-phase strategy therefore achieves its full effectiveness only in synergy with the ELBO-based variational objective, and neither component alone is sufficient.}

\item \textcolor{black}{\textbf{Look-ahead features}: While look-ahead features yield clear gains on TA instances, their removal in VG2S-lite actually improves performance on DMU benchmarks. This discrepancy may be attributed to the instance-dependent predictive power of look-ahead features: the structural barrier between the two machine subsets in DMU may weaken their effectiveness, as operations in the second subset cannot meaningfully interact with current decisions until the first subset is fully processed. In contrast, on TA instances where machine sequences are completely random and structurally similar to the training distribution, look-ahead signals remain highly informative throughout the scheduling process. This instance-dependent behavior empirically suggests that look-ahead features are not the primary driver of VG2S's overall performance gains.}
\end{enumerate}

\textcolor{black}{Beyond these core attribution factors related to performance, two additional observations merit further discussion regarding the plug-in representation modularity of the variational graph encoder.}

\begin{enumerate}
\item \textcolor{black}{\textbf{Plug-in modularity of the variational graph encoder: 
cross-framework compatibility}
The variational graph encoder is naturally suited for plug-in use across diverse neural network-based frameworks. This is because $z$ of VG2S is a time-independent, instance-level representation computed once prior to scheduling. This compatibility stems from the fact that existing DRL frameworks define their decision timestep in fundamentally different ways: \cite{park2021learning} adopts a semi-Markov Decision Process formulation where only non-trivial states (i.e., moments when a dispatching decision is actually required) constitute transitions; 
\cite{yuan2023solving} advances time in a discrete-event simulation manner, advancing to the next machine availability; and VG2S simply indexes each operation selection sequentially. Despite these differences in how decision timestep is defined, $z$ remains compatible with all of them; it can be seamlessly combined with diverse DRL frameworks via simple input modification such as concatenation ($[z \| s_t]$), requiring no architectural modifications beyond the input level.}
\item \textcolor{black}{\textbf{The scope and limits of plug-in modularity of the variational graph encoder}
While the plug-in modularity of the variational graph encoder is broadly 
applicable within the GNN-DRL paradigm, its applicability beyond this 
paradigm is more nuanced. SLIM~\cite{corsini2024self}, RASCL~\cite{iklassov2023study}, 
DRL-guided improvement search~\cite{zhang2024deep}, and MAMBA-based 
approaches~\cite{cao2026mamba} present plausible integration pathways, 
as their architectures can accommodate the instance-wise latent representation 
$z$ of VG2S; for instance, $z$ can be concatenated to SLIM's decoder input, 
RASCL's actor input, \cite{zhang2024deep}'s node embeddings, and MAMBA's 
sequence input as a global conditioning signal, respectively. In contrast, 
LLM-based approaches such as SeEvo~\cite{huang2026automatic} cannot incorporate 
continuous latent representations due to their reliance on commercial black-box 
LLMs that operate on token-level inputs without access to intermediate latent 
spaces. The empirical validation of such cross-paradigm integration constitutes a promising direction for future research.}
\end{enumerate}

\section{Conclusion}

In this study, we proposed the VG2S framework, which introduces variational \textcolor{black}{representation learning} to the JSSP domain for the first time. By deriving a probabilistic objective based on the ELBO with maximum entropy DRL, VG2S mathematically and structurally decouples representation learning from policy learning. \textcolor{black}{The two-phase training strategy fully leverages the effectiveness of ELBO-based representation learning by preventing the moving target problem.} Extensive experiments validate \textcolor{black}{the proposed framework}, demonstrating superior zero-shot generalization and robust scalability across diverse and challenging benchmark instances. UMAP analysis confirms that the variational graph encoder successfully compresses complex disjunctive graph structures into a structured latent manifold prior to policy learning. PDR similarity \textcolor{black}{interpretation} further reveals that the trained agent develops a sophisticated, context-aware strategy that cannot be fully explained by any single dispatching rule. \textcolor{black}{VG2S demonstrates a substantially more favorable speed-quality trade-off compared to 
metaheuristic approaches, particularly through VG2S-lite, which further reduces computational cost while maintaining competitive solution quality.}
\par
\textcolor{black}{However, while VG2S handles instance-level uncertainty, the stochasticity of state transitions (such as processing time fluctuations) is not considered in the present work and remains a limitation. In particular, the application of VG2S to real-world manufacturing configurations (including dynamic job arrivals, stochastic processing times, heterogeneous machine environments, and frequent disruptions) is not addressed in the present work. Validating VG2S under such operational conditions would substantially enhance its real-world applicability. This accordingly constitutes an important direction for future research.} 
\par
\textcolor{black}{Furthermore, the plug-in modularity of the variational graph encoder suggests its potential for cross-paradigm transfer, including MAMBA-based approaches \cite{cao2026mamba}, SLIM~\cite{corsini2024self}, RASCL~\cite{iklassov2023study}, and RL-guided search approaches \cite{zhang2024deep}, 
and remains an open direction for future research.} \textcolor{black}{Comparisons against alternative self-supervised pre-training strategies such as contrastive learning objectives would provide further empirical grounding for the design choices of VG2S, and are left as directions for future work. Finally, the interpretation in Section~\ref{Analysis of agent strategy via PDR similarity} points to the possibility of approximating VG2S's decision-making logic through lightweight alternatives such as learning to dynamically combine simple PDRs, and empirically validating this hypothesis remains a promising direction for future work.}

\section*{Acknowledgement}
This work was supported by the National Research Foundation of Korea (NRF) Grant funded by the Korean Government (MSIT) under Grant RS-2025-00555741; the Korea Planning \& Evaluation Institute of Industrial Technology (KEIT) Grants funded by the Korean Government (Ministry of Trade, Industry and Resources, MOTIR) under Grants RS-2026-25614399 and RS-2025-02372996; the Korea Industrial Complex Corporation (KICOX) Grant funded by the Korean Government (MOTIR) under Grant SG20251301; the Korea Institute for Advancement of Technology (KIAT) Grants funded by the Korean Government (MOTIR) under Grants RS-2025-02263945 and RS-2023-KI002688; and the 2026 Academic Research Project Fund of the Republic of Korea Navy, Maritime Institute Naval Academy.
\bibliographystyle{elsarticle-num}
\bibliography{cas-refs}

\end{document}